\documentclass[information,review,accept,moreauthors,pdftex,10pt,a4paper]{Definitions/mdpi}

\usepackage{amsmath,amssymb,amsfonts}
\usepackage{algorithmicx}
\usepackage[algo2e]{algorithm2e}

\usepackage{pseudocode}

\usepackage{mathtools}
\usepackage{algorithm}
\usepackage{rotating}
\usepackage{breakurl}
\usepackage{array}
\usepackage{graphicx}
\usepackage{color}
\usepackage{textcomp}
\usepackage{esvect}
\usepackage{amsmath,stackengine}

\setitemize{parsep=6pt,itemsep=0pt,leftmargin=*,labelsep=5.5mm}
\setenumerate{parsep=6pt,itemsep=0pt,leftmargin=*,labelsep=5.5mm}
\setlist[description]{itemsep=0mm}
\setitemize{parsep=6pt,itemsep=0pt,leftmargin=*,labelsep=5.5mm,align=parleft}
\setenumerate{parsep=6pt,itemsep=0pt,leftmargin=*,labelsep=5.5mm,align=parleft}

\usepackage{soul}
\usepackage{booktabs}
\usepackage{multirow}
\usepackage{microtype}
\makeatletter
\titleformat{\paragraph}
{\normalfont\normalsize\itshape}{\theparagraph}{1em}{}
\titlespacing*{\paragraph}
{0pt}{1.25ex plus 1ex minus .2ex}{1ex plus .2ex}

\titleformat{\subparagraph}
{\normalfont\normalsize\itshape}{\thesubparagraph}{1em}{}
\titlespacing*{\subparagraph}
{0pt}{1.25ex plus 1ex minus .2ex}{1ex plus .2ex}

\newcommand{\RNum}[1]{%
\textup{\uppercase\expandafter{\romannumeral#1}}%
}

\newcommand{\qu}[1]{``#1''}

\newcommand\given[1][]{\:#1\vert\:}

\DeclareMathOperator*{\argmin}{arg\,min}

\firstpage{1}
\pubvolume{10}
\issuenum{4}
\articlenumber{150}
\pubyear{2019}
\copyrightyear{2019}
\history{Received: 22 March 2019; Accepted: 17 April 2019; Published: 23 April 2019}
\updates{yes} 

\pdfoutput=1

\Title{Text Classification Algorithms: A Survey}

\Author{Kamran Kowsari $^{1,2,}$*\href{https://orcid.org/0000-0002-6451-4786}{\orcidicon}, Kiana Jafari Meimandi~$^{1}$ , Mojtaba Heidarysafa $^{1}$, Sanjana Mendu $^{1}$\orcidD{}, Laura Barnes~$^{1,2,3}$\orcidE{} and Donald Brown~$^{1,3}$\href{https://orcid.org/0000-0002-9140-2632}{\orcidicon}}

\AuthorNames{Kamran Kowsari, Kiana Jafari Meimandi, Mojtaba Heidarysafa, Sanjana Mendu,  Laura E. Barnes and Donald E. Brown}

\address{%
$^{1}$ \quad Department of Systems and Information Engineering, University of Virginia, Charlottesville, VA 22904,
USA;  kj6vd@virginia.edu~(K.J.M.); mh4pk@virginia.edu~(M.H.); sm7gc@virginia.edu~(S.M.); lb3dp@virginia.edu~(L.B.); deb@virginia.edu~(D.B.)\\
$^{2}$ \quad Sensing  Systems  for  Health  Lab, University of Virginia, Charlottesville, VA 22911, USA\\
$^{3}$ \quad School of Data Science, University of Virginia, Charlottesville, VA 22904, USA}

\corres{Correspondence: kk7nc@virginia.edu; Tel.: +1-202-812-3013}

\abstract{In recent years, there has been an exponential growth in the number of complex documents and texts that require a deeper understanding of machine learning methods to be able to accurately classify texts in many applications. Many machine learning approaches have achieved surpassing results in natural language processing. The success of these learning algorithms relies on their capacity to understand complex models and non-linear relationships within data. However, finding suitable structures, architectures, and techniques for text classification is a challenge for researchers. In this paper, a brief overview of text classification algorithms is discussed. This overview covers different text feature extractions, dimensionality reduction methods, existing algorithms and techniques, and evaluations methods. Finally, the limitations of each technique and their application in real-world problems are discussed.}

\keyword{text classification; text mining; text representation; text categorization; text analysis; document classification}

\begin{document}

\section{Introduction}\label{sec:Introduction}
\begin{figure}[H]
\centering
\includegraphics[width=\textwidth]{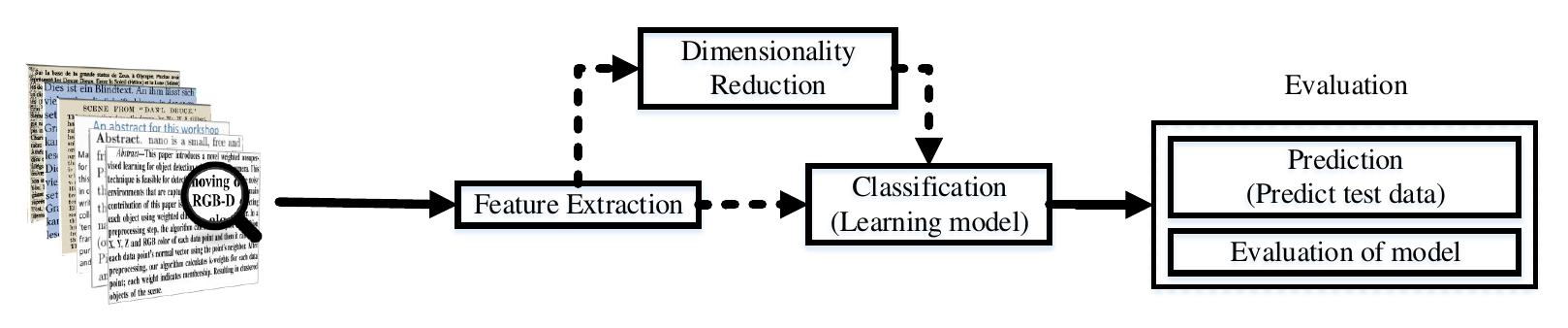}
\caption{Overview of text classification pipeline.} \label{fig:overview}
\end{figure}

\textls[-10]{Text classification problems have been widely studied and addressed in many real applications~\mbox{\cite{jiang2018text,kowsari2017HDLTex,mccallum1998comparison,
Kowsari2018RMDL,Heidarysafa2018RMDL,lai2015recurrent,aggarwal2012survey,aggarwal2012mining}} }
over the last few decades. Especially with recent breakthroughs in Natural Language Processing (NLP) and text mining, many researchers are now interested in developing applications that leverage text classification methods.
Most text classification and document categorization systems can be deconstructed into the following four phases: Feature extraction, dimension reductions, classifier selection, and evaluations. In this paper, we discuss the structure and technical implementations of text classification systems in terms of the pipeline illustrated in Figure~\ref{fig:overview} (The source code and the results are shared as free tools at \url{https://github.com/kk7nc/Text\_Classification}).

The initial pipeline input consists of some raw text data set.
In general, text data sets contain sequences of text in documents as~$D=\{X_1,X_2,\hdots,X_N\}$ where~$X_i$ refers to a data point (i.e., document, text segment) with~$s$ number of sentences such that each sentence includes~$w_s$ words with~$l_w$ letters. Each point is labeled with a class value from a set of~$k$ different discrete value indices~\cite{aggarwal2012survey}.

Then, we should create a structured set for our training purposes which call this section Feature Extraction. The dimensionality reduction step is an optional part of the pipeline which could be part of the classification system~({e.g.}, if we use Term Frequency-Inverse Document Frequency~(TF-IDF) as our feature extraction and in train set we have~$200k$ unique words, computational time is very expensive, so we could reduce this option by bringing feature space in other dimensional space). The~most significant step in document categorization is choosing the best classification algorithm. The other part of the pipeline is the evaluation step which is divided into two parts~(prediction the test set and evaluating the model). In general, the text classification system contains four different levels of scope that can be applied:

\begin{enumerate}
\item \textbf{Document level:} In the document level, the algorithm obtains the relevant categories of a full document.
\item \textbf{Paragraph level:} In the paragraph level, the algorithm obtains the relevant categories of a single paragraph~(a portion of a document).
\item \textbf{Sentence level:} In the sentence level, obtains the relevant categories of a single sentence (a~portion of a paragraph).
\item \textbf{Sub-sentence level:} In the sub-sentence level, the algorithm obtains the relevant categories of sub-expressions within a sentence~(a portion of a sentence
)).
\end{enumerate}

(\RNum{1}) \textbf{Feature Extraction:} In general, texts and documents are unstructured data sets. However, these unstructured text sequences must be converted into a structured feature space when using mathematical modeling as part of a classifier.~First, the data needs to be cleaned to omit unnecessary characters and words. After the data has been cleaned, formal feature extraction methods can be applied. The common techniques of feature extractions are Term Frequency-Inverse Document Frequency~(TF-IDF), Term Frequency~(TF)~\cite{salton1988term}, Word2Vec~\cite{goldberg2014word2vec}, and Global Vectors for Word Representation~(GloVe)~\cite{pennington2014glove}. In Section~\ref{sec:Feature_extraction}, we categorize these methods as either word embedding or weighted word techniques and discuss the technical implementation details.

(\RNum{2}) \textbf{Dimensionality Reduction:}
As text or document data sets often contain many unique words, data pre-processing steps can be lagged by high time and memory complexity. A common solution to this problem is simply using inexpensive algorithms. However, in some data sets, these kinds of cheap algorithms do not perform as well as expected. In order to avoid the decrease in performance, many researchers prefer to use dimensionality reduction to reduce the time and memory complexity for their applications. Using dimensionality reduction for pre-processing could be more efficient than developing inexpensive classifiers.

In Section~\ref{sec:Dimensionality}, we outline the most common techniques of dimensionality reduction, including Principal Component Analysis (PCA), Linear Discriminant Analysis~(LDA), and non-negative matrix factorization~(NMF). We also discuss novel techniques for unsupervised feature extraction dimensionality reduction, such as random projection, autoencoders, and t-distributed stochastic neighbor embedding~(t-SNE).

(\RNum{3}) \textbf{Classification Techniques:} The most important step of the text classification pipeline is choosing the best classifier. Without a complete conceptual understanding of each algorithm, we cannot effectively determine the most efficient model for a text classification application. In Section~\ref{sec:techniques}, we discuss the most popular techniques of text classification. First, we cover traditional methods of text classification, such as Rocchio classification. Next, we talk about ensemble-based learning techniques such as boosting and bagging, which have been used mainly for query learning strategies and text analysis~\cite{mamitsuka1998query,kim2000text,schapire2000boostexter}.
One of the simplest classification algorithms is logistic regression~(LR) which has been addressed in most data mining domains~\cite{harrell2001ordinal,hosmer2013applied,dou2018txt,chen2017comparative}.
In the earliest history of information retrieval as a feasible application, The Na\"ive Bayes Classifier~(NBC) was very popular. We have a brief overview of Na\"ive Bayes Classifier which is computationally inexpensive and also needs a very low amount of memory~\cite{larson2010introduction}.

Non-parametric techniques have been studied and used as classification tasks such as k-nearest neighbor~(KNN)~\cite{li2001gene}.
Support Vector Machine~(SVM)~\cite{manevitz2001one,han2000centroid} is another popular technique which employs a discriminative classifier for document categorization. This technique can also be used in all domains of data mining such as bioinformatics, image, video, human activity classification, safety and security, etc. This model is also used as a baseline for many researchers to compare against their own works to highlight novelty and contributions.

Tree-based classifiers such as decision tree and random forest have also been studied with respect to document categorization~\cite{xu2012improved}. Each tree-based algorithm will be covered in a separate sub-section. In recent years, graphical classifications have been considered~\cite{lafferty2001conditional} as a classification task such as conditional random fields~(CRFs). However, these techniques are mostly used for document summarization~\cite{shen2007document} and automatic keyword extraction~\cite{zhang2008automatic}.

Lately, deep learning approaches have achieved surpassing results in comparison to previous machine learning algorithms on tasks such as image classification, natural language processing, face recognition, etc. The success of these deep learning algorithms relies on their capacity to model complex and non-linear relationships within data~\cite{lecun2015deep}.

(\RNum{4}) \textbf{Evaluation:} The final part of the text classification pipeline is evaluation. Understanding how a model performs is essential to the use and development of text classification methods. There are many methods available for evaluating supervised techniques. Accuracy calculation is the simplest method of evaluation but does not work for unbalanced data sets~\cite{huang2005using}.~In Section~\ref{sec:Evaluation}, we~outline the following evaluation methods for text classification algorithms: $F_{\beta}$ Score~\cite{lock2002acute}, Matthews Correlation Coefficient~(MCC)~\cite{matthews1975comparison}, receiver operating characteristics~(ROC)~\cite{hanley1982meaning}, and area under the ROC curve~(AUC)~\cite{pencina2008evaluating}.

In Section~\ref{sec:Discussion}, we talk about the limitations and drawbacks of the methods mentioned above. We~also briefly compare the steps of pipeline including feature extractions, dimensionality reduction, classification techniques, and evaluation methods. The state-of-the-art techniques are compared in this section by many criteria such as architecture of their model, novelty of the work, feature extraction technique, corpus~(the data set/s used), validation measure, and limitation of each work. Finding the best system for an application requires choosing a feature extraction method. This choice completely depends on the goal and data set of an application as some feature extraction techniques are not efficient for a specific application. For example, since GloVe is trained on Wikipedia and when used for short text messages like short message service~(SMS), this technique does not perform as well as TF-IDF. Additionally, limited data points that this model cannot be trained as well as other techniques due to the small amount of data. The next step or in this pipeline, is a classification technique, where we briefly talk about the limitation and drawbacks of each technique.

In Section~\ref{sec:Application}, we describe the text and document classification applications. Text classification is a major challenge in many domains and fields for researchers. Information retrieval systems~\cite{jacobs2014text} and search engine~\cite{croft2010search,yammahi2014efficient} applications commonly make use of text classification methods. Extending from these applications, text classification could also be used for applications such as information filtering~(e.g., email and text message spam filtering)~\cite{chu2010tweeting}. Next, we talk about adoption of document categorization in public health~\cite{gordon1983operational} and human behavior~\cite{nobles2018identification}. Another area that has been helped by text classification is document organization and knowledge management. Finally, we will discuss recommender systems which are extensively used in marketing and advertising.

\section{Text Preprocessing}\label{sec:Feature_extraction}
Feature extraction and pre-processing are crucial steps for text classification applications.
In this section, we introduce methods for cleaning text data sets, thus removing implicit noise and allowing for informative featurization.
Furthermore, we discuss two common methods of text feature extraction: Weighted word and word embedding techniques.

\subsection{Text Cleaning and Pre-processing}

Most text and document data sets contain many unnecessary words such as stopwords, misspelling, slang, etc.
In many algorithms, especially statistical and probabilistic learning algorithms, noise and unnecessary features can have adverse effects on system performance. In this section, we~briefly explain some techniques and methods for text cleaning and pre-processing text data sets.

\subsubsection{Tokenization}
Tokenization is a pre-processing method which breaks a stream of text into words, phrases, symbols, or other meaningful elements called tokens~\cite{gupta2015text,verma2014tokenization}.~The main goal of this step is the investigation of the words in a sentence~\cite{verma2014tokenization}.~Both text classification and text mining require a parser which processes the tokenization of the documents, for example:\\
sentence~\cite{aggarwal2018machine}:
\begin{quote}
\textit{After sleeping for four hours, he decided to sleep for another four.}
\end{quote}

In this case, the tokens are as follows:
\begin{quote}
\{ \textit{\qu{After} \qu{sleeping} \qu{for} \qu{four} \qu{hours} \qu{he} \qu{decided} \qu{to} \qu{sleep} \qu{for} \qu{another} \qu{four}} \}.
\end{quote}

\subsubsection{Stop Words}
Text and document classification includes many words which do not contain important significance to be used in classification algorithms, such as \{\textit{\qu{a},
\qu{about},
\qu{above},
\qu{across},
\qu{after},
\qu{afterwards},
\qu{again},$\hdots$}\}. The most common technique to deal with these words is to remove them from the texts and documents~\cite{saif2014stopwords}.

\subsubsection{Capitalization}

Text and document data points have a diversity of capitalization to form a sentence. Since documents consist of many sentences, diverse capitalization can be hugely problematic when classifying large documents. The most common approach for dealing with inconsistent capitalization is to reduce every letter to lower case. This technique projects all words in text and document into the same feature space, but it causes a significant problem for the interpretation of some words (e.g.,~“\textit{US}”~(United States of America) to “\textit{us}”~(pronoun))~\cite{gupta2009survey}. Slang and abbreviation converters can help account for these exceptions~\cite{dalal2011automatic}.

\subsubsection{Slang and Abbreviation}
Slang and abbreviation are other forms of text anomalies that are handled in the pre-processing step. An abbreviation~\cite{whitney2010abbreviations} is a shortened form of a word or phrase which contain mostly first letters form the words, such as SVM which stands for Support Vector Machine.

Slang is a subset of the language used in informal talk or text that has different meanings such as \qu{lost the plot}, which essentially means that they've gone mad~\cite{helm2003recovery}. A common method for dealing with these words is converting them into formal language~\cite{dhuliawala2016slangnet}.

\subsubsection{Noise Removal}

Most of the text and document data sets contain many unnecessary characters such as punctuation and special characters. Critical punctuation and special characters are important for human understanding of documents, but it can be detrimental for classification algorithms~\cite{pahwasentiment}.

\subsubsection{Spelling Correction}
Spelling correction is an optional pre-processing step. Typos (short for typographical errors) are commonly present in texts and documents, especially in social media text data sets (e.g., Twitter). Many algorithms, techniques, and methods have addressed this problem in NLP~\cite{mawardi2018spelling}. Many techniques and methods are available for researchers including hashing-based and context-sensitive spelling correction techniques~\cite{dziadek2017improving}, as well as spelling correction using Trie and Damerau--Levenshtein distance bigram~\cite{mawardi2018fast}.

\subsubsection{Stemming}
In NLP, one word could appear in different forms (i.e., singular and plural noun form) while the semantic meaning of each form is the same~\cite{spirovski2018comparison}. One method for consolidating different forms of a word into the same feature space is stemming. Text stemming modifies words to obtain variant word forms using different linguistic processes such as affixation (addition of affixes)~\cite{singh2016text,sampson2005language}. For example, the stem of the word ``studying'' is ``study''.

\subsubsection{Lemmatization}
Lemmatization is a NLP process that replaces the suffix of a word with a different one or removes the suffix of a word completely to get the basic word form~(lemma)~\cite{plisson2004rule,korenius2004stemming,sampson2005language}.

\subsection{Syntactic Word Representation}
Many researchers have worked on this text feature extraction technique to solve the loosing syntactic and semantic relation between words. Many researchers addressed novel techniques for solving this problem, but many of these techniques still have limitations. In~\cite{caropreso2006beyond}, a model was introduced in which the usefulness of including syntactic and semantic knowledge in the text representation for the selection of sentences comes from technical genomic texts.
The other solution for syntactic problem is using the n-gram technique for feature extraction.

\subsubsection{N-Gram}

The n-gram technique is a set of~$n\text{-}word$ which occurs \qu{in that order} in a text set. This is not a representation of a text, but it could be used as a feature to represent a text.

BOW
is a representation of a text using its words~$(1\text{-}gram)$ which loses their order~(syntactic). This model is very easy to obtain and the text can be represented through a vector, generally of a manageable size of the text. On the other hand,~$n\text{-}gram$ is a feature of BOW for a representation of a text using~$1\text{-}gram$. It is very common to use~$2\text{-}gram$ and $3\text{-}gram$. In this way, the text feature extracted could detect more information in comparison to~$1\text{-}gram$.

\subsubsection*{An Example of $2\text{-}Gram$}

\begin{quote}
\textit{After sleeping for four hours, he decided to sleep for another four.}
\end{quote}

In this case, the tokens are as follows:

\begin{quote}
\{ \textit{\qu{After sleeping}, \qu{sleeping for}, \qu{for four}, \qu{four hours}, \qu{four he} \qu{he decided}, \qu{decided to}, \qu{to~sleep}, \qu{sleep for}, \qu{for another}, \qu{another four}} \}.
\end{quote}

\subsubsection*{An Example of $3\text{-}Gram$}

\begin{quote}
\textit{After sleeping for four hours, he decided to sleep for another four.}
\end{quote}

In this case, the tokens are as follows:

\begin{quote}
\{ \textit{\qu{After sleeping for}, \qu{sleeping for four}, \qu{four hours he}, \qu{ hours he decided}, \qu{he decided to}, \qu{to~sleep for}, \qu{sleep for another}, \qu{for another four}} \}.
\end{quote}

\subsubsection{Syntactic N-Gram}

In~\cite{sidorov2012syntactic}, syntactic n-grams are discussed which is defined by paths in syntactic dependency or constituent trees rather than the linear structure of the text.

\subsection{Weighted Words}
The most basic form of weighted word feature extraction is TF, where each word is mapped to a number corresponding to the number of occurrences of that word in the whole corpora. Methods that extend the results of TF generally use word frequency as a boolean or logarithmically scaled weighting. In all weight words methods, each document is translated to a vector (with length equal to that of the document) containing the frequency of the words in that document. Although this approach is intuitive, it is limited by the fact that particular words that are commonly used in the language may dominate such representations.

\subsubsection{Bag of Words~(BoW)}
The bag-of-words model (BoW model) is a reduced and simplified representation of a text document from selected parts of the text, based on specific criteria, such as word frequency.

The BoW technique is used in several domains such as computer vision, NLP, Bayesian spam filters, as well as document classification and information retrieval by Machine Learning.

In a BoW, a body of text, such as a document or a sentence, is thought of like a bag of words. Lists of words are created in the BoW process. These words in a matrix are not sentences which structure sentences and grammar, and the semantic relationship between these words are ignored in their collection and construction. 
The words are often representative of the content of a sentence. While grammar and order of appearance are ignored, multiplicity is counted and may be used later to determine the focus points of the documents.

Here is an example of BoW:

\subsubsection*{Document}

\begin{quote}
\textit{\qu{As the home to UVA's recognized undergraduate and graduate degree programs in systems engineering. In the UVA Department of Systems and Information Engineering, our students are exposed to a wide range of range}}
\end{quote}

\subsubsection*{Bag-of-Words~(BoW)}

\begin{quote}
\textit{\{\qu{As}, \qu{the}, \qu{home}, \qu{to}, \qu{UVA's}, \qu{recognized}, \qu{undergraduate}, \qu{and}, \qu{graduate}, \qu{degree}, \qu{program}, \qu{in}, \qu{systems}, \qu{engineering}, \qu{in}, \qu{Department}, \qu{Information},\qu{students}, \qu{ },\qu{are}, \qu{exposed}, \qu{wide}, \qu{range} \}}
\end{quote}

\subsubsection*{Bag-of-Feature~(BoF)}

\begin{quote}
Feature = \{1,1,1,3,2,1,2,1,2,3,1,1,1,2,1,1,1,1,1,1\}
\end{quote}

\subsubsection{Limitation of Bag-of-Words}
Bag0of-words models encode every word in the vocabulary as one-hot-encoded vector e.g., for the vocabulary of size $|\Sigma|$, each word is represented by a $|\Sigma|$ dimensional sparse vector with~$1$ at index corresponding to the word and 0 at every other index. As vocabulary may potentially run into millions, bag-of-word models face scalability challenges~({e.g.}, \qu{This is good} and \qu{Is this good} have exactly the same vector representation).
The technical problem of the bag-of-word is also the main challenge for the computer science and data science community.

Term frequency, also called bag-of-words, is the simplest technique of text feature extraction. This~method is based on counting the number of words in each document and assigning it to the feature space.

\subsubsection{Term Frequency-Inverse Document Frequency}

K. Sparck Jones~\cite{sparck1972statistical}
proposed Inverse Document Frequency~(IDF) as a method to be used in conjunction with term frequency in order to lessen the effect of implicitly common words in the corpus. IDF assigns a higher weight to words with either high or low frequencies term in the document.
This combination of TF and IDF is well known as Term Frequency-Inverse document frequency (TF-IDF). The mathematical representation of the weight of a term in a document by TF-IDF is given in Equation~\eqref{tf-idf}.
\begin{equation}\label{tf-idf}
W(d,t)=TF(d,t)* log(\frac{N}{df(t)})
\end{equation}

Here \emph{N}
is the number of documents and~$df(t)$ is the number of documents containing the term \emph{t} in the corpus. The first term in Equation~\eqref{tf-idf} improves the recall while the second term improves the precision of the word embedding~\cite{tokunaga1994text}. Although TF-IDF tries to overcome the problem of common terms in the document, it still suffers from some other descriptive limitations. Namely, TF-IDF cannot account for the similarity between the words in the document since each word is independently presented as an index.
However, with the development of more complex models in recent years, new methods, such as word embedding, have been presented that can incorporate concepts such as similarity of words and part of speech tagging.

\subsection{Word Embedding}
Even though we have syntactic word representations, it does not mean that the model captures the semantics meaning of the words.
On the other hand, bag-of-word models do not respect the semantics of the word. For example, words \textit{\qu{airplane}, \qu{aeroplane}, \qu{plane}}, and \textit{\qu{aircraft}} are often used in the same context. However, the vectors corresponding to these words are orthogonal in the bag-of-words model. This issue presents a serious problem to understanding  sentences within the model. The other problem in the bag-of-word is that the order of words in the phrase is not respected. The $n\text{-}gram$ does not solve this problem so a similarity needs to be found for each word in the sentence. Many researchers worked on word embedding to solve this problem. The
Skip-gram and continuous bag-of-words (CBOW) models of~\cite{mikolov2013efficient} propose a simple single-layer architecture based on the inner product between two word vectors.

Word embedding is a feature learning technique in which each word or phrase from the vocabulary is mapped to a~$N$ dimension vector of real numbers. Various word embedding methods have been proposed to translate unigrams into understandable input for machine learning algorithms. This~work focuses on Word2Vec, GloVe, and FastText, three of the most common methods that have been successfully used for deep learning techniques. Recently, the Novel technique of word representation was introduced where word vectors depend on the context of the word called \textit{\qu{Contextualized Word Representations}} or \textit{\qu{Deep Contextualized Word Representations}}.

\subsubsection{Word2Vec}

{T. Mikolov et~al.}~\cite{mikolov2013efficient,mikolov2013distributed} presented \textit{"word to vector"} representation as an improved word embedding architecture. The Word2Vec approach uses shallow neural networks with two hidden layers, continuous bag-of-words~(CBOW), and the Skip-gram model to create a high dimension vector for each word. The Skip-gram model dives a corpus of words~$w$ and context~$c$~\cite{goldberg2014word2vec}. The goal is to maximize the probability:
\begin{equation}
arg \max_\theta \prod_{w \in T} \bigg[ \prod_{c\in c(w)} p(c\given w;\theta) \bigg]
\end{equation}
where~$T$ refers to Text, and $\theta$ is parameter of $p(c\given w;\theta)$.

\begin{figure}[H]
\centering
\includegraphics[width=\textwidth]{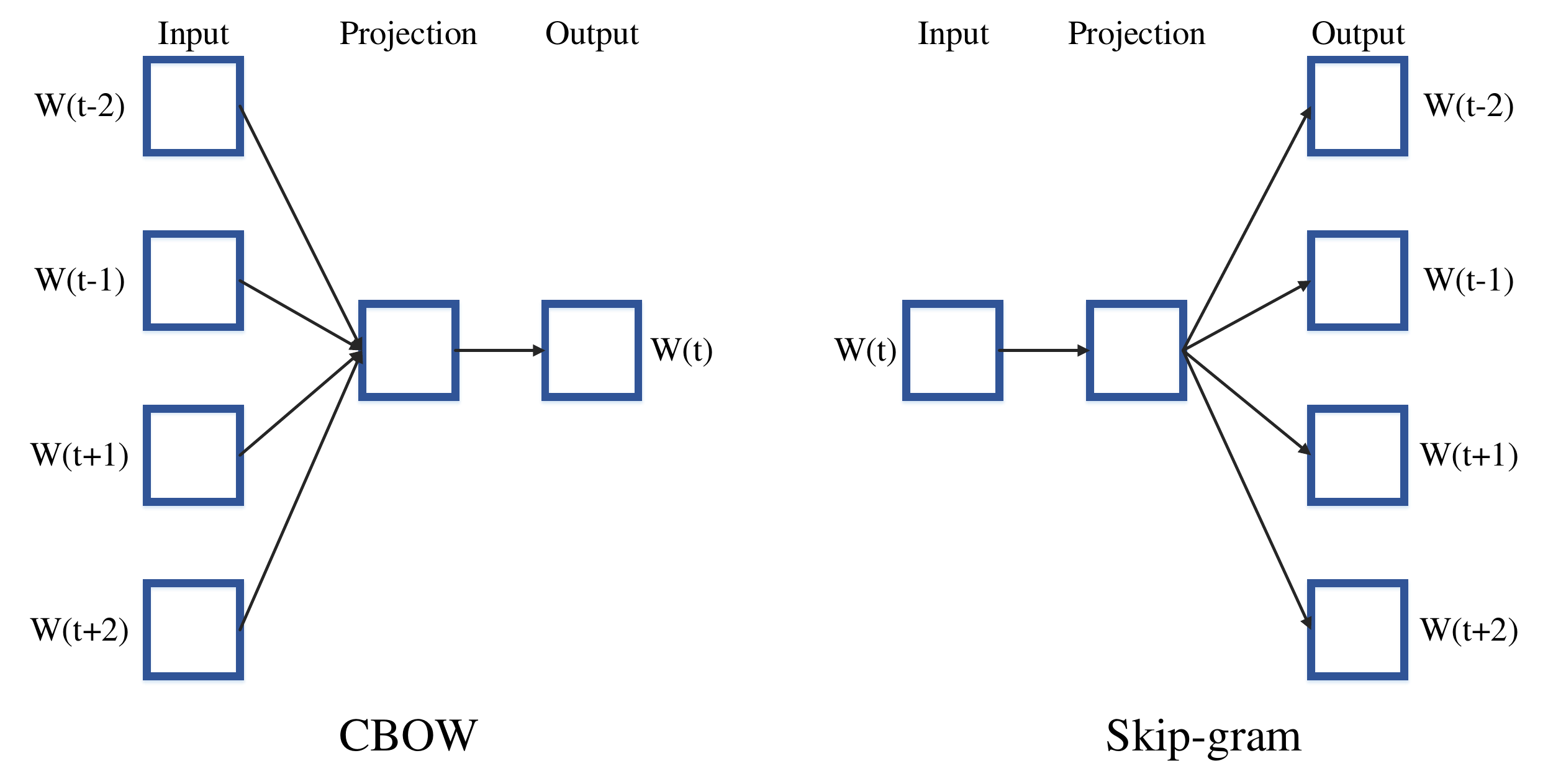}
\caption{The continuous bag-of-words (CBOW) architecture predicts the current word based on the
context, and the Skip-gram predicts surrounding words based on the given current word~\cite{mikolov2013efficient}.}
\label{fig:CBOW}
\end{figure}

Figure~\ref{fig:CBOW} shows a simple CBOW model which tries to find the word based on previous words, while Skip-gram tries to find words that might come in the vicinity of each word.
The weights between the input layer
and output layer represent $v\times N$~\cite{rong2014word2vec} as a matrix of $w$.
\begin{equation}
h = W^Tc = W^T_{k,.} := v^T_{wI}
\end{equation}

This method provides a very powerful tool for discovering relationships in the text corpus as well as similarity between words. For example, this embedding would consider the two words such as ``big'' and ``bigger'' close to each other in the vector space it assigns them.

\subsubsection*{Continuous Bag-of-Words Model}

The continuous bag-of-words model is represented by multiple words for a given target of words. For example, the word \qu{airplane} and \qu{military} as context words for~\qu{air-force} as the target word. This consists of replicating the input to hidden layer connections $\beta$ times which is the number of context words~\cite{mikolov2013efficient}. Thus, the bag-of-words model is mostly used to represent an unordered collection of words as a vector. The first thing to do is create a vocabulary, which means all the unique words in the corpus. The output of the shallow neural network will be~$w_i$ that the task as \textit{\qu{predicting the word given its context}}. The number of words used depends on the setting for the window size~(common size is 4--5 words).

\subsubsection*{Continuous Skip-Gram Model}
Another model architecture which is very similar to CBOW~\cite{mikolov2013efficient} is the continuous Skip-gram model, however this model, instead of predicting the current word based on the context, tries to maximize classification of a word based on another word in the same sentence. The continuous bag-of-words model and continuous Skip-gram model are used to keep syntactic and semantic information of sentences for machine learning algorithms.

\subsubsection{Global Vectors for Word Representation~(GloVe)}
\begin{figure}[H]
\centering
\includegraphics[width=0.5\textwidth]{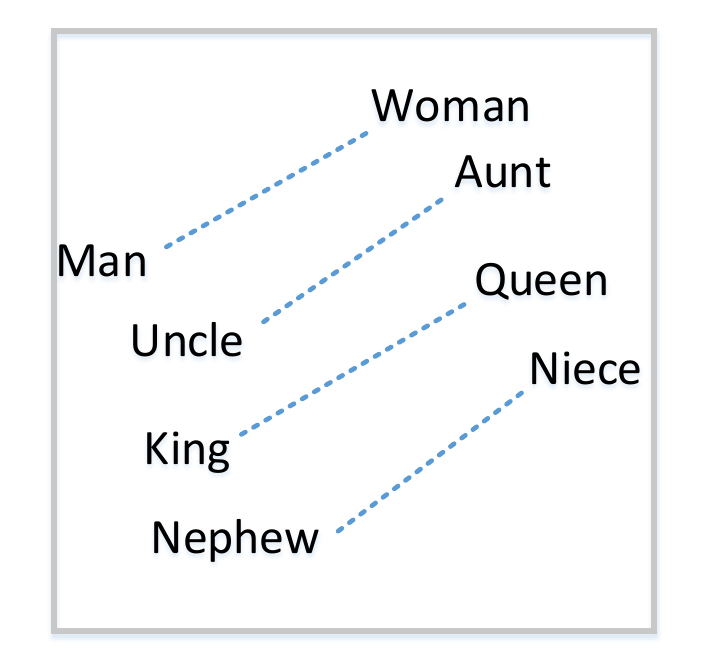}
\caption{GloVe: Global Vectors for Word Representation.} \label{fig:Glove}
\end{figure}

Another powerful word embedding technique that has been used for text classification is Global Vectors~(GloVe)~\cite{pennington2014glove}. The approach is very similar to the Word2Vec method, where each word is presented by a high dimension vector and trained based on the surrounding words over a huge corpus. The pre-trained word embedding used in many works is based on 400,000 vocabularies trained over Wikipedia~2014 and Gigaword~5 as the corpus and~50 dimensions for word presentation. GloVe also provides other pre-trained word vectorizations with~100, 200, 300 dimensions which are trained over even bigger corpora, including Twitter content. Figure~\ref{fig:Glove} shows a visualization of the word distances over a sample data set using the same t-SNE technique~\cite{maaten2008visualizing}. The objective function is as follows:
\begin{equation}
f(w_i-w_j,\widetilde{w}_k) =\frac{P_{ik}}{P_{jk}}
\end{equation}
where~$w_i$ refers to the word vector of word~$i$, and~$P_{ik}$ denotes to the probability of word~$k$ to occur in the context of word~$i$.

\subsubsection{FastText}
Many other word embedding representations ignore the morphology of words by assigning a distinct vector to each word~\cite{bojanowski2016enriching}. Facebook AI Research lab released a novel technique to solve this issue by introducing a new word embedding method called~FastText.
Each word,~$w$, is represented as a bag of character n-gram. For example, given the word~\qu{\textit{introduce}} and~$n = 3$, FastText will produce the following representation composed of character tri-grams:
\begin{equation*}
<in,~int,~ntr,~tro,~rod,~odu,~duc,~uce,~ce >
\end{equation*}

Note that the sequence <\emph{int}>,
corresponding to the word here is different from the tri-gram ``\emph{int}'' from the word introduce.

Suppose we have a dictionary of n-grams of size~$G$, and given a word~$w$ which is associated as a vector representation $z_g$ to each n-gram~$g$. The obtained scoring function~\cite{bojanowski2016enriching} in this case is:

\begin{equation}
s(w,c)=\sum_{g \in g_w} z_g^Tv_c
\end{equation}
where~$g_w \in \{1,2,\hdots,G\}$ .

Facebook published pre-trained word vectors for~$294$ languages which are trained on Wikipedia using FastText based on~$300$ dimension. The FastText used the Skip-gram model~\cite{bojanowski2016enriching} with default parameters.

\subsubsection{Contextualized Word Representations}
Contextualized word representations are another word embedding technique which is based on the context2vec~\cite{melamud2016context2vec} technique introduced by {B. McCann et al}. The context2vec method uses bidirectional long short-term memory (LSTM). {M.E. Peters~et al.}~\cite{peters2018deep} built upon this technique to create the deep contextualized word representations technique. This technique contains both the main feature of word representation:
(\RNum{1}) complex characteristics of word use ({e.g.,} syntax and semantics) and
(\RNum{2}) how these uses vary across linguistic contexts~({e.g.,} to model polysemy)~\cite{peters2018deep}.

The main idea behind these word embedding techniques is that the resulting word vectors are learned from a bidirectional language model~(biLM), which consist of both forward and backward~LMs.

The forward LMs are as follows:
\begin{equation}
p(t_1,t_2,\dots,t_N) = \prod_{k=1}^N p(t_k|t_1,t_2,\dots, t_{k-1})
\end{equation}

The backward LMs are as follows:
\begin{equation}
p(t_1,t_2,\dots,t_N) = \prod_{k=1}^N p(t_k|t_{k+1},t_{k+2},\dots, t_N)
\end{equation}

This formulation jointly maximizes the log-likelihood of the forward and backward directions as~follows:
\begin{equation}
\sum_{k=1}^N \left(
\begin{split}
&\log p(t_k|t_1,\dots, t_{k-1};\Theta_x, \overrightarrow{\Theta}_{LSTM},\Theta_s ) +\\  &\log p(t_k|t_{k+1},\dots, t_N;\Theta_x, \overleftarrow{\Theta}_{LSTM},\Theta_s )
\end{split}
\right)
\end{equation}
where $\Theta_x$ is the token representation and  $\Theta_x$ refers to the softmax layer. Then, ELMo
is computed as a task-specific weighting for all biLM layers as follows:
\begin{equation}
\begin{split}
ELMo_k^{task} = E(R_k; \Theta^{task})
= \gamma^{task} \sum_{j=0}^L s_j^{task} h_{k,j}^{LM}
\end{split}
\end{equation}
where  $h_{k,j}^{LM}$ is calculated by:
\begin{equation}
h_{k,j}^{LM} = \left[\overrightarrow{h}_{k,j}^{LM},\overleftarrow{h}_{k,j}^{LM}\right]
\end{equation}
where $s^{task}$ stands for softmax-normalized weights, and $\gamma^{task}$ is the scalar parameter.

\subsection{Limitations}
Although the continuous bag-of-words model and continuous Skip-gram model are used to keep syntactic and semantic information of per-sentences for machine learning algorithms, there remains the issue how to  keep full meaning of coherent documents for machine learning.

\vspace{6pt}
\noindent\textbf{Example:}
\vspace{6pt}

\noindent Document: \{\qu{\textit{Maryam went to Paris on July 4th, 2018. She missed the independence day fireworks and celebrations. This day is a federal holiday in the United States commemorating the Declaration of Independence of the United States on July 4, 1776. The Continental Congress declared that the thirteen American colonies were no longer subject to the monarch of Britain and were now united, free, and independent states. She wants to stay in the country for next year and celebrate with her friends.}}\}

\vspace{6pt}
\noindent\textbf{Sentence level of this document:}
\vspace{6pt}

\noindent S1: \{\qu{\textit{Maryam went to Paris on July 4th, 2018.}}\}\\
\noindent S2: \{\qu{\textit{She missed the independence day fireworks and celebrations.}}\}\\
\noindent S3: \{\qu{\textit{This day is a federal holiday in the United States commemorating the Declaration of Independence of the United States on July 4, 1776.}}\}\\
\noindent S4: \{\qu{\textit{The Continental Congress declared that the thirteen American colonies were no longer subject to the monarch of Britain and were now united, free, and independent states.}}\}\\
\noindent S5: \{\qu{\textit{She has a plan for next year to stay in the country and celebrate with her friends.}}\}

\vspace{6pt}
\noindent\textbf{Limitation:}
\vspace{6pt}

\begin{figure}[H]
\centering
\includegraphics[width=0.9\textwidth]{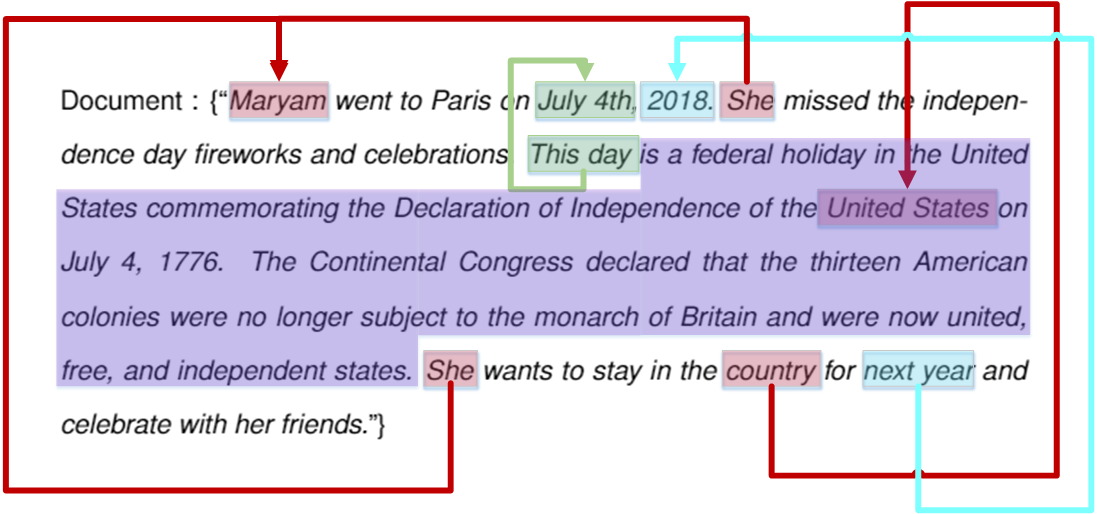}
\caption{Limitation of document feature extraction by per-sentence level.} \label{Document_per}
\end{figure}

Figure~\ref{Document_per} shows how the feature extraction fails for per-sentence level. The~purple color shown in figure is the brief history of~\qu{This day}. Furthermore,~\qu{This day} refers to ~\qu{July 4th}. In S5, \qu{She} refers to the S1~\qu{Maryam}.

\section{Dimensionality Reduction}\label{sec:Dimensionality}
Text sequences in term-based vector models consist of many features. Thus, time complexity and memory consumption are very expensive for these methods. To address this issue, many researchers use dimensionality reduction to reduce the size of feature space. In this section, existing dimensionality reduction algorithms are discussed in detail.

\subsection{Component Analysis}\unskip
\subsubsection{Principal Component Analysis (PCA)}
Principal component analysis~(PCA) is the most popular technique in multivariate analysis and dimensionality reduction. PCA is a method to identify a subspace in which the data approximately lies~\cite{abdi2010principal}. This means finding new variables that are uncorrelated and maximizing the variance to ``preserve as much variability as possible''~\cite{jolliffe2016principal}.

Suppose a data set ${x^{(i)}; i=1,...,m}$ is given and $x^{(\emph{i})} \in \mathbb{R}^n$ for each \emph{i}
($n \ll m$). The $j$th column of matrix $X$ is vector, $x_j$ that is the observations on the $j$th variable. The linear combination of $x_j$s can be written as:
\begin{equation}
\sum_{j=1}^{m}a_jx_j=Xa
\end{equation}
where $a$ is a vector of constants $a_1,a_2,...,a_m$. The variance of this linear combination can be given as:
\begin{equation}
var(Xa)=a^TSa
\end{equation}
where $S$ is the sample co-variance matrix. The goal is to find the linear combination with maximum variance. This translates into maximizing $a^TSa-\lambda(a^Ta-1)$, where $\lambda$ is a Lagrange multiplier.

PCA can be used as a pre-processing tool to reduce the dimension of a data set before running a supervised learning algorithm on it ($x^{(i)}s$
as inputs). PCA is also a valuable tool as a noise reduction algorithm and can be helpful in avoiding the over-fitting problem \cite{ng2015principal}.
kernel principal component analysis~(KPCA) is another dimensionality reduction method that generalizes linear PCA into the nonlinear case by using the kernel method~\cite{cao2003comparison}.

\subsubsection{Independent Component Analysis (ICA)}
Independent component analysis (ICA) was introduced by {H. Jeanny}~\cite{herault1984reseaux}.~This technique was then further developed by {C. Jutten and J. Herault}~\cite{jutten1991blind}. ICA is a statistical modeling method where the observed data are expressed as a linear transformation~\cite{hyvarinen2001topographic}.~Assume that~4$n$ linear mixtures~($x_1,x_2,\hdots,x_n$) are observed where independent components:
\begin{equation}
x_j=a_{j1}s_1+a_{j2}s_2+\hdots+a_{jn}s_n~\forall j
\end{equation}

The vector-matrix notation
is written as:
\begin{equation}
X=As
\end{equation}

Denoting them by~$a_i$, the model can also be written~\cite{hyvarinen2000independent}
as follows:
\begin{equation}
X=\sum_{i=1}^n a_is_i
\end{equation}

\subsection{Linear Discriminant Analysis (LDA)}
LDA is a commonly used technique for data classification and dimensionality reduction~\cite{sugiyama2007dimensionality}.~LDA is particularly helpful where the within-class frequencies are unequal and their performances have been evaluated on randomly generated test data. Class-dependent and class-independent transformation are two approaches to LDA in which the ratio of between class variance to within class variance and the ratio of the overall variance to within class variance are used respectively~\cite{balakrishnama1998linear}.

Let $x_i \in \mathbb{R}^d$ which be $d$-dimensional samples and $y_i \in \{1,2,..., c\}$ be associated target or output~\cite{sugiyama2007dimensionality}, where~$n$ is the number of documents and~$c$ is the number of categories. The number of samples in each class is calculated as follows:
\begin{equation}
S_w = \sum_{l=1}^c s_l
\end{equation}
where
\begin{equation}
S_i = \sum_{x\in w_i} (x-\mu_i)(x-\mu_i)^T,~\mu_i = \frac{1}{N_i}\sum_{x\in w_i}x
\end{equation}

The generalization between the class scatter matrix is defined as follows:
\begin{equation}
S_B = \sum_{i=1}^c N_i(\mu_i-\mu)(\mu_i-\mu)^T
\end{equation}
where
\begin{equation}
\mu=\frac{1}{N}\sum_{\forall x} x
\end{equation}

Respect to~$c-1$ projection vector of~$w_i$ that can be projected into~$W$ matrix:
\begin{equation}
W = \left[ w_1|w_2|\hdots|w_{c-1}\right]
\end{equation}
\begin{equation}
y_i = w_i^Tx
\end{equation}

Thus, the~$\mu$ (mean) vector and~$S$ matrices (scatter matrices) for the projected to lower dimension
as follows:
\begin{align}
\widetilde{S}_w =& \sum_{i=1}^c \sum_{y\in w_i} (y-\widetilde{\mu}_i)(y-\widetilde{\mu}_i)^T\\
\widetilde{S}_B =& \sum_{i=1}^c (\widetilde{\mu}_i-\widetilde{\mu})(\widetilde{\mu}_i-\widetilde{\mu})^T
\end{align}

If the projection is not scalar ($c-1$ dimensions), the determinant of the scatter matrices will be used as follows:
\begin{equation}
J(W) = \frac{|\widetilde{S}_B|}{|\widetilde{S}_W|}
\end{equation}

From the fisher discriminant analysis~(FDA)~\cite{sugiyama2006local,sugiyama2007dimensionality}, we can re-write the equation as:
\begin{equation}
J(W) = \frac{|W^TS_BW|}{|W^TS_WW|}
\end{equation}

\subsection{Non-Negative Matrix Factorization (NMF)}

Non-negative matrix factorization~(NMF) or non-negative matrix approximation has been shown to be a very powerful technique for very high dimensional data such as text and sequences analysis~\cite{pauca2004text}. This technique is a promising method for dimension reduction~\cite{tsuge2001dimensionality}. In this section, a brief overview of NMF is discussed for text and document data sets.
Given a non-negative~$n\times m$ in matrix~$V$ is an approximation of:
\begin{equation}
V \approx W H
\end{equation}
where~$W=\mathbb{R}^{n\times r}$ and~$H= \mathbb{R}^{r \times m}$. Suppose~$(n+m)~r <nm$, then the product~$WH$ can be regarded as a compressed form of the data in~$V$. Then~$v_i$ and~$h_i$ are the corresponding columns of~$V$
and~$H$. The~computation of each corresponding column can be re-written as follows:
\begin{equation}
u_i\approx W h_i
\end{equation}

The computational time of each iteration, as introduced by {S. Tsuge et al.}~\cite{tsuge2001dimensionality}, can be written as~follows:
\begin{equation}
\overline{H}_{ij} = H_{ij} \frac{(W^TV)_{ij}}{(W^TWH)_{ij}}
\end{equation}
\begin{equation}
\overline{W}_{ij} = W_{ij} \frac{(VH^T)_{ij}}{(WHH^T)_{ij}}
\end{equation}

Thus, the local minimum of the objective function is calculated as follows:
\begin{equation}
F = \sum_i\sum_j(V_{ij}-(WH)_{ij})^2
\end{equation}

The maximization of the objective function can be re-written as follows:
\begin{equation}
F = \sum_i\sum_j(V_{ij}\log((WH)_{ij})-(WH)_{ij})
\end{equation}

The objective function, given by the  Kullback--Leibler~\cite{kullback1951information,johnson2001symmetrizing} divergence, is defined as follows:
\begin{align}
\overline{H}_{ij} &= H_{ij} \sum_k W_{kj}\frac{V_{kj}}{(WH)_{kj}}\\
\hat{W}_{ij} &= W_{ij} \sum_k{\frac{V_{ik}}{(WH)_{ik}}}H_{jk}\\
\overline{W}_{ij}&= \frac{\hat{W_{ij}}}{\sum_k\hat{W_{kj}}}
\end{align}

This NMF-based dimensionality reduction contains the following 5 steps~\cite{tsuge2001dimensionality} (step \RNum{6} is optional but commonly used in information retrieval:

\begin{enumerate}[label=(\Roman*),labelsep=4.5mm]
\item Extract index term after pre-processing stem like feature extraction and text cleaning as discussed in~Section~\ref{sec:Feature_extraction}. Then we have~$n$ documents with~$m$ features;
\item Create~$n$ documents~($d\in\{d_1,d_2,\hdots,d_n\}$), where vector $a_{ij}=L_{ij}\times G_i$ where~$L_{ij}$ refers to local weights of~$i_{-th}$ term in document~$j$, and~$G_i$ is global weights for document~$i$;
\item Apply NMF to all terms in all documents one by one;
\item Project the trained document vector into~$r$-dimensional space;
\item Using the same transformation, map the test set into the $r$-dimensional space;
\item Calculate the similarity between the transformed document vectors and a query vector.
\end{enumerate}

\subsection{Random Projection}

\begin{figure}[H]
\centering
\includegraphics[width=0.764\textwidth]{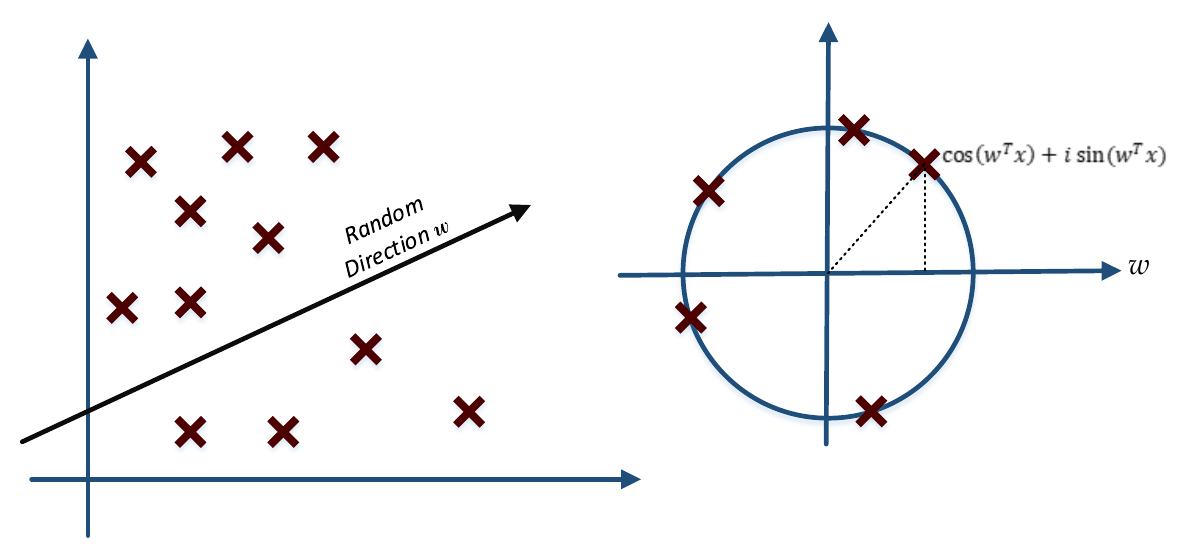}
\caption{The plot on the left shows how we generate random direction, and the plot on the right shows how we project the data set into the new space using complex numbers.} \label{RKS_cos}
\end{figure}

Random projection is a novel technique for dimensionality reduction which is mostly used for high volume data set or high dimension feature spaces. Texts and documents, especially with weighted feature extraction, generate a huge number of features.
Many researchers have applied random projection to text data~\cite{bingham2001random,chakrabarti2003fast} for text mining, text classification, and dimensionality reduction.
In this section, we review some basic random projection techniques. As shown in Figure~\ref{RKS_cos}, the overview of random projection is shown.

\subsubsection{Random Kitchen Sinks}\label{RKS}
The key idea of random kitchen sinks~\cite{rahimi2009weighted} is sampling via monte carlo integration~\cite{morokoff1995quasi} to approximate the kernel as part of dimensionality reduction. This technique works only for shift-invariant kernel:
\begin{equation}
K(x,x') = < \phi (x) , \phi (x') > \approx  K(x - x')
\end{equation}
where shift-invariant kernel, which is an approximation kernel of:
\begin{equation}
K(x - x') = z(x)z(x')
\end{equation}
\begin{equation}
K(x,x') = \int_{R^D} { P(w) e^{iw^T(x-x')} }
\end{equation}
where $D$ is the target number of samples, $P(w)$ is a probability distribution, $w$ stands for random direction, and $w \in \mathbb{R}^{F \times D}$ where $F$ is the number of features and $D$ is the target.

\unskip

\begin{equation}
K(x,x') = K(x-x') \approx \frac{1}{D}{\sum_{j=1}^{D} e^{i w_j^T(x-x')}}
\end{equation}
\begin{equation}
\left. \begin{aligned}
\frac{1}{D}{\sum_{j=1}^{D} e^{i w_j^T(x-x')}}=
& \frac{1}{D}{\sum_{j=1}^{D} e^{i w_j^Tx}   e^{i w_j^Tx'}}  =\\
& \frac{1}{\sqrt{D}}{\sum_{j=1}^{D} e^{i w_j^Tx}} \frac{1}{\sqrt{D}}{\sum_{j=1}^{D}  e^{i w_j^Tx'}}
\end{aligned}\right.
\end{equation}
\begin{equation}
k(x-x') \approx \phi(x) \phi(x')
\end{equation}
\begin{equation}
\phi(x)  = \cos(w^Tx + b_i)
\end{equation}
where the $b_i$ is uniform random variable  ( $b_i \in [0,\pi]$ ).

\subsubsection{Johnson Lindenstrauss Lemma}
William B. Johnson and Joram Lindenstrauss \cite{johnson1984extensions,dasgupta2003elementary} proved that for any $n$ point Euclidean space can be bounded in $k = O(\frac{log n}{\epsilon^2})$
for any ~$u~$ and ~$v \in n$ and $~n \in \mathbb{R}^d$: $\exists f: \mathbb{R}^d \rightarrow \mathbb{R}^k | \epsilon \in [0,1]$. With $x = u-v$ to the lower bound of the success probability.

\begin{equation}
(i-\epsilon)||u-v||^2 \leq ||f(u) - f (v) ||^2 \leq (i+ \epsilon)||u-v||^2
\end{equation}

\subsubsection*{Johnson Lindenstrauss Lemma Proof~\cite{Vempala2009}:}

For any $V$ sets of data point from $n$ where $V \in n$ and random variable $w \in R^{k\times d}$:
\begin{equation}
Pr[success] \geq 1 - 2 m^2 e^ {\frac{-k(\epsilon^2 - \epsilon^3)}{4}}
\end{equation}

If we let $k=  \frac{ 16 log n}{\epsilon^2}$ :
\begin{equation}
\begin{split}
1 - 2 m^2 e^ {\frac{-k(\epsilon^3 - \epsilon^3)}{4}} \geq  1 - 2 m^2 e^ {\frac{(-\frac{ 8 log n}{\epsilon^2}) (\epsilon^2 - \epsilon^3)}{4}} \\
1 - 2 m^2 e^ {\frac{-\frac{ 16 log n}{\epsilon^2}) (\epsilon^3 - \epsilon^3)}{4}}  =\\
1-2 m^{4\epsilon -2} > 1-2m^{- \frac{1}{2}}  > 0
\end{split}
\end{equation}

\subsubsection*{Lemma 1 Proof~\cite{Vempala2009}:}

Let $\Psi$ be a random variable with $k$ degrees of freedom, then for $\epsilon \in [0,1]$
\begin{equation} \label{equation_lemma1_1}
Pr[ (1- \epsilon)k \leq \Psi \leq (1+ \epsilon) k] \geq 1 - 2 e^ {\frac{-k(\epsilon^2 - \epsilon^3)}{4}}
\end{equation}

We start with Markov's inequality~\cite{mao2006stochastic}:
\begin{equation}
Pr[(\Psi \geq (1 - \epsilon) k)] \leq \frac{E [\Psi]}{(1- \epsilon)k}
\end{equation}
\begin{equation}
\begin{split}
Pr[e^{\lambda \Psi} \geq e^{\lambda (1- \epsilon) k)}] \leq \frac{E [e^{\lambda \Psi}]}{e^{\lambda (1- \epsilon) k}} \\
E[e^{\lambda \Psi}] = (1-2 \lambda)^{-\frac{k}{2}}
\end{split}
\end{equation}
where $\lambda~<~ 0.5$ and using the fact of $(1- \epsilon) \leq e^{\epsilon - \frac{(\epsilon^2 - \epsilon^3)}{2}}$; thus, we can proof $Pr[(\Psi \geq (1 - \epsilon) k)] \leq \frac{E [\Psi]}{(1- \epsilon) k} $ and the $Pr[(\Psi \leq (1 + \epsilon) k)] \leq \frac{E [\Psi]}{(1+ \epsilon)F} $ is similar.
\begin{equation}
\frac{(1 + \epsilon)}{e^\epsilon} \leq \bigg( \frac{e^{\epsilon - \frac{(\epsilon^2 - \epsilon^3)}{2}}}{e^\epsilon}\bigg)^{\frac{k}{2}} = e^ {\frac{-k(\epsilon^3 - \epsilon^3)}{4}}
\end{equation}
\begin{equation}
\left. \begin{aligned}
Pr[\overline{(1- \epsilon)k \leq \Psi \leq (1+ \epsilon) k }] \leq \\Pr[ (1- \epsilon)k \geq \Psi  \cup   \Psi \leq (1+ \epsilon) k ]  = \\ 2 e^ {\frac{-k(\epsilon^3 - \epsilon^3)}{4}}
\end{aligned} \right.
\end{equation}

\subsubsection*{Lemma 2 Proof~\cite{Vempala2009}:}
Let $w$ be a random variable of $ w \in \mathbb{R}^{k \times d} $ and $k~<~d $, $x$ is data points $x \in  \mathbb{R}^{d}$ then for any $\epsilon \in [0,1]$~:
\begin{equation} \label{equation_lemma2_1}
\begin{split}
Pr[ (1- \epsilon)||x||^2 \leq || \frac{1}{\sqrt{k}} w x||^2 \leq \\ (1+ \epsilon) ||x||^2 ] \geq   1 - 2 e^ {\frac{-k(\epsilon^3 - \epsilon^3)}{4}}
\end{split}
\end{equation}

In Equation~\eqref{equation_lemma2_1}, $\frac{1}{\sqrt{k}} w x$ is the random approximation value and $ \widehat{x} = w x$, so we can rewrite the Equation~\eqref{equation_lemma2_1} by $Pr[ (1- \epsilon)||x||^2 \leq || \frac{1}{\sqrt{k}}\widehat{x}||^2 \leq (1+ \epsilon) ||x||^2 ] \geq   1 - 2 e^ {\frac{-k(\epsilon^3 - \epsilon^3)}{4}}$.

Call $ \zeta_i = \frac{\widehat{x_i}}{||x||} \sim N(0,1)$ and~~$\Psi= \sum_{i=1}^k \zeta_i^2$ thus:
\begin{equation} \label{equation_lemma2_2}
\begin{split}
Pr[ (1- \epsilon)k \leq || \sum_{i=0}^k\zeta_i ||^2 \leq (1+ \epsilon) k ] =  \\  Pr[ (1- \epsilon)k \leq || w||^2 \leq (1+ \epsilon) k ]
\end{split}
\end{equation}
where we can prove Equation \eqref{equation_lemma2_2} by using Equation \eqref{equation_lemma1_1}:
\begin{equation}
Pr[ (1- \epsilon)k \leq \Psi \leq (1+ \epsilon) k] \geq 1 - 2 e^ {\frac{-k(\epsilon^3 - \epsilon^3)}{4}}
\end{equation}

\subsection{Autoencoder}
An autoencoder is a type of neural network that is trained to attempt to copy its input to its output~\cite{goodfellow2016deep}. The autoencoder has achieved great success as a dimensionality reduction method via the powerful reprehensibility of neural networks~\cite{wang2014generalized}. The first version of autoencoder was introduced by {D.E. Rumelhart et~al.}~\cite{rumelhart1985learning} in 1985. The main idea is that one hidden layer between input and output layers has fewer units~\cite{liang2017text} and could thus be used to reduce the dimensions of a feature space. Especially for texts, documents, and sequences that contain many features, using an autoencoder could help allow for faster, more efficient data processsing.

\subsubsection{General Framework}
\begin{figure}[H]
\centering
\includegraphics[width=0.64\textwidth]{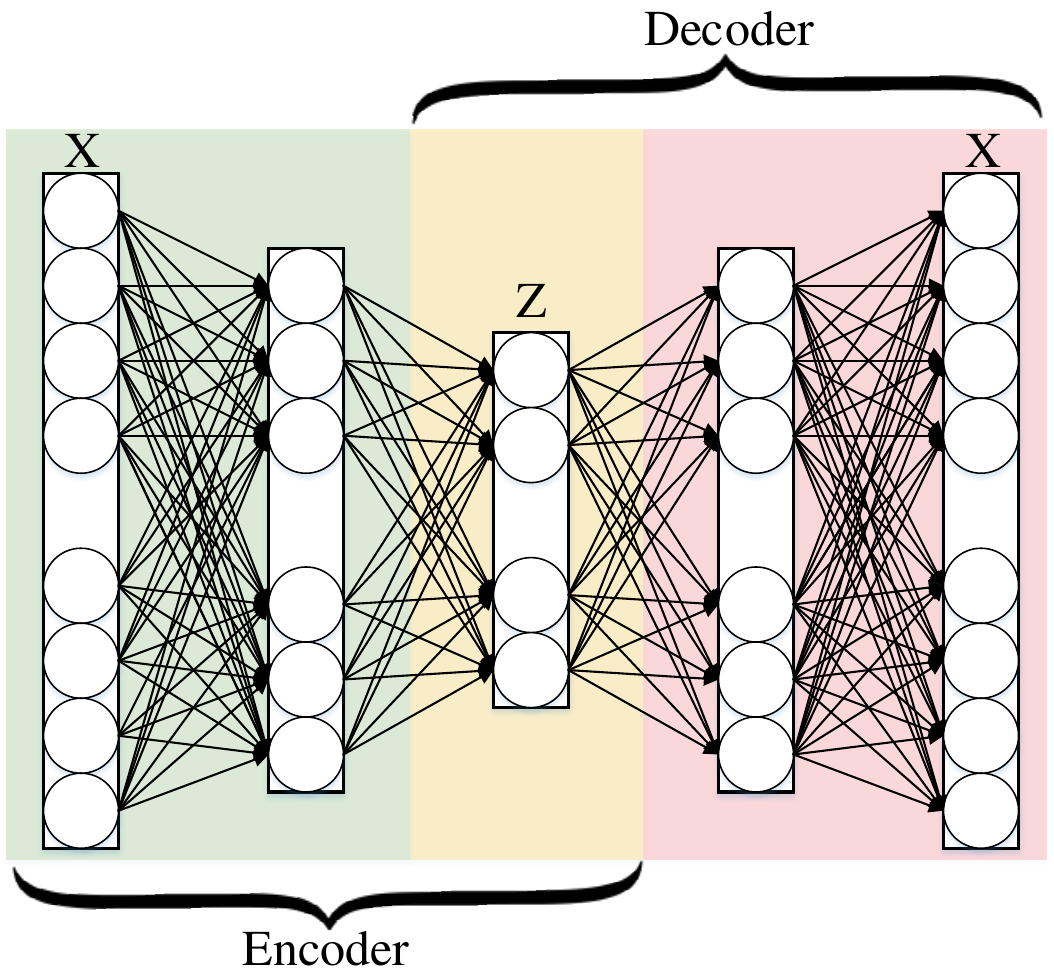}
\caption{This figure shows how a simple autoencoder works. The model depicted contains the following layers: $Z$ is code and two hidden layers are used for encoding and two are used for decoding.} \label{fig:Autoencoder}
\end{figure}

As shown in Figure~\ref{fig:Autoencoder}, the input and output layers of an autoencoder contain~$n$ units where~$x = \mathbb{R}^n$, and hidden layer~$Z$ contains~$p$ units with respect to~$p<n$~\cite{baldi2012autoencoders}. For this technique of dimensionality reduction, the dimensions of the final feature space are reduced from~$n\rightarrow p$. The encoder representation involves a sum of the representation of all words~(for bag-of-words), reflecting the relative frequency of each word~\cite{ap2014autoencoder}:
\begin{equation}
a(x)= c+\sum_{i=1}^{|x|} W_{.,x_i}, \phi(x)=h(a(x))
\end{equation}
where~$h(.)$ is an element-wise non-linearity such as the sigmoid~(Equation~\eqref{eq:sigm}).

\subsubsection{Conventional Autoencoder Architecture}
A convolutional neural networks~(CNN)-based autoencoder can be divided into two main steps~\cite{masci2011stacked}~(encoding and decoding).
\begin{equation}
O_m(i, j) = a\bigg(\sum_{d=1}^{D}\sum_{u=-2k-1}^{2k+1}\sum_{v=-2k -1}^{2k +1} F^{(1)}_{m_d}(u, v)I_d(i -u, j -v)\bigg)~~\quad \forall  m = 1, \cdots, n
\end{equation}
where~$F \in \{F^{(1)}_{1},F^{(1)}_{2},\hdots,F^{(1)}_{n},\}$ which is a convolutional filter, with convolution among an input volume defined by~$I = \left\{I_1,\cdots, I_D\right\}$ which learns to represent input combining non-linear functions:
\begin{equation}
z_m = O_m = a(I * F^{(1)}_{m} + b^{(1)}_m) \quad m = 1, \cdots, m
\end{equation}

where~$b^{(1)}_m$ is the bias, and the number of zeros we want to pad the input with is such that: \text{dim}(I)~=~\text{dim}(\text{decode}(\text{encode}(I))). Finally, the encoding convolution is equal to:
\begin{equation}
\begin{split}
O_w = O_h &= (I_w + 2(2k +1) -2) - (2k + 1) + 1 \\&= I_w + (2k + 1) - 1
\end{split}
\end{equation}

The decoding convolution step produces~$n$ feature maps~$z_{m=1,\hdots,n}$. The reconstructed results~$\hat{I}$ is the result of the convolution between the volume of feature maps~$Z=\{z_{i=1}\}^n$ and this convolutional filters volume~$F^{(2)}$~\cite{chen2015page,geng2015high,masci2011stacked}.
\begin{equation}
\tilde{I} = a(Z * F^{(2)}_{m} + b^{(2)})
\end{equation}
\begin{equation}\label{eq:a:CNN}
\begin{split}
O_w = O_h = &( I_w + (2k + 1) - 1 ) -\\&  (2k + 1) + 1 = I_w = I_h
\end{split}
\end{equation}
where Equation~\eqref{eq:a:CNN} shows the decoding convolution with~$I$ dimensions. Input's dimensions are equal to the output's dimensions.

\subsubsection{Recurrent Autoencoder Architecture}
\begin{figure}[H]
\centering
\includegraphics[width=0.874\textwidth]{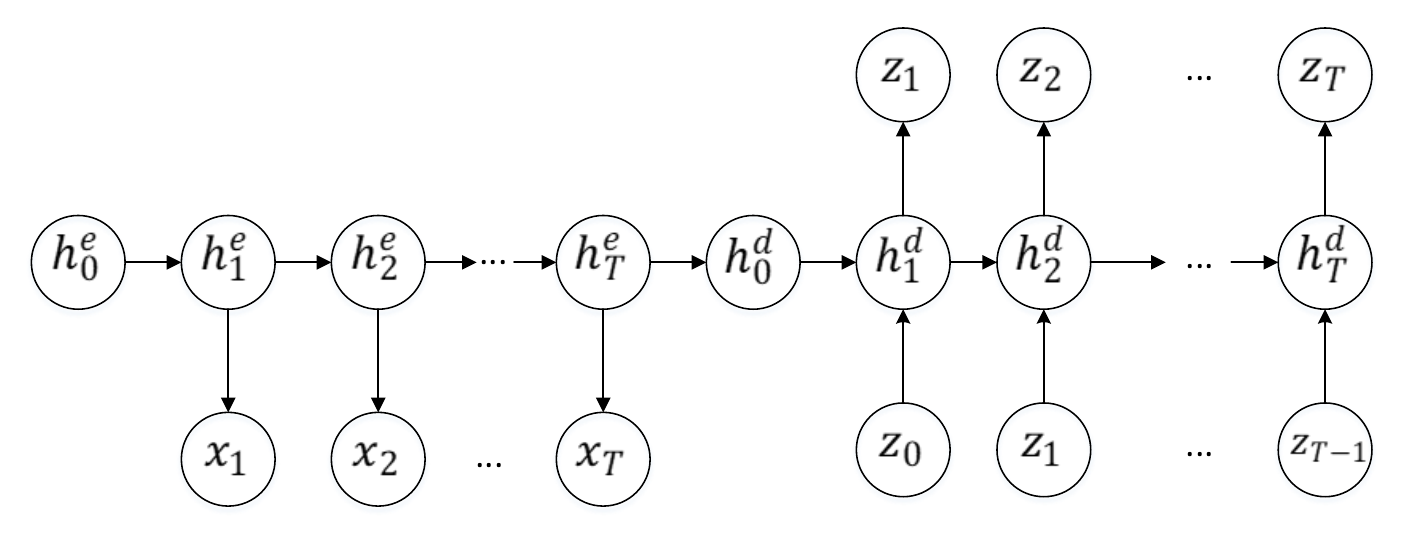}
\caption{A recurrent autoencoder structure.} \label{fig:AERNN}
\end{figure}

A recurrent neural network (RNN) is a natural generalization of feedforward neural networks to sequences~\cite{sutskever2014sequence}. Figure~\ref{fig:AERNN} illustrate recurrent autoencoder architecture.  A standard RNN compute the econding as a sequences of output by iteration:
\begin{align}
h_t &= sigm(W^{hx}x_t +W^{hh}h_{t-1})\\
y_t &= W^{y^h}h_t
\end{align}
where x is inputs $ (x_1, . . . , x_T )$ and $y$ refers to output~($y_1,...,y_T$). A multinomial distribution (1-of-K coding) can be output using a softmax activation function~\cite{cho2014learning}:
\begin{equation}
p(x_{t,j}=1\given x_{t-1,...,x_1}) = \frac{\exp(w_jh_t)}{\sum_{j'=1}^K \exp(w_{j'}h_t)}
\end{equation}

By combining these probabilities, we can compute the probability of the sequence~$x$ as:
\begin{equation}
p(x) = \prod_{t=1}^T p(x_t \given x_{t-1},...,x_1)
\end{equation}

\subsection{T-distributed Stochastic Neighbor Embedding (t-SNE)}
\begin{figure}[H]
\centering
\includegraphics[width=0.44\textwidth]{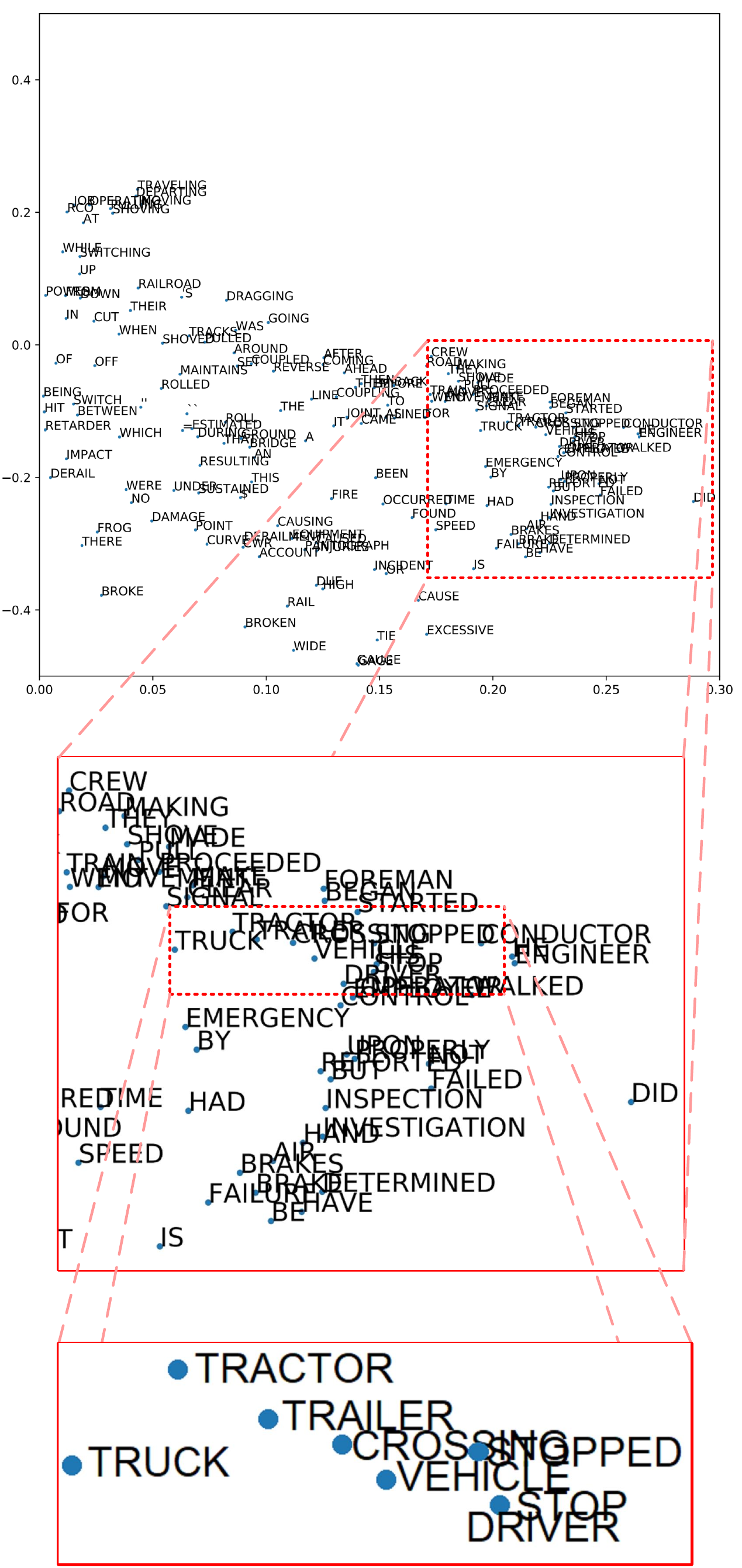}
\caption{This figure presents the t-distributed stochastic neighbor embedding (t-SNE)  visualization of Word2vec of the Federal Railroad Administration (FRA) data set.} \label{fig:T-SNE}
\end{figure}

T-SNE is a nonlinear dimensionality reduction method for embedding high-dimensional data. This method is mostly commonly used for visualization in a low-dimensional feature space~\cite{maaten2008visualizing}, as shown in Figure~\ref{fig:T-SNE}. This approach is based on G. Hinton and S. T. Roweis~\cite{hinton2003stochastic}. SNE works by converting the high dimensional Euclidean distances into conditional probabilities which represent similarities~\cite{maaten2008visualizing}. The conditional probability~$p_{j|i}$ is calculated by:
\begin{equation}
p_{j|i} = \frac{\exp \left(-\frac{||x_i-x_j||^2}{2\sigma_i^2}\right)}{\sum_{k\neq i}\exp \left(-\frac{||x_i-x_k||^2}{2\sigma_i^2}\right)}
\end{equation}
where $\sigma_i$ is the variance of the centered on data point~$x_i$. The similarity of $y_j$ to~$y_i$ is calculated as~follows:
\begin{equation}
q_{j|i} = \frac{\exp \left(-||y_i-y_j||^2\right)}{\sum_{k\neq i}\exp \left(-||y_i-y_k||^2\right)}
\end{equation}

The cost function~$C$ is as follows:
\begin{equation}
C= \sum_i KL(p_i|Q_i)
\end{equation}
where $KL(P_i|Q_i)$ is the Kullback--Leibler divergence~\cite{joyce2011kullback}, which is calculated as:
\begin{equation}
KL(P_i|Q_i) = \sum_j p_{j|i} \log\frac{p_{j|i}}{q_{j|i}}
\end{equation}

The gradient update with a momentum term is as follows:
\begin{equation}
\gamma^{(t)} = \gamma^{(t-1)}+ \eta \frac{\delta C}{\delta \gamma} +\alpha(t) \left(\gamma^{(t-1) - \gamma^{(t-2)}}\right)
\end{equation}
where~$\eta$ is the learning rate,~$\gamma^{(t)}$ refers to the solution at iteration \textit{t},
and~$\alpha(t)$ indicates momentum at iteration \textit{t}. Now we can re-write symmetric SNE in the high-dimensional space and a joint probability
distribution,~$Q$, in the low-dimensional space as follows~\cite{maaten2008visualizing}:
\begin{equation}
C= KL(P||Q) = \sum_i \sum_j p_{ij} \log\frac{p_{ij}}{q_{ij}}
\end{equation}
in the high-dimensional space~$p_{ij}$  is:
\begin{equation}
p_{ij} = \frac{\exp \left(-\frac{||x_i-x_j||^2}{2\sigma^2}\right)}{\sum_{k\neq l}\exp \left(-\frac{||x_i-x_l||^2}{2\sigma^2}\right)}
\end{equation}

The gradient of symmetric S is as follows:
\begin{equation}
\frac{\delta C}{\delta y_i} = 4\sum_j (p_{ij}-q_{ij})(y_i-y_j)
\end{equation}

\section{Existing Classification Techniques}\label{sec:techniques}
In this section, we outline existing text and document classification algorithms. First, we describe the Rocchio algorithm which is used for text classification. Then, we address two popular techniques in ensemble learning algorithms: Boosting and bagging. Some methods, such as logistic regression, Na\"ive Bayes, and k-nearest neighbor, are more traditional but still commonly used in the scientific community. Support vector machines (SVMs), especially kernel SVMs, are also broadly used as a classification technique. Tree-based classification algorithms, such as decision tree and random forests are fast and accurate for document categorization. We also describe neural network based algorithms such as deep neural networks~(DNN), CNN, RNN, deep belief network~(DBN), hierarchical attention networks~(HAN), and combination techniques.

\subsection{Rocchio Classification}
The Rocchio algorithm was first introduced by~{J.J. Rocchio}~\cite{rocchio1971relevance} in 1971 as method of using relevance feedback to query full-text databases. Since then, many researchers have addressed and developed this technique for text and document classification~\cite{partalas2015lshtc,sowmya2016large}. This classification algorithm uses TF-IDF weights for each informative word instead of boolean features. Using a training set of documents, the Rocchio algorithm builds a prototype vector for each class. This prototype is an average vector over the training documents' vectors that belong to a certain class. It then assigns each test document to the class with the maximum similarity between the test document and each of the prototype vectors~\cite{korde2012text}. The average vector computes the centroid of a class \emph{c}~(center of mass of its~members):
\begin{equation}
\Vec{\mu}(c)=\frac{1}{|D_c|}\sum_{d \in D_c}\Vec{v}_d
\end{equation}
where $D_c$ is the set of documents in \emph{D} that belongs to class \emph{c} and $\Vec{v}_d$ is the weighted vector representation of document \emph{d}. The predicted label of document \emph{d} is the one with the smallest Euclidean distance between the document and the centroid:
\begin{equation}
c^*=\argmin_c\lVert \Vec{\mu}_c-\Vec{v}_d\rVert
\end{equation}

Centroids can be normalized to unit-length as follows:
\begin{equation}
\Vec{\mu}_c=\frac{\sum_{d \in D_c}\Vec{v}_d}{\lVert\sum_{d\in D_c}\Vec{v}_d\rVert}
\end{equation}

Therefore, the label of test documents can be obtained as follows:
\begin{equation}
c_*=\argmin_c \Vec{\mu}_c\cdot\Vec{v}_d
\end{equation}

\subsubsection*{Limitation of Rocchio Algorithm}
The Rocchio algorithm for text classifcation contains many limitations such as the fact that the user can only retrieve a few relevant documents using this model~\cite{selvi2017text}. Furthermore, this algorithms' results illustrate by taking semantics into consideration~\cite{albitar2012towards}.

\subsection{Boosting and Bagging}
Voting classification techniques, such as bagging and boosting, have been successfully developed for document and text data set classification~\cite{farzi2016estimation}. While boosting adaptively changes the distribution of the training set based on the performance of previous classifiers, bagging does not look at the previous classifier~\cite{bauer1999empirical}.

\SetInd{0.25em}{0.1em}

\subsubsection{Boosting}
The boosting algorithm was first introduced by R.E. Schapire~\cite{schapire1990strength} in 1990 as a technique for boosting the performance of a weak learning algorithm. This technique was further developed by Freund~\cite{freund1992improved,bloehdorn2004boosting}.

\begin{figure}[H]
\centering
\includegraphics[width=\textwidth]{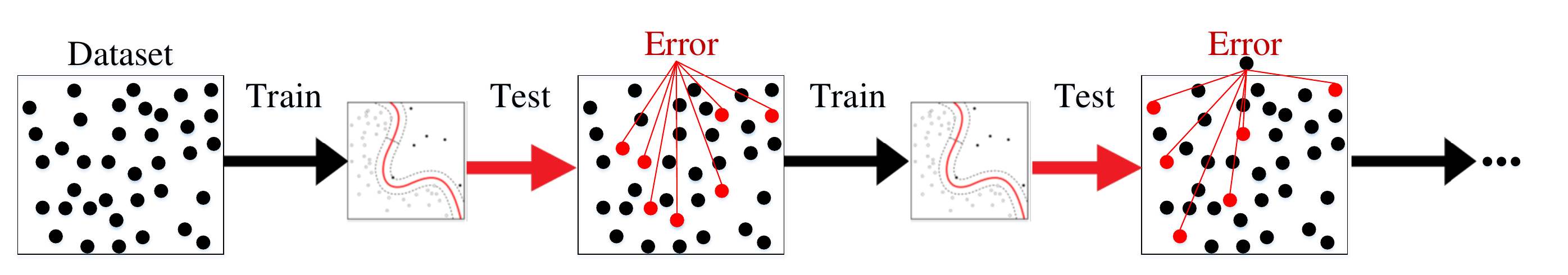}
\caption{This figure is the boosting technique architecture.} \label{fig:Boosting}
\end{figure}

Figure~\ref{fig:Boosting} shows how a boosting algorithm works for 2D data sets, as shown we have labeled the data, then trained by multi-model architectures~(ensemble learning). These developments resulted in the AdaBoost (Adaptive Boosting)~\cite{freund1995efficient}. Suppose we construct~$D_t$ such that $D_1(i)= \frac{1}{m}$ given~$D_t$ and~$h_t$:
\begin{equation}
\begin{split}
D_{t+1}(i) &= \frac{D_t(i)}{Z_t}\times
\begin{cases}
e^{-\alpha_t}& if y_i = h_t(x_i)\\
e^{\alpha_t}& if y_i \neq h_t(x_i)
\end{cases}\\
&=\frac{D_t(i)}{Z_t} exp(-\alpha y_i h_i(x_i))
\end{split}
\end{equation}
where $Z_t$ refers to the normalization factor and~$\alpha_t$ is as follows:
\begin{equation}
\alpha_t = \frac{1}{2}\ln \bigg(\frac{1-\epsilon_t}{\epsilon_t}\bigg)\\\vspace{-6pt}
\end{equation}

As shown in Algorithm~\ref{alg:AdaBoost_Method},  training set~S of size $m$, inducer~$\tau$ and integer $N$ as input. Then this algorithm find the weights of each~$x_j$, and finally, the output is the optimal classifier~(\emph{$C^*$}).

\begin{algorithm}[H]
\caption{The AdaBoost method}\label{alg:AdaBoost_Method}
\SetKwInOut{Input}{input}\SetKwInOut{Output}{output}
\SetKwFor{Foreach}{for}{do}{endfor}
\SetKwIF{If}{ElseIf}{Else}{if}{then}{else if}{else}{endif}
\BlankLine
\Input{training set~S of size $m$, inducer~$\tau$, integer $N$}
\BlankLine
\Foreach { $i=1$ to $N$}{

~~~~$C_i=\tau(S')$

\vspace{-10pt}
\begin{flalign}\nonumber
~~~&\epsilon_i = \frac{1}{m} \sum_{x_j\in S'; C_i(x_j)\notin y_i} weight(x)&&
\end{flalign}
\vspace{-10pt}

\If{$\epsilon_i > \frac{1}{2}$}{

~~~~~~~~set S' to a bootstrap sample from S with weight 1 for

~~~~~~~~all instance and go top

~~~~}

~~~~$\beta_i = \frac{\epsilon_i}{1-\epsilon_i}$

\Foreach { $x_i\in S'$}{

\If{$C_i(x_j)=y_i$}{

~~~~~~~~$weight(x_j) = weight(x_j) . \beta_i$

}

}

~~~~Normalize weights of instances

}
\vspace{-10pt}
\begin{flalign}\nonumber
&C^*(x)= arg\max_{y\in Y} \sum_{i,C_i(x)=y} \log \frac{1}{\beta_i}&&
\end{flalign}
\vspace{-10pt}

\Output{~Classifier $C^*$}
\end{algorithm}

The final classifier formulation can be written as:
\begin{equation}
H_f(x)=sign\bigg(\sum_t\alpha_th_t(x)\bigg)
\end{equation}

\subsubsection{Bagging}
\begin{figure}[H]
\centering
\includegraphics[width=0.5\textwidth]{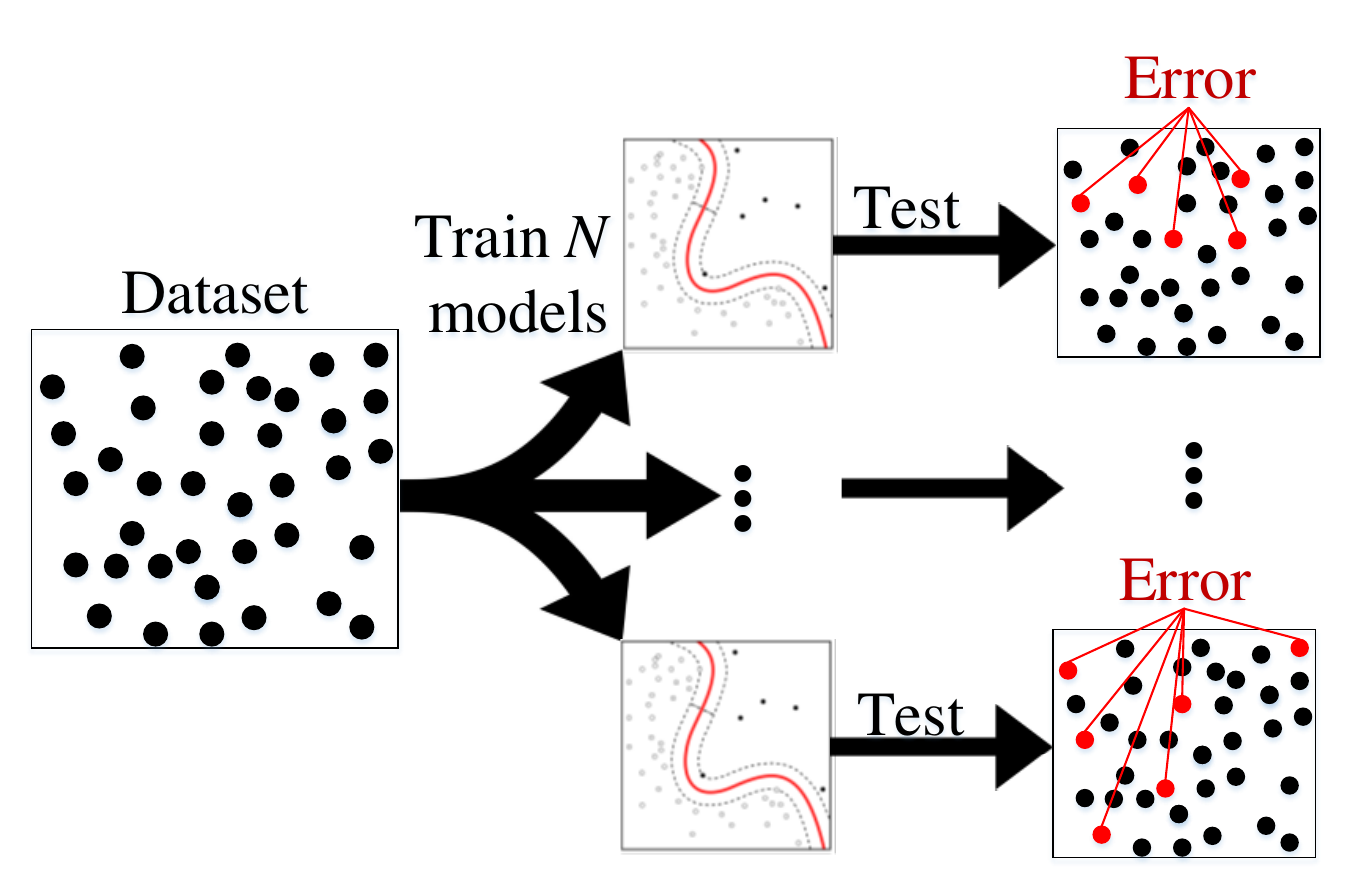}
\caption{This figure shows a simple model of the bagging technique.} \label{fig:Bagging}
\end{figure}

The bagging algorithm was introduced by~{L. Breiman}~\cite{breiman1996bagging} in 1996 as a voting classifier method. The algorithm is generated by different bootstrap samples~\cite{bauer1999empirical}. A bootstrap generates a uniform sample from the training set. If~$N$ bootstrap samples $B_1, B_2,\hdots, B_N$ have been generated, then we have $N$ classifiers~($C$)  which $C_i$ is built from each bootstrap sample $B_i$. Finally, our classifier~$C$ contain or generated from $C_1,C_2,...,C_N$ whose output is the class predicted most often by its sub-classifiers, with ties broken arbitrarily~\cite{bauer1999empirical,breiman1996bagging}.
Figure~\ref{fig:Bagging} shows a simple bagging algorithm which trained~$N$ models. As shown in Algorithm~\ref{alg:bagging}, We have training set~\emph{S} which is trained and find the best classifier~\emph{C}.

\unskip
\begin{algorithm}[H]
\SetKwInOut{Input}{input}\SetKwInOut{Output}{output}
\SetKwFor{Foreach}{for}{do}{endfor}
\SetKwIF{If}{ElseIf}{Else}{if}{then}{else if}{else}{endif}
\BlankLine
\Input{training set~S, inducer~$\tau$, integer $N$}
\BlankLine
\Foreach {$i=1$ to $N$}{
~~~~$S'$ = bootstrap sample from S

~~~~$C_i$ = $\tau(S')$
}
\vspace{-10pt}
\begin{flalign}\nonumber
&C^*(x)= arg\max_{y\in Y} \sum_{i,C_i=y} 1&&
\end{flalign}
\vspace{-10pt}

\Output{~Classifier $C^*$}
\caption{Bagging}\label{alg:bagging} 
\end{algorithm}

\subsubsection{Limitation of Boosting and Bagging}

Boosting and bagging methods also have many limitations and disadvantages, such as  the computational complexity and loss of interpretability~\cite{geurts2000some}, which means that the feature importance could not be discovered by these models.

\subsection{Logistic Regression}
One of the earliest methods of classification is logistic regression~(LR). LR was introduced and developed by statistician David Cox in 1958~\cite{cox2018analysis}. LR is a linear classifier with decision boundary of $\theta^Tx=0$. LR predicts probabilities rather than classes~\cite{fan2008liblinear,genkin2007large}.

\subsubsection{Basic Framework}
The goal of LR is to train from the probability of variable~$Y$ being 0 or 1 given $x$. Let us have text data which is $X \in \mathbb{R}^{n\times d}$. If we have binary classification problems, the Bernoulli mixture models function should be used~\cite{juan2002use} as follows:
\begin{equation}
\begin{split}
L(\theta \given x )= &~p(y\given x;\theta) =\\& \prod_{i=1}^n \beta (y_i \given sigm(x_i\theta)) =\\& \prod_{i=1}^n sigm  (x_i)^{y_i}(1-sigm (x_i))^{1-y_i} =\\
& \prod_{i=1}^n \bigg[ \frac{1}{1+e^{-x_i\theta}}\bigg]^{y_i}\bigg[ 1-  \frac{1}{1+e^{-x_i\theta}}\bigg]^{(1-y_i)}
\end{split}
\end{equation}
where $x_i\theta = \theta_0+\sum_{j=1}^d (x_{ij}\theta_j)$, and $sigm(.)$ is a sigmoid function which is defined as shown in Equation~\eqref{eq:sigm}.
\begin{equation}\label{eq:sigm}
sigm(\eta) = \frac{1}{1-e^{-\eta}} = \frac{e^\eta}{1-e^{\eta}}
\end{equation}

\subsubsection{Combining Instance-Based Learning and LR}
The LR model specifies the probability of binary output $y_i=\{0,1\}$ given the input $x_i$. we can consider posterior probability as:
\begin{equation}
\pi_0 =P(y_0 = +1\given y_i)
\end{equation}
where:
\begin{equation}
\frac{\pi_0}{1-\pi_0} = \frac{P(y_i\given y_0 = +1)}{P(y_i\given y_0 = +1)}. \frac{p_0}{1-p_0}
\end{equation}
where~$p$ is the likelihood ratio it could be re-written as:
\begin{align}
\frac{\pi_0}{1-\pi_0} &= p. \frac{p_0}{1-p_0}\\
\log\bigg(\frac{\pi_0}{1-\pi_0}\bigg) &= \log(p)+ w_0
\end{align}
with respect to:
\begin{equation}
w_0=\log(p_0)-\log(1-p_0)
\end{equation}

To obey the basic principle underlying instance-based learning~(IBL)~\cite{cheng2009combining}, the classifier should be a function of the distance~$\delta_i$.
$p$ will be large if~$\delta_i\rightarrow 0$ then~$y_i=+1$, and  small for $y_i=-1$. $p$ should be close to~$1$ if~$\delta_i\rightarrow \infty$; then, neither in favor of~$y_0 = +1$ nor in favor of~$y_0 = -1$, so the parameterized function is as follows:
\begin{equation}
p = p(\delta) = \exp\big(y_i.\frac{\alpha}{\delta}\big)
\end{equation}

Finally,
\begin{equation}
\log\bigg(\frac{\pi_0}{1-\pi_0}\bigg) = w_0+\alpha \sum_{x_i\in N(x_0)} k(x_0,x_i).y_i
\end{equation}
where~$k(x_0,x_i)$ is similarity measure.

\subsubsection{Multinomial Logistic Regression}
Multinomial (or multilabeled) logistic classification~\cite{krishnapuram2005sparse} uses the probability of~$x$ belonging to class~$i$ (as defined in Equation~\eqref{eq:probx})
\begin{equation}\label{eq:probx}
p\left(y^{(i)} = 1 \given x,\theta\right) = \frac{\exp\left({\theta^{(i)}}^Tx\right)}{\sum_{j=1}^m \exp\left({\theta ^{(i)}}^Tx\right)}
\end{equation}
where $\theta^{(i)}$ is the weight vector corresponding to class~$i$.

For binary classification~($m=2$) which is known as a basic LR, but for multinomial logistic regression~($m>2$) is usually uses the~$softmax$ function.

The normalization function is:
\begin{equation}
\sum_{i=1}^m  p\left(y^{(i)} = 1 \given x,\theta\right) = 1
\end{equation}

In a classification task as supervised learning context, the component of~$\theta$ is calculated from the subset of the training data~$D$ which belongs to class~$i$ where~$i \in\{1,\hdots,n\}$. To perform maximum likelihood~(ML) estimation of~$\theta$, we need to maximize the log-likelihood function as follows:
\begin{equation}
\begin{split}
\ell(\theta) &= \sum_{j=1}^n \log  p\left(y_j = 1 \given x_j,\theta\right) \\
&=\sum_{j=1}^n \left[
\sum_{i=1}^m y^{(i)}_j {\theta^{(i)}}^Tx_j -\log \sum_{i=1}^m \exp\left({\theta^{(i)}}^Tx_j \right)
\right]
\end{split}
\end{equation}

The adoption of a maximum a posteriori~(MAP) estimates as follows:
\begin{equation}
\Hat{\theta}_{MAP} = arg\max_{\theta} L(\theta) =  arg\max_{\theta}\left[ \ell(\theta) + \log p(\theta)\right]
\end{equation}

\subsubsection{Limitation of Logistic Regression}
Logistic regression classifier works well for predicting categorical outcomes.~However, this prediction requires that each data point be independent~\cite{huang2015unconstrained} which is attempting to predict outcomes based on a set of independent variables~\cite{guerin2016using}

\subsection{Na\"ive Bayes Classifier}
Na\"{i}ve Bayes text classification has been widely used for document categorization tasks since the 1950s~\cite{kaufmann1969cuba,porter1980algorithm}. The Na\"{i}ve Bayes classifier method is theoretically based on Bayes theorem, which was formulated by Thomas Bayes between 1701--1761~\cite{pearson1925bayes,hill1968posterior}). Recent studies have widely addressed this technique in information retrieval~\cite{qu2018improved}.
This technique is a generative model, which is the most traditional method of text categorization. We start with the most basic version of NBC which was developed by using TF~(bag-of-word), a feature extraction technique which counts the number of words in documents.

\subsubsection{High-Level Description of Na\"ive Bayes Classifier}
If the number of documents~($n$) fit into $k$ categories where $k \in \{ c_1,c_2,...,c_k\}$, the predicted class as output is $c \in C$. The Na\"ive Bayes algorithm can be described as follows:
\begin{equation}
P(c \given d) = \frac{P(d \given c)P(c)}{P(d)}
\end{equation}
where $d$ is document and $c$ indicates classes.
\begin{equation} \label{eq1}
\begin{split}
C_{MAP} &=~  arg \max_{c \in C} P(d \given c)P(c) \\&= arg \max_{c\in C} P(x_1,x_2,...,x_n \given c) p(c)
\end{split}
\end{equation}

This model is used as baseline of many papers which is word-level of Na\"ive Bayes classifier~\cite{kim2006some,mccallum1998comparison} as follows:
\begin{equation}
P(c_j \given d_i;\hat{\theta}) = \frac{P(c_j \given \hat{\theta})P(d_i \given c_j ; \hat{\theta_j})}{P(d_i \given \hat{\theta})}
\end{equation}

\subsubsection{Multinomial Na\"ive Bayes Classifier}

If the number of documents~($n$) fit into $k$ categories where $k \in \{ c_1,c_2,...,c_k\}$ the predicted class as output is $c \in C$. The Na\"ive Bayes algorithm can be written as:
\begin{equation}
P(c \given d) = \frac{P(c) \prod_{w\in d} P(d \given c)^{n_{wd}}}{P(d)}
\end{equation}
where $n_{wd}$ is denoted to the number of times word~$w$ occurs in document, and~$P(w|c)$ is the probability of observing word~$w$ given class~$c$~\cite{frank2006naive}.

$P(w|c)$ is calculated as:
\begin{equation}\label{eq:MNBC}
P(w \given c) = \frac{1+\sum_{d \in D_c}n_{wd}}{k+\sum_{w'}\sum_{d \in D_c} n_{w'd}}
\end{equation}

\subsubsection{Na\"ive Bayes Classifier for Unbalanced Classes}
One of the limitations of NBC is that the technique performs poorly on data sets with unbalanced classes~\cite{liu2009imbalanced}. 
Eibe Frank and Remco R. Bouckaert~\cite{frank2006naive} developed a method for introducing normalization in each class by Equation~\eqref{eq:NBC_Normalization} and then uses the centroid classifier~\cite{han2000centroid} in NBC for unbalanced classes.
The centroid~$c_c$ for class~$c$ is given in Equation~\eqref{eq:NBC_Centroid}.
\begin{equation}\label{eq:NBC_Normalization}
\alpha \times \frac{n_{wd}}{\sum_{w'}\sum_{d \in D_c}n_{w'd}}
\end{equation}
\begin{equation}\label{eq:NBC_Centroid}
\begin{split}
c_c = \bigg\{& \frac{\sum_{d \in D_c}n_{w_1d}}{\sqrt{\sum_{w} \big( \sum_{d \in D_c} n_{wd}\big)^2} },\hdots ,\\& \frac{\sum_{d \in D_c}n_{w_id}}{\sqrt{\sum_{w} \big( \sum_{d \in D_c} n_{wd}\big)^2} }, \hdots,\\& \frac{\sum_{d \in D_c}n_{w_kd}}{\sqrt{\sum_{w} \big( \sum_{d \in D_c} n_{wd}\big)^2} }\bigg\}
\end{split}
\end{equation}

The scoring function is defined as:
\begin{equation}
x_d.c_1-x_d.c_2
\end{equation}

So log of multinomial Na\"ive Bayes classifier can be calculated as:
\begin{equation}
\left[ \log P(c_1)+\sum_{i=1}^k n_{w_id} \log(P(w_i \given c_1)) \right]-
\left[ \log P(c_2)+\sum_{i=1}^k n_{w_id} \log(P(w_i \given c_2)) \right]
\end{equation}

Using Equations~\eqref{eq:MNBC} and~\eqref{eq:NBC_Normalization}, and if~$\alpha=1$ we can rewrite:
\begin{equation}
P(w \given c) = \frac{1+ \frac{n_{wd}}{\sum_{w'}\sum_{d \in D_c}n_{w'd}}}{K+1}
\end{equation}
with respect to:
\begin{equation}
\frac{\sum_{d \in D_c}n_{wd}}{\sum_{w'}\sum_{d \in D_c} n_{w'd}} <<1
\end{equation}

For text data sets and $log(x+1)\approx x$ and $x <<1$~\cite{frank2006naive}. In this technique of NBC, the experimental results is very similar to the centroid classifier~\cite{han2000centroid}.

\subsubsection{Limitation of Na\"ive Bayes Algorithm}
Na\"ive Bayes algorithm also has several limitations. NBC makes a strong assumption about the shape of the data distribution~\cite{soheily2017intrusion,wang2012nonparametric}. NBC is also limited by data scarcity for which any possible value in feature space, a likelihood value must be estimated by a frequentist~\cite{ranjan2017document}.

\subsection{K-Nearest Neighbor}
The k-nearest Neighbors algorithm (KNN) is a non-parametric technique used for classification. This method is used for text classification applications in many research domains~\cite{jiang2012improved} in past decades.

\subsubsection{Basic Concept of KNN}
\begin{figure}[H]
\centering
\includegraphics[width=0.5\textwidth]{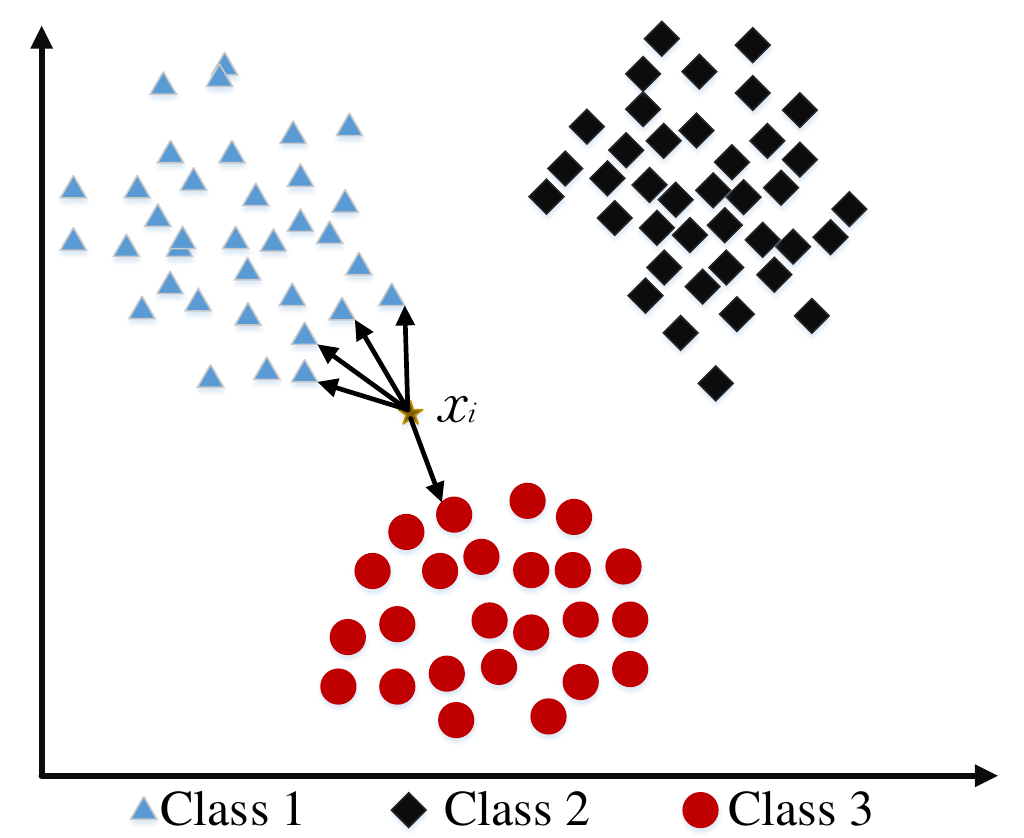}
\caption{A architecture of k-nearest Neighbor (KNN) model for the 2D data set and three classes.} \label{fig:KNN}
\end{figure}

Given a test document~$x$, the KNN algorithm finds the~$k$ nearest neighbors of~$x$ among all the documents in the training set, and scores the category candidates based the class of~$k$ neighbors. The~similarity of~$x$ and each neighbor's document could be the score of the category of the neighbor documents. Multiple KNN documents may belong to the same category; in this case, the summation of these scores would be the similarity score of class~$k$ with respect to the test document~$x$. After sorting the score values, the algorithm assigns the candidate to the class with the highest score from the test document~$x$~\cite{jiang2012improved}. Figure~\ref{fig:KNN} illustrates KNN architecture, but for simplicity, this figure is designed by a 2D data set~(similar and with higher dimensional space like the text data set). The decision rule of KNN is:
\begin{equation}
\begin{split}
f(x) = &arg \max_j S(x,C_j) \\ = &\sum_{d_i\in KNN} sim(x,d_i)y(d_i,C_j)
\end{split}
\end{equation}
where S refers to score value with respect to $S(x,C_j)$, the score value of candidate~$i$ to class of~$j$, and output of $f(x)$ is a label to the test set document.

\subsubsection{Weight Adjusted K-Nearest Neighbor Classification}
The weight adjusted k-nearest neighbor classification (WAKNN) is a version of KNN which tries to learn the weight vectors for classification~\cite{han2001text}. The weighted cosine measure~\cite{salton1989automatic} is calculated as~follows:
\begin{equation}
\begin{split}
\cos(x,y,w) \dfrac{\sum\limits_{t\in T} (x_t\times w_t)\times(y_t\times w_t)}{\sqrt{\sum\limits_{t\in T} (x_t\times w_t)^2} \times \sqrt{\sum\limits_{t\in T} (y_t\times w_t)^2}}
\end{split}
\end{equation}
where~$T$ refers to the set of words, and~$x_t$ and~$y_t$ are TF, as discussed in Section~\ref{sec:Feature_extraction}.  For the training model~($d \in D$), let $N_d = \{n_1, n_2,\hdots,n_k\}$ be the set of k-nearest Neighbors of~$d$. Given~$N_d$, the similarity sum of~$d$ neighbors that belong to class~$c$ is defined as follows:
\begin{equation}
S_c = \sum_{n_i \in N; C(n_i)=c} \cos(d,n_i,w)
\end{equation}

Total similarity is calculated as follows:
\begin{equation}
T = \sum_{c \in C }S_c
\end{equation}

The contribution of $d$ is defined in terms of~$S_c$ of classes~$c$ and~$T$ as follows:
\begin{equation}
cont(d) =
\begin{cases}
1       & \quad \stackunder{\text{if } $\forall c \in C, c \neq class(d),$\quad\quad}{$S_{class(d)>S_s}$ and $\frac{S_{class(d)}}{T}\leq p $}\\
0  & \quad \text{ otherwise}
\end{cases}
\end{equation}
where~$cont(d)$ stands for~$contribution(d)$

\subsubsection{Limitation of K-Nearest Neighbor}
KNN is a classification method that is easy to implement and adapts to any kind of feature space. This model also naturally handles multi-class cases~\cite{sahgal2015road,patel2014ant}. However, KNN is limited by data storage constraints for large search problems to find nearest neighbors. 
Additionally, the performance of KNN is dependent on finding a meaningful distance function, thus making this technique a very data dependent algorithm~\cite{sahgal2014object,sanjay4comparing}.

\subsection{Support Vector Machine~(SVM)}\label{me_svm}
\begin{figure}[H]
\centering
\includegraphics[width=0.8\textwidth]{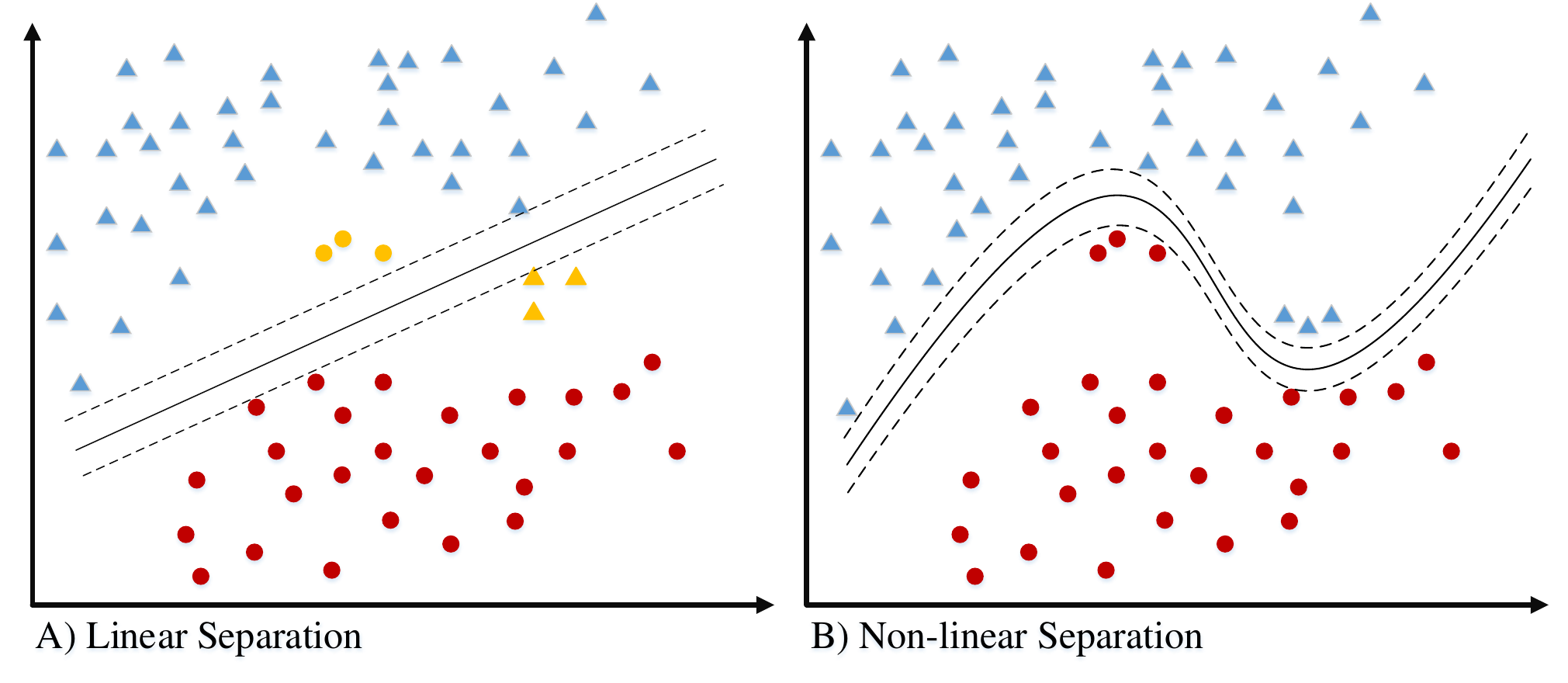}
\caption{This figure shows 
the linear and non-linear Support Vector Machine (SVM) for a 2D data set~(for text data we have thousands of dimensions). The red is class~$1$, the blue color is class~$2$ and yellow color is miss-classified data points.} 
\label{fig:SVM}
\end{figure}

The original version of SVM was developed by {Vapnik and~Chervonenkis}~\cite{vapnik1964class} in 1963. {B.E.~Boser~et~al.}~\cite{boser1992training} adapted this version into a nonlinear formulation in the early 1990s. SVM~was originally designed for binary classification tasks. However, many researchers work on multi-class problems using this dominate technique~\cite{bo2006svm}. The Figure~\ref{fig:SVM} indicates the linear and non-linear classifier which is used for $2-dimension$ datasets.

\subsubsection{Binary-Class SVM}
In the context of text classification, let $x_1, x_2,...,x_l$ be training examples belonging to one class $X$, where $X$ is a compact subset of $R^N$~\cite{manevitz2001one}. Then we can formulate a binary classifier as follows:
\begin{equation}
\min \frac{1}{2}||w||^2+ \frac{1}{vl}\sum_{i=1}^l \xi_i - p
\end{equation}
subject to:
\begin{equation}
(w. \Phi(x_i)) \geq p-\xi_i~~i=1,2,\hdots,l~~ \xi \geq 0
\end{equation}

If $w$ and $p$ solve this problem, then the decision function is given by:
\begin{equation}
f(x) = sign((w.\Phi(x))-p)
\end{equation}

\subsubsection{Multi-Class SVM}
Since SVMs are traditionally used for the binary classification, we need to generate a Multiple-SVM~(MSVM)~\cite{mohri2012foundations} for multi-class problems. One-vs-One is a technique for multi-class SVM that builds $N(N-1)$ 
classifiers as follows:
\begin{equation}
f(x) = arg \max_{i}\big(\sum_j f_{ij}(x)\big)
\end{equation}

The natural way to solve the k-class problem is to construct a decision function of all $k$ classes at once~\cite{chen2016turning,weston1998multi}. In general, multi-class SVM is an optimization problem of the following form:
\begin{align}
&\min_{w_1,w_2,..,w_k,\zeta}~\frac{1}{2}\sum_{k}w_k^Tw_k + C\sum_{(x_i,y_i) \in D} \zeta_i\\
\begin{split}
st.~~ &w_{y_i}^Tx - w_k^Tx \leq i - \zeta_i,~~\\
&\forall (x_i,y_i) \in D, k\in\{1,2,...,K\}, k\neq y_i
\end{split}
\end{align}
where $(x_i,y_i)$ represent the training data points such that $(x_i,y_i) \in D$, $C$ is the penalty parameter, $\zeta$ is a slack parameter, and $k$ stands for the class.

Another technique of multi-class classification using SVM is All-vs-One. Feature extraction via SVM generally uses one of two methods: Word sequences feature extracting~\cite{zhang2008text} and TF-IDF. But for an unstructured sequence such as RNA and DNA sequences, string kernel is used. However, string kernel can be used for a document categorization~\cite{lodhi2002text}.

\subsubsection{String Kernel}\label{String_kernel}
Text classification has also been studied using string kernel~\cite{lodhi2002text}. The basic idea of string kernel~(SK) is using $\Phi(.)$ to map the string in the feature space. 

Spectrum kernel  as part of SK 
has been applied to many different applications, including text, DNA, and protein classification~\cite{leslie2002spectrum,eskin2002mismatch}. The basic idea of Spectrum kernel is counting the number of times a word appears in string $x_i$ as a feature map where defining feature maps from $x \rightarrow R^{l^k}$.
\begin{align}
\Phi_k (x)& = \Phi_j(x)_{j \in \Sigma^k} \\
where~~~~ &\nonumber \\
\Phi_j (x)& = number~ of~ j~~feature~appears~ in~x
\end{align}

The feature map $\Phi_i(x)$ is generated by the sequence $x_i$ and kernel defines as follows:
\begin{align}\label{feature_mis}
F = &\Sigma ^ k  \\
\label{SP_kernel}
K_i(x,x') = &<\Phi_i(x),\Phi_i(x')>
\end{align}

The main limitation of SVM when applied to string sequence classification is time complexity~\cite{singh2017gakco}. The features are generated using dictionary size $\Sigma$ and $F$ is the number of features and bounded by Equation \eqref{feature_mis}. The kernel calculation is similar with SP and uses Equation~\eqref{SP_kernel}, and finally normalizes the kernel using Equation~\eqref{miskernel}.
\begin{align}
K^{Norm}(x,y) & \leftarrow~ \frac{K(x,y)}{\sqrt{K(x,x)}~ \sqrt{K(y,y)} }
\label{miskernel}\\
<f^x , f^y > &= \sum_{i=1}^{n_1}  \sum_{j=1}^{n_2} h(u_i^{s_1} , u_j^{s_2} )  \label{miskernel_2}
\end{align}
where two sequences,~$u_i^{s_1}$ and $u_j^{s_2}$, are lengths of $s_1$ and $s_2$ respectively.

\subsubsection{Stacking Support Vector Machine~(SVM)}
\begin{figure}[H]
\centering
\includegraphics[width=0.65\textwidth]{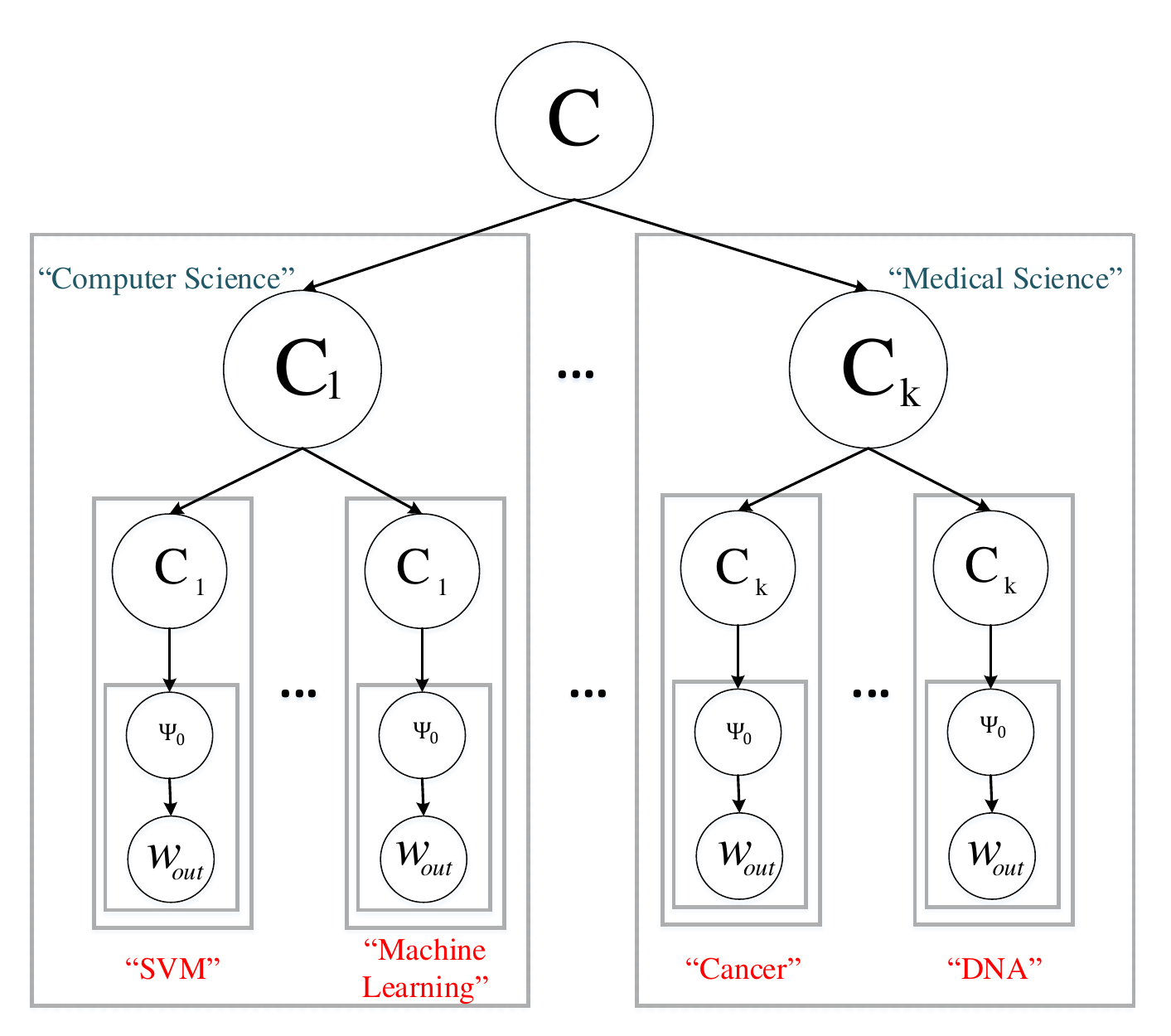}
\caption{Hierarchical classification  method.}  \label{Fig:HSVM_HDLTex}
\end{figure}

Stacking SVM is a hierarchical classification method used for category tree structure based on a top-down level-based approach~\cite{sun2001hierarchical}. This technique provides a hierarchical model of individual SVM classifiers, and thus generally produces more accurate results than single-SVM models~\cite{sebastiani2002machine}. As shown in the Figure~\ref{Fig:HSVM_HDLTex}, the stacking model employs hierarchical classifier which contains several layers~(in this Figure we have two level like mane domain, and sub-domains).

\subsubsection{Multiple Instance Learning~(MIL)}
Multiple instance learning~(MIL) is a supervised learning method~\cite{maron1998framework} and is typically formulated as one of two SVM-based methods~(mi-SVM and MI-SVM)~\cite{andrews2003support}. MIL takes in a set of labeled bags as input instead of instances. A bag is labeled positive if there is at least one instance in it with a positive label, and labeled negative if all instances of it are negative. Then, the learner tries to infer a concept that label individual instances correctly~\cite{maron1998framework}. In statistical pattern recognition, it is assumed that a training set of labeled patterns is available where each pair $(x_i,y_i)\in R^d\times Y$ has been generated from an unknown distribution independently. The goal is to find a classifier from patterns to labels i.e., $f:R^d\rightarrow Y$. In MIL, the algorithm assumes the input is available as a set of input patterns $x_1,...,x_n$ grouped into bags $B_1,...,B_m$ where $B_I=\{x_i:i\in I\}$ for the given index sets $I \subseteq \{1,...,n\}$. Each bag $B_I$ is associated with label $Y_I$ where $Y_I=-1$ if $y_i=-1$ for all $i \in I$ and $Y_I=1$ if there is at least one instance $x_i\in B_I$ with positive label~\cite{andrews2003support}. The relationship between instance labels $y_i$ and bag labels $Y_I$ can be expressed as $Y_I=\max_{i\in I}y_i$ or a set of linear constraints:
\begin{equation}
\begin{split}
\sum_{i\in I}&\frac{y_i+1}{2}\geq 1,\\ &\forall I\ s.t.\ Y_I=1,\quad\\ & y_i=-1, \forall I\ s.t.\ Y_I=-1.
\end{split}
\end{equation}

The discriminant function~$f:X\rightarrow R$ is called MI-separating with respect to a multiple-instance data set if $sgn\max_{i\in I}\ f(x_i)=Y_I$ for all bags $B_I$ holds.

\subsubsection{Limitation of Support Vector Machine (SVM)}
SVM has been one of the most efficient machine learning algorithms since its introduction in the 1990s~\cite{karamizadeh2014advantage}. However, the SVM algorithms for text classification are limited by the lack of transparency in results caused by a high number of dimensions. Due to this, it cannot show the company score as a parametric function based on financial ratios nor any other functional form~\cite{karamizadeh2014advantage}. A further limitation is a variable financial ratios rate~\cite{guo2014soft}.

\subsection{Decision Tree}
One earlier classification algorithm for text and data mining is decision tree~\cite{morgan1963problems}. Decision tree classifiers (DTCs) are used successfully in many diverse areas for classification~\cite{safavian1991survey}. The structure of this technique is a hierarchical decomposition of the data space
~\cite{morgan1963problems,aggarwal2012survey}. Decision tree as classification task was introduced by~{D. Morgan}~\cite{magerman1995statistical} and developed by~{J.R. Quinlan}~\cite{quinlan1986induction}.
The main idea is creating a tree based on the attribute for categorized data points, but the main challenge of a decision tree is which attribute or feature could be in parents' level and which one should be in child level.
To solve this problem,~{De M{\'a}ntaras}~\cite{de1991distance} introduced statistical modeling for feature selection in tree. For a training set containing~$p$ positive and $n$ negative:
\begin{equation}
\begin{split}
H  \bigg(\frac{p}{n+p},\frac{n}{n+p} \bigg) =& -\frac{p}{n+p} \log_2\frac{p}{n+p}\\&-\frac{n}{n+p} \log_2\frac{n}{n+p}
\end{split}
\end{equation}

Choose attribute \emph{A} with \emph{k} distinct value, divides the training set \emph{E} into subsets of $\{E_1,E_2,\hdots, E_k\}$. The expect entropy (EH) remain after trying attribute \emph{A} (with branches $i = 1, 2, \hdots, k$): 
\begin{equation}
\begin{split}
EH(A) = \sum_{i=1}^K \frac{p_i+n_i}{p+n} H  \bigg(\frac{p_i}{n_i+p_i},\frac{n_i}{n_i+p_i} \bigg)
\end{split}
\end{equation}

Information gain (\emph{I}) 
or reduction in entropy for this attribute is :
\begin{equation}
\begin{split}
A(I) =H  \bigg(\frac{p}{n+p},\frac{n}{n+p} \bigg)-EH(A)
\end{split}
\end{equation}

Choose the attribute with largest information gain as parent's node.

\subsubsection*{Limitation of Decision Tree Algorithm}
The decision tree is a very fast algorithm for both learning and prediction; but it is also extremely sensitive to small perturbations in the data~\cite{giovanelli2017towards}, and can be easily overfit~\cite{quinlan1987simplifying}. These effects can be negated by validation methods and pruning, but this is a grey area~\cite{giovanelli2017towards}. This model also has problems with out-of-sample prediction~\cite{jasimdata}.

\subsection{Random Forest}
\begin{figure}[H]
\centering
\includegraphics[width=0.76\textwidth]{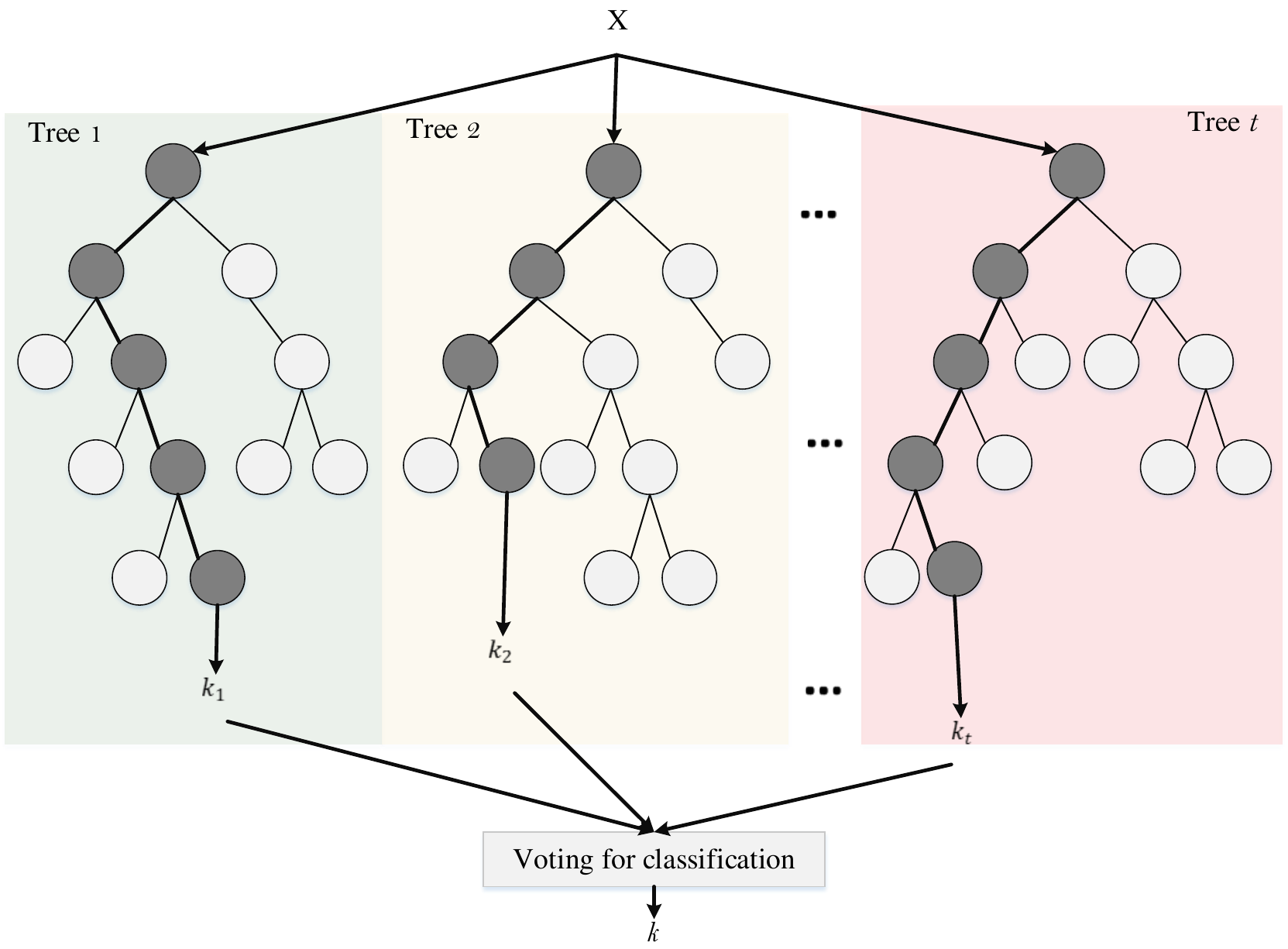}
\caption{Random forest.}\label{Fig:RF}
\end{figure}

Random forests or random decision forests technique is an ensemble learning method for text classification. This method, which used $t$ tree as parallel, was introduced by~{T. Kam Ho}~\cite{Ho1995RF} in 1995.  As~shown in Figure~\ref{Fig:RF}, the main idea of RF is generating random decision trees. This~technique was further developed in 1999 by~{L. Breiman}~\cite{breiman1999random}, who found convergence for RF as margin measures~($mg(X,Y)$) as follows:
\begin{equation}
\begin{split}
mg(X,Y)= &av_k I(h_k(X)=Y)-\\&\max_{j\neq Y} av_k I(h_k(X)=j)
\end{split}
\end{equation}
where $I(.)$ 
refers to indicator function.

\subsubsection{Voting}
After training all trees as forest, predictions are assigned based on voting~\cite{wu2004probability} as follows:
\begin{equation}
\delta _V  = arg\max_i \sum_{j:j \neq j} I_{\{r_{ij} > r_{ji}\}}
\end{equation}
such that
\begin{equation}
r_{ij}+r_{ji} = 1
\end{equation}

\subsubsection{Limitation of Random Forests}

Random forests~(i.e., ensembles of decision trees) are very fast to train for text data sets in comparison to other techniques such as deep learning, but quite slow to create predictions once trained~\cite{Himani2018}. Thus, in order to  achieve a faster structure, the number of trees in forest must be reduced, as more trees in forest increases time complexity in the prediction step.

\subsection{Conditional Random Field~(CRF)}
\begin{figure}[H]
\centering
\includegraphics[width=0.4\textwidth]{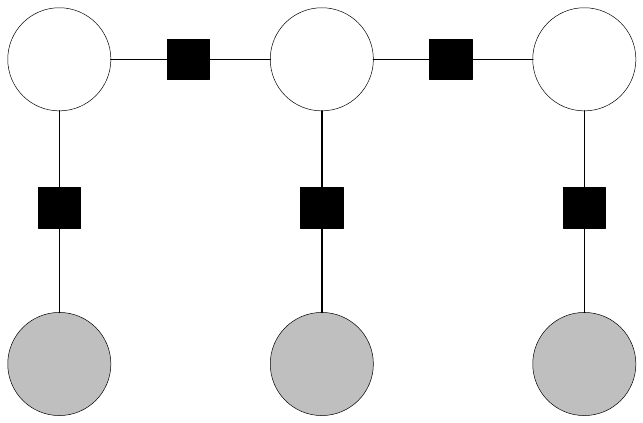}
\caption{Linear-chain conditional random field (CRF). The black boxes are transition clique}\label{fig:CRF} 
\end{figure}

CRF is an undirected graphical model, as shown in Figure \ref{fig:CRF}. CRFs are essentially a way of combining the advantages of classification and graphical modeling that combine the ability to compactly model multivariate data, and the ability to leverage a high dimensional features space for prediction~\cite{sutton2012introduction}~(this model is very powerful for text data due to high feature space). CRFs state the conditional probability of a label sequence~$Y$ given a sequence of observation~$X$ i.e., $P(Y|X)$. CRFs can incorporate complex features into an observation sequence without violating the independence assumption by modeling the conditional probability of the label sequence rather than the joint probability $P(X,Y)$~\cite{vail2007conditional,chen2017improving}. Clique (i.e., fully connected subgraph) potential is used for computing $P(X|Y)$. With respect to the potential function for each clique in the graph, the probability of a variable configuration corresponds to the product of a series of a non-negative potential functions. The value computed by each potential function is equivalent to the probability of the variables in the corresponding clique for a particular configuration~\cite{vail2007conditional}. That is:
\begin{equation}
P(V)=\frac{1}{Z}\prod_{c\in cliques(V)}\psi(c)
\end{equation}
where $Z$ is the normalization term. The conditional probability $P(X|Y)$ can be formulated as:
\begin{equation}
P(Y|X)=\frac{1}{Z}\prod_{t=1}^T\psi(t, y_{t-1},y_t,X)
\end{equation}

Given the potential function~($\psi(t, y_{t-1},y_t,X)=\exp(w.f(t,y_{t-1},y_t,X))$), the conditional probability can be rewritten as:
\begin{equation}
P(Y|X)=\prod_{t=1}^T\exp(w.f(t,y_{t-1},y_t,X))
\end{equation}
where $w$ is the weight vector associated with a feature vector computed by $f$.


\subsubsection*{Limitation of Conditional Random Field (CRF)}
With regards to CRF, the most evident disadvantage of CRF is the high computational complexity of the training step~\cite{sutton2006introduction}, especially for text data sets due to high feature space. Furthermore, this algorithm does not perform with unseen words~(i.e., with words that were not present in the training data sample)~\cite{tseng2005conditional}.


\subsection{Deep Learning}
Deep learning models have achieved state-of-the-art results across many domains, including a wide variety of NLP applications. Deep learning for text and document classification includes three basic architectures of deep learning in parallel. 
We describe each individual model in detail below.

\subsubsection{Deep Neural Networks}
\begin{figure}[H]
\centering
\includegraphics[width=0.74\textwidth]{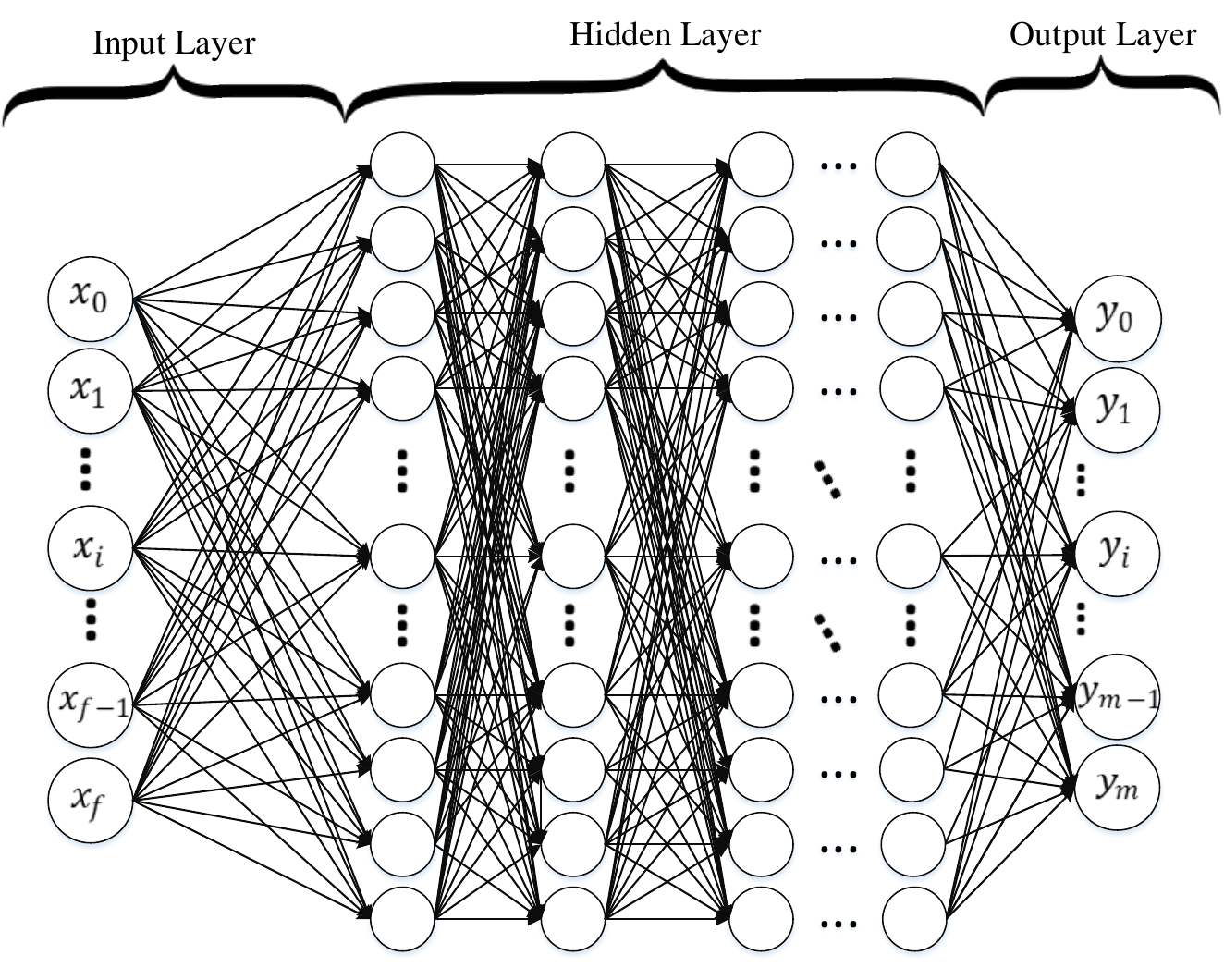}
\caption{Standard, fully connected deep  neural network (DNN).}\label{fig:DNN}
\end{figure}

Deep neural networks~(DNN) are designed to learn by multi-connection of layers that every single layer only receives the connection from previous and provides connections only to the next layer in a hidden part~\cite{kowsari2017HDLTex}. Figure~\ref{fig:DNN} depicts the structure of a standard DNN. The input consists of the connection of the input feature space~(as discussed in Section~\ref{sec:Feature_extraction}) with the first hidden layer of the DNN. The~input layer may be constructed via TF-IDF, word embedding, or some other feature extraction method. The~output layer is equal to the number of classes for multi-class classification or only one for binary classification.
In multi-class DNNs, each learning model is generated~(number of nodes in each layer and the number of layers are completely randomly assigned). The implementation of DNN is a discriminative trained model that uses a standard back-propagation algorithm using sigmoid~(Equation~\eqref{sigmoid}), ReLU~ 
\cite{nair2010rectified} (Equation~\eqref{relu}) as an activation function.~The output layer for multi-class classification should be a $Softmax$ function~(as shown in Equation~\eqref{Softmax}).
\begin{align}
f(x) &= \frac{1}{1+e^{-x}}\in (0,1)\label{sigmoid}\\
f(x) &= \max(0,x)\label{relu}\\
\sigma(z)_j &= \frac{e^{z_j}}{\sum_{k=1}^K e^{z_k}}\label{Softmax}\\
&\forall  ~j \in \{1,\hdots, K\} \nonumber
\end{align}

Given a set of example pairs $(x,y)$,$x\in X,y\in Y $, the goal is to learn the relationship between these input and target spaces using hidden layers. In text classification applications, the input is a string which is generated via vectorization of the raw text data.

\subsubsection{Recurrent Neural Network~(RNN)}
\begin{figure}[H]
\centering
\includegraphics[width=0.54\textwidth]{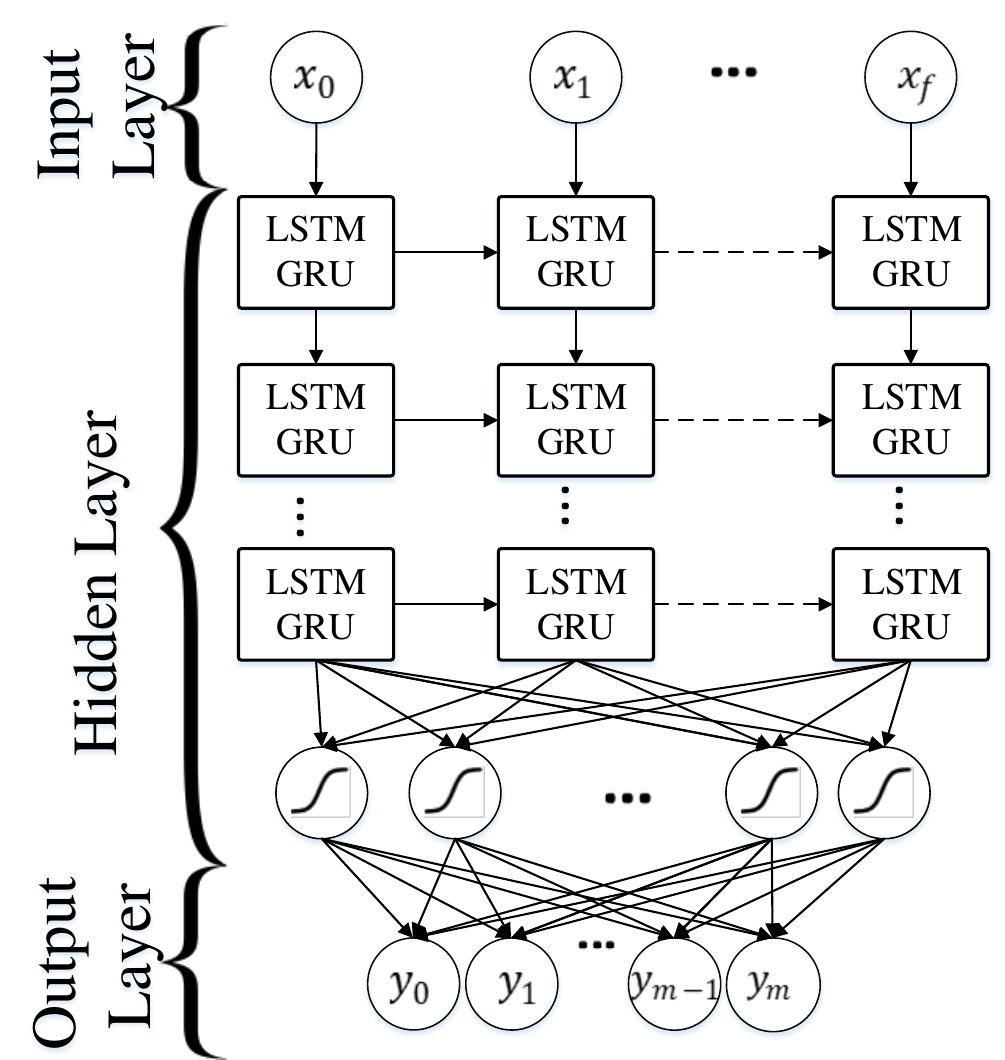}
\caption{Standard long short-term memory (LSTM)/GRU 
recurrent neural networks.}\label{fig:RNN}
\end{figure}

Another neural network architecture that researchers have used for text mining and classification is recurrent neural network~(RNN)~\cite{sutskever2011generating,mandic2001recurrent}. RNN assigns more weights to the previous data points of a sequence. Therefore, this technique is a powerful method for text, string, and sequential data classification. 
A RNN considers the information of previous nodes in a very sophisticated method which allows for better semantic analysis of a data set's structure. RNN mostly works by using LSTM or GRU 
for text classification, as shown in Figure~\ref{fig:RNN} which contains input layer~(word embedding), hidden layers, and finally output layer.  This method can be formulated as:

\begin{equation}
\label{rnn_gen}
x_{t}=F(x_{t-1},\boldsymbol{u_t},\theta)
\end{equation}
where $x_t$ is the state at time $t$ and $\boldsymbol{u_t}$ refers to the input at step~$t$.
More specifically, we can use weights to formulate Equation~\eqref{rnn_gen}, parameterized by:

\begin{equation}\label{rnn_spec}
x_{t}=\mathbf{W_{rec}}\sigma(x_{t-1})+\mathbf{W_{in}}\mathbf{u_t}+\mathbf{b}
\end{equation}
where $\mathbf{W_{rec}}$ refers to recurrent matrix weight, $\mathbf{W_{in}}$ refers to input weights, $\mathbf{b}$ is the bias, and $\sigma$ denotes an element-wise function.

Figure~\ref{fig:RNN} illustrates an extended RNN architecture. Despite the benefits described above, RNN is vulnerable to the problems of vanishing gradient and exploding gradient when the error of the gradient descent algorithm is back propagated through the network~\cite{bengio1994learning}.

\subsubsection*{Long Short-Term Memory~(LSTM)}
LSTM was introduced by S. Hochreiter and J. Schmidhuber~\cite{hochreiter1997long}, and~has since been augmented by many research scientists~\cite{graves2005framewise}.

LSTM is a special type of RNN that addresses these problems by preserving long term dependency in a more effective way in comparison to the basic RNN. LSTM is particularly useful with respect to overcoming the vanishing gradient problem~\cite{pascanu2013difficulty}. Although LSTM has a chain-like structure similar to RNN, LSTM uses multiple gates to carefully regulate the amount of information that is allowed into each node state. Figure~\ref{fig:LSTM} shows the basic cell of an LSTM model. A step-by-step explanation of a LSTM cell is as follows:
\begin{align}
&&i_{t}=&\sigma(W_{i}[x_{t},h_{t-1}]+b_{i}),&& \label{eq:lstm1}\\
&&\tilde{C_{t}}=&\tanh(W_{c}[x_{t},h_{t-1}]+b_{c}),&& \label{eq:lstm2} \\
&&f_{t}=&\sigma(W_{f}[x_{t},h_{t-1}]+b_{f}),&& \label{eq:lstm3}\\
&&C_{t}=&  i_{t}* \tilde{C_{t}}+f_{t} C_{t-1},&& \label{eq:lstm4}\\
&&o_{t}=& \sigma(W_{o}[x_{t},h_{t-1}]+b_{o}),&& \label{eq:lstm5}\\
&&h_{t}=&o_{t}\tanh(C_{t}),&&\label{eq:lstm6}
\end{align}
where Equation~\eqref{eq:lstm1} represents the input gate, Equation~\eqref{eq:lstm2} represents the candid memory cell value, Equation~\eqref{eq:lstm3} defines forget-gate activation, Equation~\eqref{eq:lstm4} calculates the new memory cell value, and  Equations~\eqref{eq:lstm5} and~\eqref{eq:lstm6} define the final output gate value.
In the above description, each $b$ represents a bias vector, each $W$ represent a weight matrix, and $x_{t}$ represents input to the memory cell at time~$t$. Furthermore,~$i$,$c$,$f$,$o$ indices refer to input, cell memory, forget and output gates respectively.
Figure~\ref{fig:LSTM} shows a graphical representation of the structure of these gates.

A RNN can be biased when later words are more influential than earlier ones. Convolutional neural network~(CNN) models~(discussed in Section~\ref{sec:CNN}) were introduced to overcome this bias by deploying a max-pooling layer to determine discriminative phrases in text data~\cite{lai2015recurrent}.

\subsubsection*{Gated Recurrent Unit~(GRU)}\label{subsec:GRU}
\begin{figure}[H]
\centering
\includegraphics[width=\textwidth]{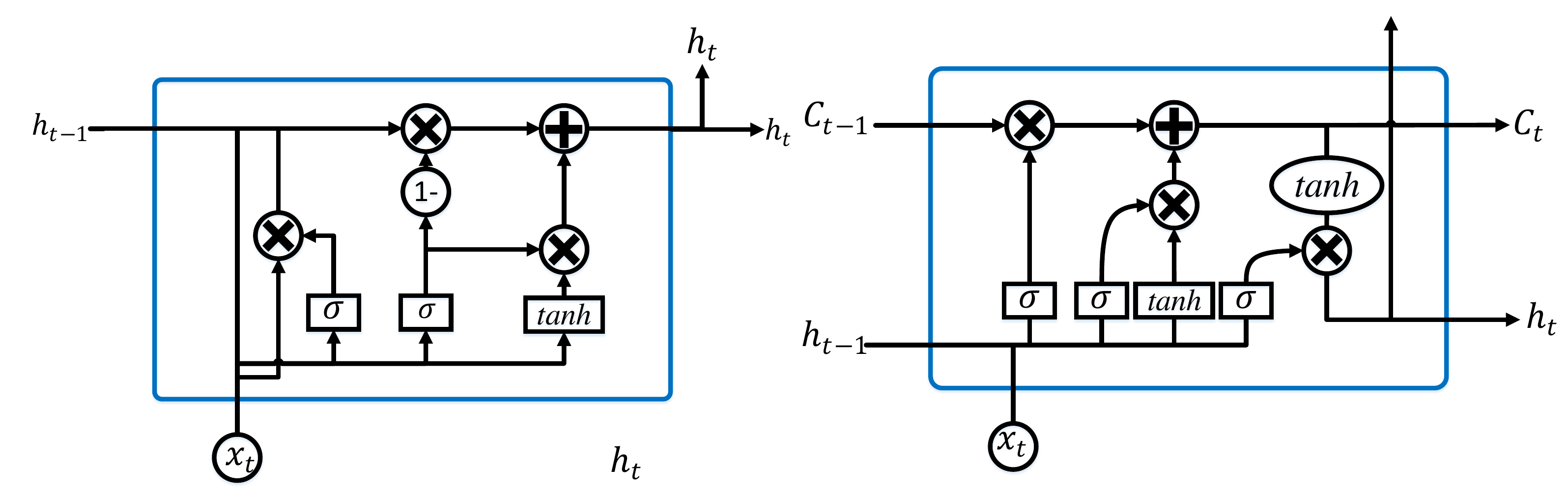}
\caption{The left figure is a GRU cell while the right figure is a LSTM cell.}  \label{fig:LSTM}
\end{figure}

GRUs are a gating mechanism for RNN formulated by~{J. Chung et al.}~\cite{chung2014empirical} and~{K. Cho et al.}~\cite{cho2014learning}. GRUs are a simplified variant of the LSTM architecture. However, a GRU differs from LSTM because it contains two gates and a GRU does not possess internal memory (i.e., the $C_{t-1}$ in Figure~\ref{fig:LSTM}). Furthermore, a second non-linearity is not applied (i.e.,~tanh in Figure~\ref{fig:LSTM}). A~step-by-step explanation of a GRU cell is as follows:
\begin{equation}
z_{t}=\sigma_g(W_{z}x_{t}+U_zh_{t-1}+b_{z}), \label{eq:gru1}
\end{equation}
where~$z_t$ refers to the update gate vector of~$t$,~$x_t$ stands for input vector, $W$, $U$, and~$b$ represent parameter matrices/vectors. The activation function~($\sigma_g$) is either a sigmoid or ReLU and can be formulated as~follow:
\begin{equation}
\tilde{r_{t}}=\sigma_g(W_{r}x_{t}+U_rh_{t-1}+b_{r}), \label{eq:gru2}
\end{equation}
where $r_t$ stands for reset gate vector of~$t$, $z_t$ is update gate vector of~$t$.
\begin{equation}
\begin{split}
h_t =  &z_t \circ h_{t-1} + (1-z_t) \circ \\ &\sigma_h(W_{h} x_t + U_{h} (r_t \circ h_{t-1}) + b_h)\label{eq:gru6}
\end{split}
\end{equation}
where~$h_t$ is output vector of~$t$, and $\sigma_h$ indicates the hyperbolic tangent function.

\subsubsection{Convolutional Neural Networks~(CNN)}\label{sec:CNN}
A convolutional neural network~(CNN) is a deep learning architecture that is commonly used for hierarchical document classification~\cite{jaderberg2016reading,lai2015recurrent}. Although originally built for image processing, CNNs have also been effectively used for text classification~\cite{lecun1998gradient,lecun2015deep}. In a basic CNN for image processing, an image tensor is convolved with a set of kernels of size $d \times d$. These convolution layers are called feature maps and can be stacked to provide multiple filters on the input. To reduce the computational complexity, CNNs use pooling to reduce the size of the output from one layer to the next in the network. Different pooling techniques are used to reduce outputs while preserving important features~\cite{scherer2010evaluation}.

\begin{figure}[H]
\centering
\includegraphics[width=\textwidth]{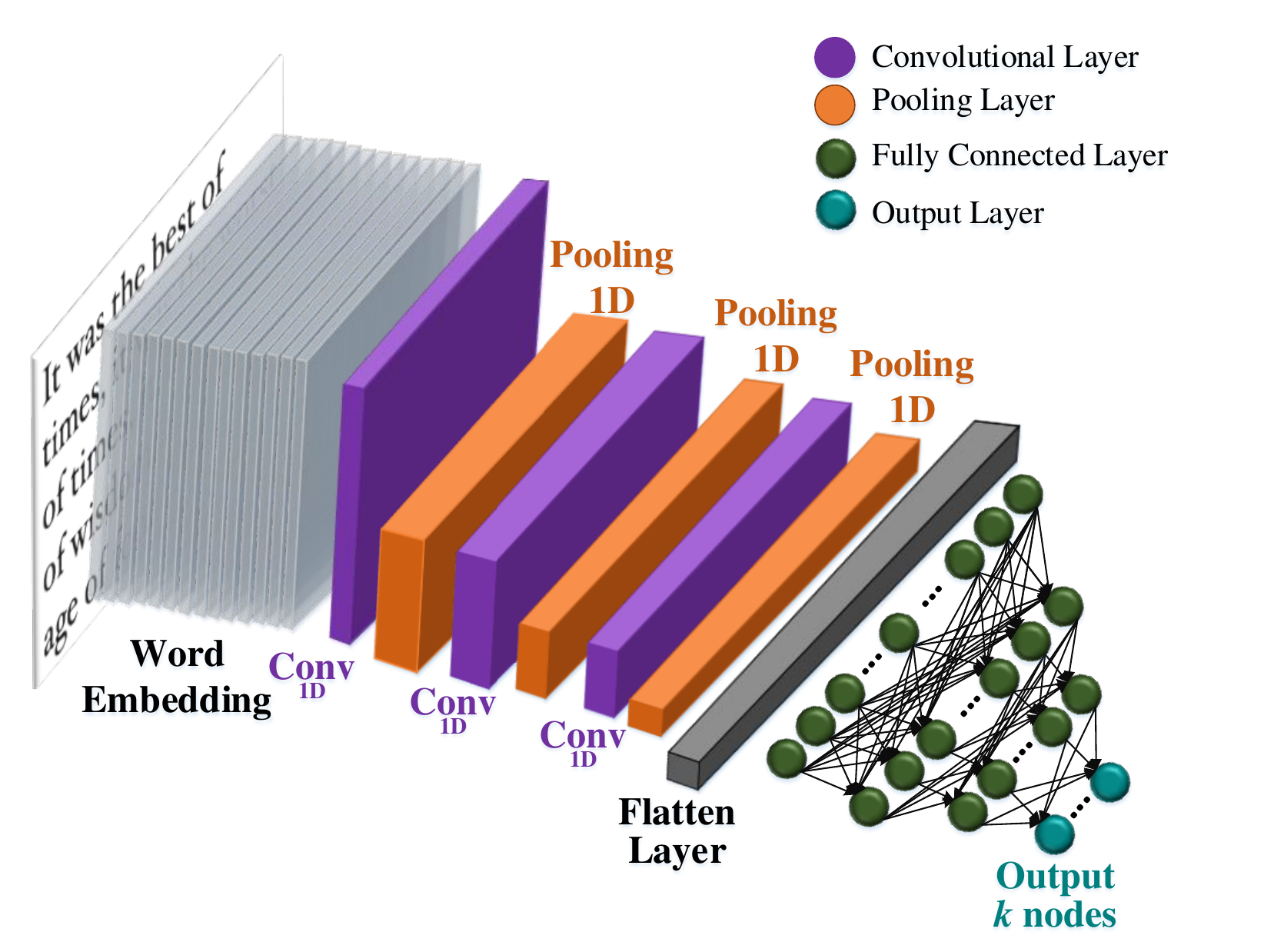}
\caption{Convolutional neural network (CNN) architecture for text classification.}  \label{fig:CNN_Ar}
\end{figure}

The most common pooling method is max pooling where the maximum element in the pooling window is selected. In~order to feed the pooled output from stacked featured maps to the next layer, the maps are flattened into one column. The final layers in a CNN are typically fully connected.
In~general, during the back-propagation step of a convolutional neural network, both the weights and the feature detector filters are adjusted. A potential problem that arises when using CNN for text classification is the number of 'channels', $\Sigma$~(size of the feature space). While image classification application generally have few channels~({e.g.}, only 3 channels of RGB), $\Sigma$ may be very large~({e.g.}, 50 K) for text classification applications~\cite{johnson2014effective}, thus resulting in very high dimensionality. Figure~\ref{fig:CNN_Ar} illustrate the CNN architecture for text classification which contains word embedding as input layer 1D convolutional layers, 1D~pooling layer, fully connected layers, and finally output layer.

\subsubsection{Deep Belief Network~(DBN)}
A deep belief network~(DBN) is a deep learning structure that is superposed by restricted Boltzmann machines~(RBMs)~\cite{jiang2018text}. A RBM is a generative artificial neural network which could learn probability distribution over samples. Contrastive divergence~(CD)~\cite{hinton2002training} is a training technique used for RBMs~\cite{hinton2006fast,mohamed2012acoustic}.

The energy function is as follows:
\begin{equation}
E(v,h) = -\sum_i a_i v_i - \sum_j b_j h_j -\sum_i \sum_j v_i w_{i,j} h_j
\end{equation}
where~$a_i$ is visible units and~$b_i$ refers to hidden units in matrix notation. This expression can be simplified as:
\begin{equation}
E(v,h) = -a^{\mathrm{T}} v - b^{\mathrm{T}} h -v^{\mathrm{T}} W h
\end{equation}

Given a configuration of the hidden units~$h$ is defined as follows:
\begin{equation}
P(v|h) = \prod_{i=1}^m P(v_i|h)
\end{equation}

For Bernoulli, the logistic function for visible units is replaced as follows:
\begin{equation}
P(v_i^k = 1|h) = \frac{\exp(a_i^k + \Sigma_j W_{ij}^k h_j)} {\Sigma_{k'=1}^K \exp(a_i^{k'} + \Sigma_j W_{ij}^{k'} h_j)}
\end{equation}

The update function with gradient descent is as follows:
\begin{equation}
w_{ij}(t+1) = w_{ij}(t) + \eta\frac{\partial \log(p(v))}{\partial w_{ij}}
\end{equation}

\subsubsection{Hierarchical Attention Networks (HAN)}
\begin{figure}[H]
\centering
\includegraphics[width=0.60553\textwidth]{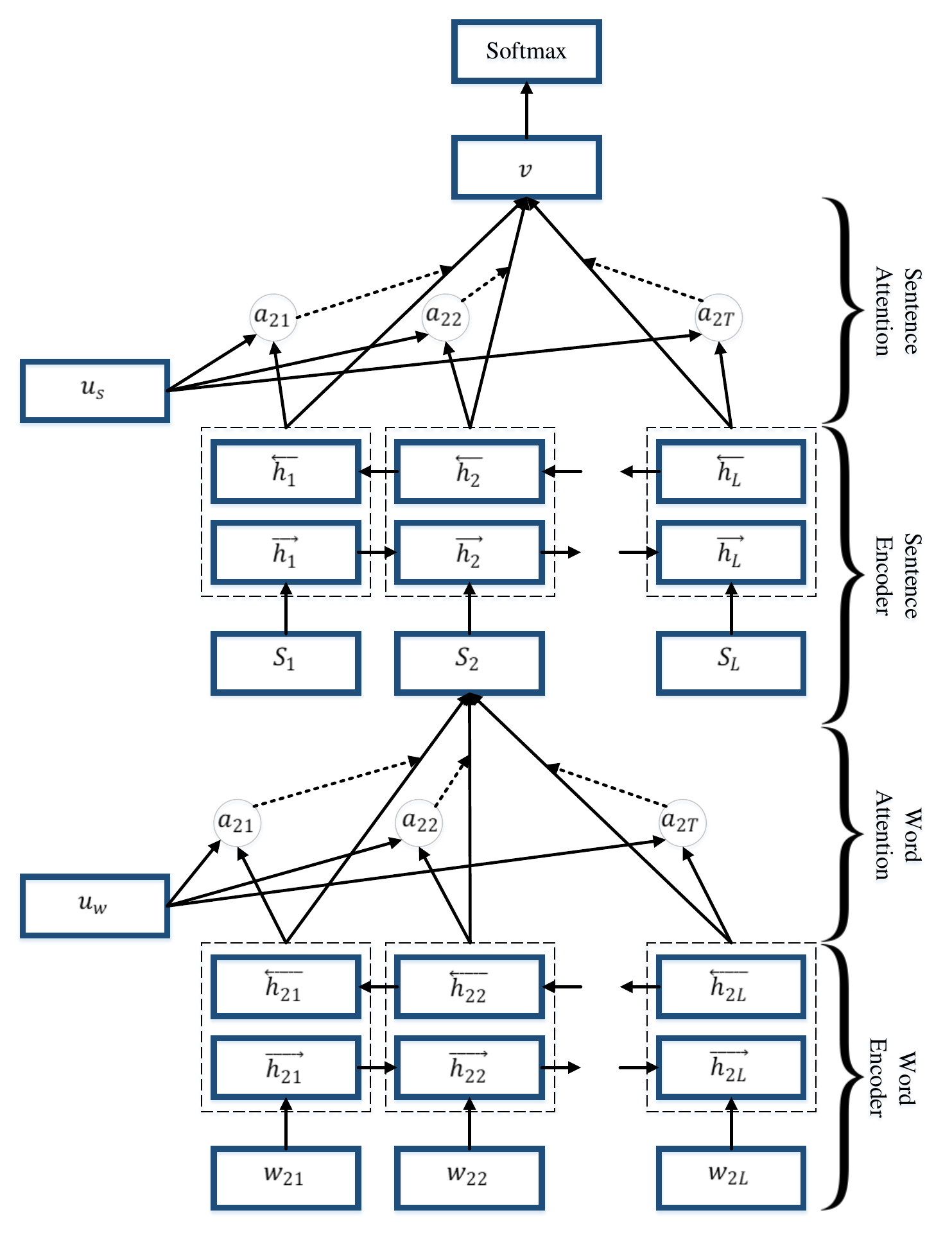}
\caption{Hierarchical attention networks for document classification.}\label{fig:HAN}
\end{figure}

One of the successful deep architecture for text and document classification is hierarchical attention networks (HAN). This technique was introduced by~{Z. Yang et al.}~\cite{yang2016hierarchical} and {S.P. Hongsuck et al.}~\cite{seo2016hierarchical}. The structure of a HAN focuses on the document-level classification which a document has~$L$ sentences and each sentence contains~$T_i$ words, where~$w_{it}$ with~$t \int [1, T]$ represents the words in the \emph{i}th
sentence. HAN architecture is illustrated in Figure~\ref{fig:HAN}, where the lower level contains word encoding and word attention and the upper level contains sentence encoding and sentence attention.

\subsubsection{Combination Techniques}
Many researchers combine or concatenate standard deep learning architectures in order to develop novel techniques with more robust and accurate architectures for classification tasks. In this sub-section, we describe recent and popular deep learning architectures and structure.

\subsubsection*{Random Multimodel Deep Learning~(RMDL)}

\begin{figure}[H]
\centering
\includegraphics[width=\textwidth]{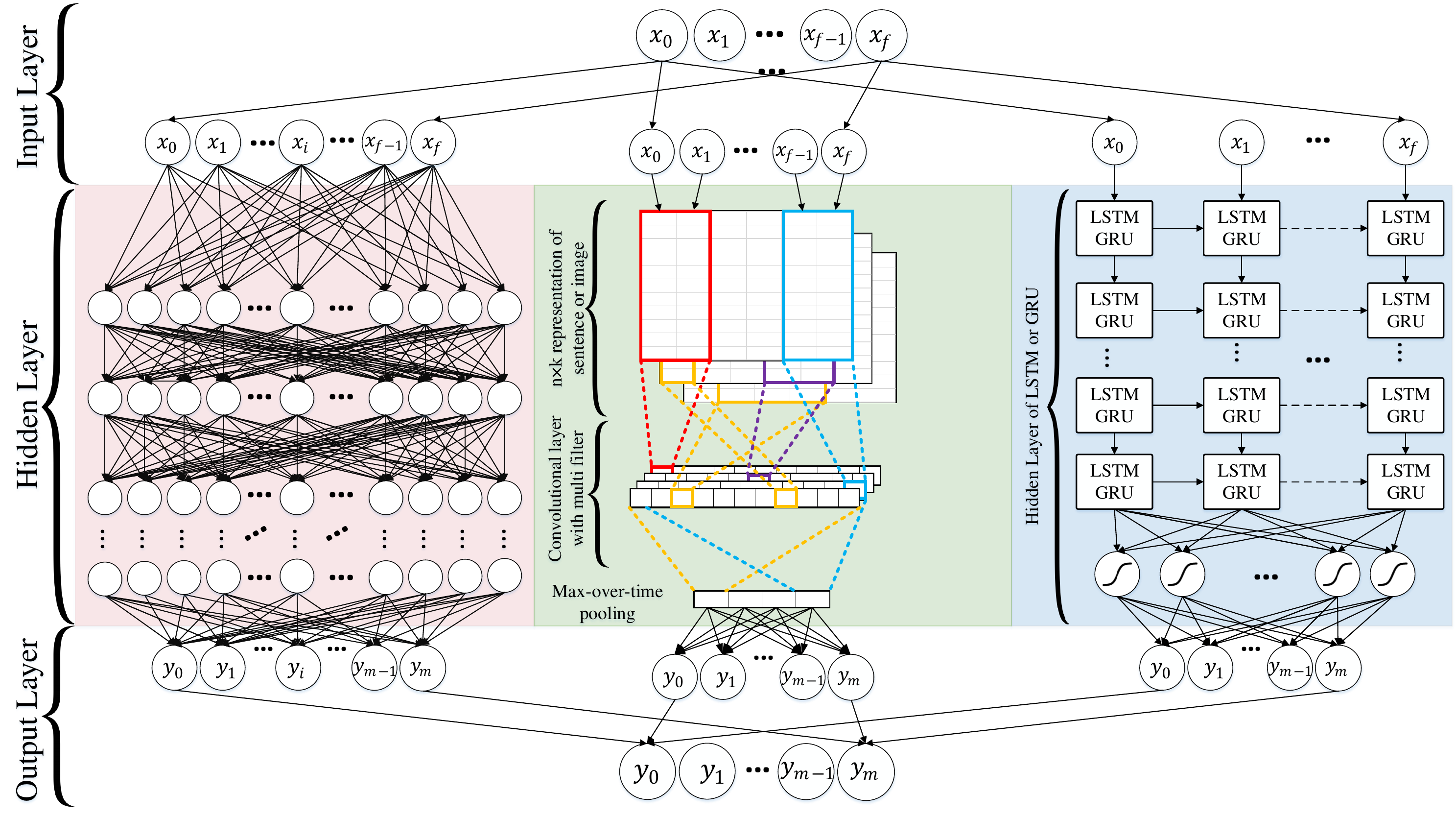}
\caption{{R}andom {m}ultimodel {e}eep {l}earning~(RDML) 
architecture for classification. RMDL includes 3 random models: A deep neural network (DNN) classifier ({\bf left}), a deep CNN classifier ({\bf middle}), and a deep recurrent neural network (RNN) classifier ({\bf right}). Each unit could be a LSTM or GRU).}  \label{fig:RMDL}
\end{figure}

Random multimodel deep learning (RMDL) was introduced by~{K. Kowsari et al.}~\cite{Kowsari2018RMDL,Heidarysafa2018RMDL} as a novel deep learning technique for classification. RMDL can be used in any kind of data set for classification. An overview of this technique is shown in Figure~\ref{fig:RMDL} which illustrates the architecture using multi-DNN, deep CNN, and deep RNN. The number of layers and nodes for all of these deep learning multi-models are generated randomly ({e.g.}, 9 random models in RMDL constructed from~$3$ CNNs,~$3$ RNNs, and $3$~DNNs, all of which are unique due to random creation).
\begin{align}
\label{eq:majority}
M(y_{i1},y_{i2},...,y_{in}) =& \bigg\lfloor \frac{1}{2}+ \frac{(\sum_{j=1}^n y_{ij}) - \frac{1}{2}}{n}\bigg\rfloor
\end{align}
where $n$ is the number of random models, and $y_{ij}$ is the output prediction of model for data point $i$ in  model $j$~(Equation~\eqref{eq:majority} is used for binary classification, $k\in\{0~ \text{or}~1\}$). The output space uses majority vote to calculate the final value of $\hat{y_i}$. Therefore,~$\hat{y_i}$ is given as follows:
\begin{equation}
\hat{y_i} =
\begin{bmatrix}
\hat{y}_{i1}~
\hdots~
\hat{y}_{ij}~
\hdots~
\hat{y}_{in}~
\end{bmatrix}^T
\end{equation}
where $n$ is the number of the random model, and $\hat{y}_{ij}$ shows the prediction of the label of data point (e.g., document) of $D_i \in \{x_i,y_i\}$ for model $j$ and $\hat{y}_{i,j}$ is defined as follows:

\begin{equation}
\hat{y}_{i,j} = arg \max_{k} [ softmax(y_{i,j}^*)]
\end{equation}

After all the RDL models~(RMDL) are trained, the final prediction is calculated using majority vote on the output of these models. The main idea of using multi-model with different optimizers is that if one optimizer does not provide a good fit for a specific data set, the RMDL model with~$n$ random models~(where some of them might use different optimizers) could ignore~$k$ models which are not efficient if and only if $n>k$. Using multi-techniques of optimizers (e.g., SGD, Adam, RMSProp, Adagrad, Adamax) helps the RMDL model be more suitable for any type of data sets. While we only used~$2$ optimizers~(Adam and RMSProp) for evaluating the model in this research, the RMDL model can use any kind of optimizer.
In this part, we describe common optimization techniques used in deep learning architectures.

\vspace{6pt}
\noindent\textbf{Stochastic Gradient Descent~(SGD) Optimizer:}
\vspace{6pt}

The basic equation for stochastic gradient descent~(SGD)~\cite{bottou2010large} is shown in Equation~\eqref{SGD}. SGD~uses a momentum on re-scaled gradient which is shown in Equation~\eqref{momentum} for updating parameters.
\begin{align}
\theta &\leftarrow \theta - \alpha \nabla_\theta J(\theta , x_i,y_i)\label{SGD}\\
\theta &\leftarrow \theta -\big( \gamma \theta + \alpha \nabla_\theta J(\theta , x_i,y_i)\big)\label{momentum}
\end{align}

\vspace{6pt}
\noindent\textbf{RMSprop:}
\vspace{6pt}

T. Tieleman and G. Hinton~\cite{tieleman2012lecture} introduced RMSprop as a novel optimizer which divides	the	learning rate for a weight by a running average	of the magnitudes of recent gradients for that weight. The equation of the momentum method for RMSprop is as follows:
\begin{equation}
v(t) = \alpha~ v(t-1)-\epsilon \frac{\partial E}{\partial w}(t)
\end{equation}
\begin{align}
\Delta w(t) &= v(t)\nonumber\\
&=\alpha~ v(t-1)-\epsilon \frac{\partial E}{\partial w}(t)\\
&=\alpha~ \Delta v(t-1)-\epsilon \frac{\partial E}{\partial w}(t)\nonumber
\end{align}

RMSProp does not do bias correction, which causes significant problems when dealing with a sparse gradient.

\vspace{6pt}
\noindent\textbf{Adam Optimizer}
\vspace{6pt}

Adam is another stochastic gradient optimizer which uses only the first two moments of gradient~($v$ and $m$, shown in Equations~\eqref{adam}--\eqref{adam3}) and calculate the average over them. It can handle non-stationary of the objective function as in RMSProp while overcoming the sparse gradient issue limitation of RMSProp~\cite{kingma2014adam}.
\begin{align}
\theta  &\leftarrow \theta - \frac{\alpha}{\sqrt{\hat{v}}+\epsilon} \hat{m}\label{adam}\\
g_{i,t} &=  \nabla_\theta J(\theta_i , x_i,y_i) \label{adam1}\\
m_t &= \beta_1 m_{t-1} + (1-\beta_1)g_{i,t}\label{adam2}\\
m_t &= \beta_2 v_{t-1} + (1-\beta_2)g_{i,t}^2\label{adam3}
\end{align}
where $m_t$ is the first moment and $v_t$ indicates second moment that both are estimated. $\hat{m_t}=\frac{m_t}{1-\beta_1^t}$ and $\hat{v_t}=\frac{v_t}{1-\beta_2^t}$.

\vspace{6pt}
\noindent\textbf{Adagrad:}
\vspace{6pt}

Adagrad is addressed in~\cite{duchi2011adaptive} as a novel family of sub-gradient methods which dynamically absorb knowledge of the geometry of the data to perform more informative gradient-based learning.

AdaGrad is an extension of SGD. In iteration $k$, the gradient is defined as:
\begin{equation}
G^{(k)} = diag\bigg[\sum_{i=1}^k g^{(i)}(g^{(i)})^T\bigg]^{1/2}
\end{equation}
diagonal matrix:
\begin{equation}
G^{(k)}_{jj} = \sqrt{\sum_{i=1}^k (g_{i}^{(i)})^2}
\end{equation}
update rule:
\begin{equation}
\begin{aligned}
x^{(k+1)}&= arg\min_{x\in X}\{\langle{\nabla f(x^{(k)}), x}\rangle +\\ &~\qquad~\qquad~\quad\frac{1}{2\alpha_k}||{x-x^{(k)}}||^2_{G^{(k)}}\}\\
&= x^{(k)} - \alpha B^{-1}\nabla f(x^{(k)}) \qquad(\text{if }{X} = \!R^n)
\end{aligned}
\end{equation}

\vspace{6pt}
\noindent\textbf{Adadelta:}
\vspace{6pt}

AdaDelta, introduced by~{M.D. Zeiler}~\cite{zeiler2012adadelta}, uses the exponentially decaying average of~$g_t$ as $2_{nd}$ moment of gradient. 
This method is an updated version of Adagrad which relies on only first order information. The update rule for Adadelta is:
\begin{equation}
g_{t+1} = \gamma g_t + (1-\gamma)\nabla \mathcal{L}(\theta)^2
\end{equation}
\begin{equation}
x_{t+1} = \gamma x_t + (1-\gamma) v_{t+1}^2
\end{equation}
\begin{equation}
v_{t+1} = - \frac{\sqrt{x_t+\epsilon} \delta L(\theta_t)}{\sqrt{g_{t+1}+\epsilon}}
\end{equation}

\subsubsection*{Hierarchical Deep Learning for Text~(HDLTex)}

\begin{figure}[H]
\centering
\includegraphics[scale=0.23]{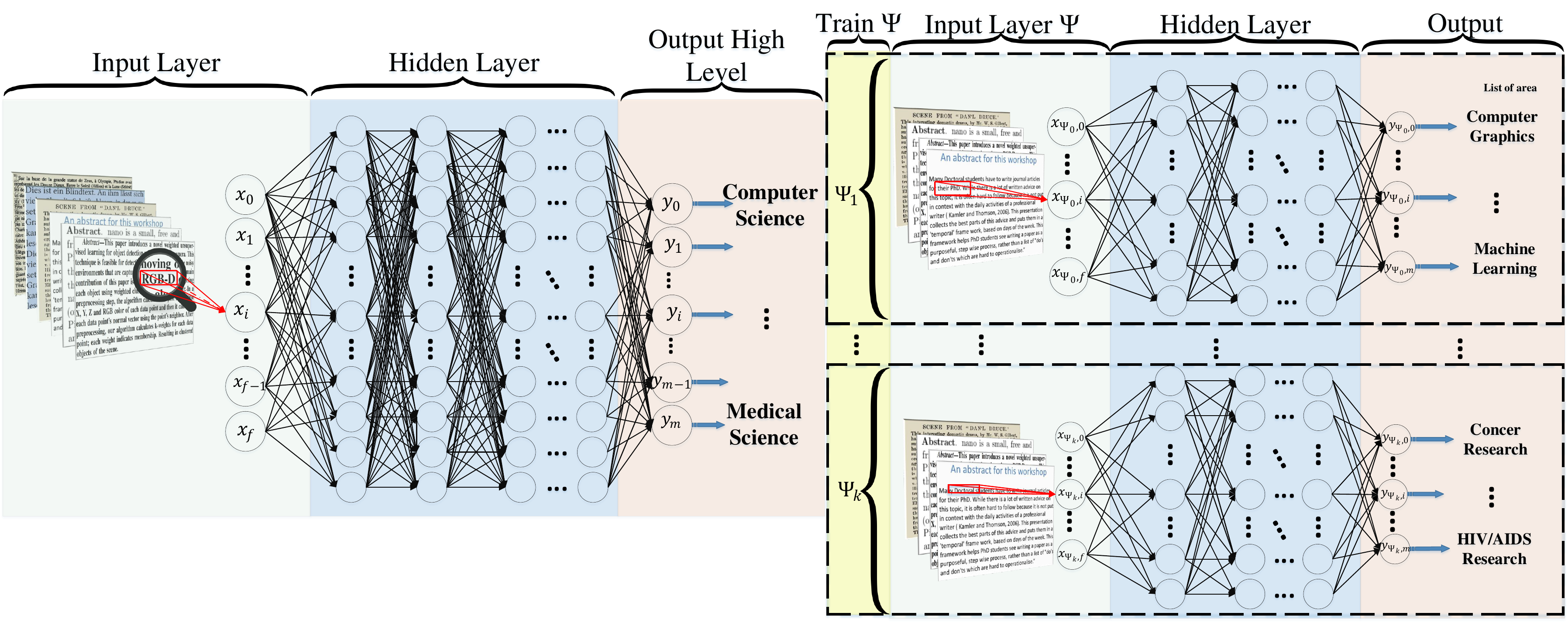}
\caption{HDLTex: {H}ierarchical {d}eep {l}earning for {tex}t classification.
DNN approach for the text classification. The top figure depicts the parent-level of our model, and the bottom figure depicts child-level models~($\Psi_i$) as input documents in the parent level.}  \label{fig:HDLTex}
\end{figure}

The primary contribution of the hierarchical deep learning for text (HDLTex) architecture is hierarchical classification of documents~\cite{kowsari2017HDLTex}. A traditional multi-class classification technique can work well for a limited number of classes, but performance drops with an increasing number of classes, as is present in hierarchically organized documents. In this hierarchical deep learning model, this problem was solved by creating architectures that specialize deep learning approaches for their level of the document hierarchy~({e.g.}, see Figure~\ref{fig:HDLTex}). The structure of the HDLTex architecture for each deep learning model is as follows:
\begin{description}
\item [DNN:] $8$ hidden layers with $1024$ cells in each hidden layer.
\item [RNN:] GRU and LSTM are used in this implementation, $100$ cells with GRU with two hidden layers.
\item [CNN:] Filter sizes of $\{3, 4, 5, 6, 7\}$ and max-pool of $5$, layer sizes of $\{128,128,128\}$ with max pooling of $\{5,5,35\}$, the CNN contains $8$ hidden layers.
\end{description}

All models were constructed using the following parameters: \textit{Batch Size = 128}, \textit{learning parameters = $0.001$, $\beta_1$ = 0.9, $\beta_2$ = 0.999, $\epsilon=1e^{08}$,   $decay=0.0$, Dropout = 0.5~(DNN), and Dropout = 0.25 (CNN and~RNN)}. 

HDLTex uses the following cost function for the deep learning models evaluation:
\begin{equation}
\label{eq_cost}
\begin{split}
Acc(X) =&\sum_{\varrho} \bigg{[} \frac{Acc(X_{\Psi_\varrho})}{k_{\varrho}-1}\\
&\sum_{\Psi \in \{ \Psi_1,..\Psi_k\}} Acc(X_{\Psi_i}).n_{\Psi_k}\bigg{]}
\end{split}
\end{equation}
where $\varrho$ is the number of levels, $k$ indicates number of classes for each level, and $\Psi$ refers to the number of classes in the child's level of the hierarchical model.

\subsubsection*{Other Techniques}
In this section, we discuss other techniques of text classification that come from combining deep learning architectures. Recurrent convolutional neural networks~(RCNN) is used for text classification~\cite{lai2015recurrent,wang2017scene}. RCNNs can capture contextual information with the recurrent structure and construct the representation of text using a CNN~\cite{lai2015recurrent}. This architecture is a combination of RNN and CNN that leverages the advantages of both techniques in a model.

C-LSTM is another technique of text and document classification that was introduced by {C.~Zhou~et~al.}~\cite{zhou2015c}. C-LSTM combines CNN with LSTM in order to learn phrase-level features using convolutional layers. This architecture feeds sequences of higher level representations into the LSTM to learn long-term dependent.

\subsubsection{Limitation of Deep Learning}
\begin{figure}[H]
\centering
\includegraphics[width=0.7\textwidth]{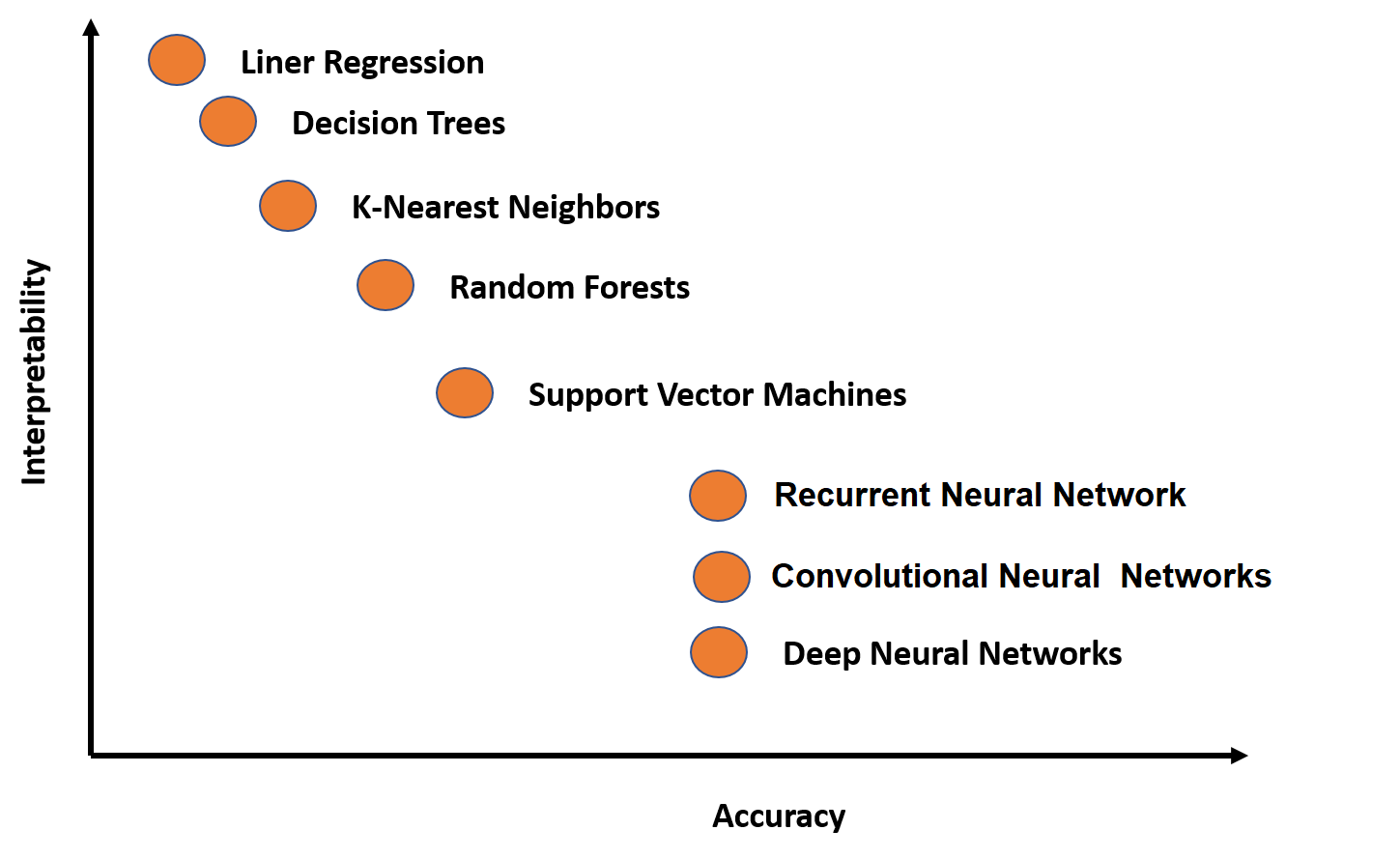}
\caption{The model
interpretability comparison between traditional and deep learning techniques.} \label{fig:interpretability}
\end{figure}

Model interpretability of deep learning (DL), especially DNN, has~always been a limiting factor for use cases requiring explanations of the features involved in modelling and such is the case for many healthcare problems. This problem is due to scientists preferring to use traditional techniques such as  linear models, Bayesian Models, SVM, decision trees, etc. for their works. The weights in a neural network are a measure of how strong each connection is between each neuron to find the important feature space. As shown in Figure~\ref{fig:interpretability}, the more accurate model, the interpretability is lower which means the complex algorithms such as deep learning is hard to understand.

Deep learning~(DL) is one of the most powerful techniques in artificial intelligence~(AI), and many researchers and scientists focus on deep learning architectures to improve the robustness and computational power of this tool.  However, deep learning architectures also have some disadvantages and limitations when applied to classification tasks. One of the main problems of this model is that DL does not facilitate a comprehensive theoretical understanding of learning~\cite{shwartz2017opening}. A well-known disadvantage of DL methods is their \qu{black box} nature~\cite{gray1996alternatives,shrikumar2017learning}. That is, the method by which DL methods come up with the convolved output is not readily understandable. Another limitation of DL is that it usually requires much more data than traditional machine learning algorithms, which means that this technique cannot be applied to classification tasks over small data sets~\cite{anthes2013deep,lampinen2017one}. Additionally, the massive amount of data needed for DL classification algorithms further exacerbates the computational complexity during the training step~\cite{severyn2015learning}.

\subsection{Semi-Supervised Learning for Text Classification}

Many researchers have developed many efficient classifiers for both labeled and unlabeled documents. Semi-supervised learning is a type of supervised learning problem that uses unlabeled data to train a model. Usually, researchers and scientists prefer to use semi-supervised techniques when a small part of the data set contains labeled data points and a large amount of data set does not include labels~\cite{gowda2016semi}. Most of the semi-supervised learning algorithms for classification tasks use a clustering technique~(generally used for unsupervised learning~\cite{kowsari2014investigation}) as follows: Initially a clustering technique is applied on $D^T$ with~$K=K$~(the number of classes), since $D^T$ has labeled samples of all~$K$ classes~\cite{gowda2016semi}. If a partition $P_i$ has labeled samples, then, all data points on that cluster belongs to that~label.

The research goal for clustering techniques is to determine if we have more than one class labeled on one cluster, and what happens if we have no labeled data point in one cluster~\cite{kowsari2015construction}. In~this part, we~briefly describe the most popular technique of semi-supervised text and document classification.~{O.~Chapelle and A.~Zien}~\cite{chapelle2005semi} worked on semi-supervised classification via low density separation, which combines graph distance computation with transductive support vector machine~(TSVM) training. {K.~Nigam~et~al.}~\cite{nigam2006semi} developed a technique for text classification using expectation maximization~(EM) and generative models for semi-supervised learning with labeled and unlabeled data in the field of text classifications. {L.~Shi~et~al.}~\cite{shi2010cross} introduced a method for transferring classification knowledge across languages via translated features. This technique uses an EM algorithm that naturally takes into account the ambiguity associated with the translation of a word. {J.~Su~et~al.}~\cite{shi2010cross} introduced~\qu{Semi-supervised Frequency Estimate~(SFE)}, a MNBC method for large scale text classification. {S.~Zhou~et~al.}~\cite{zhou2014fuzzy} invented a novel deep learning method that uses fuzzy DBN for semi-supervised sentiment classification. This method employs a fuzzy membership function for each category of reviews based on the learned architecture.

\section{Evaluation}\label{sec:Evaluation}
\begin{figure}[H]
\centering
\includegraphics[width=0.56\textwidth]{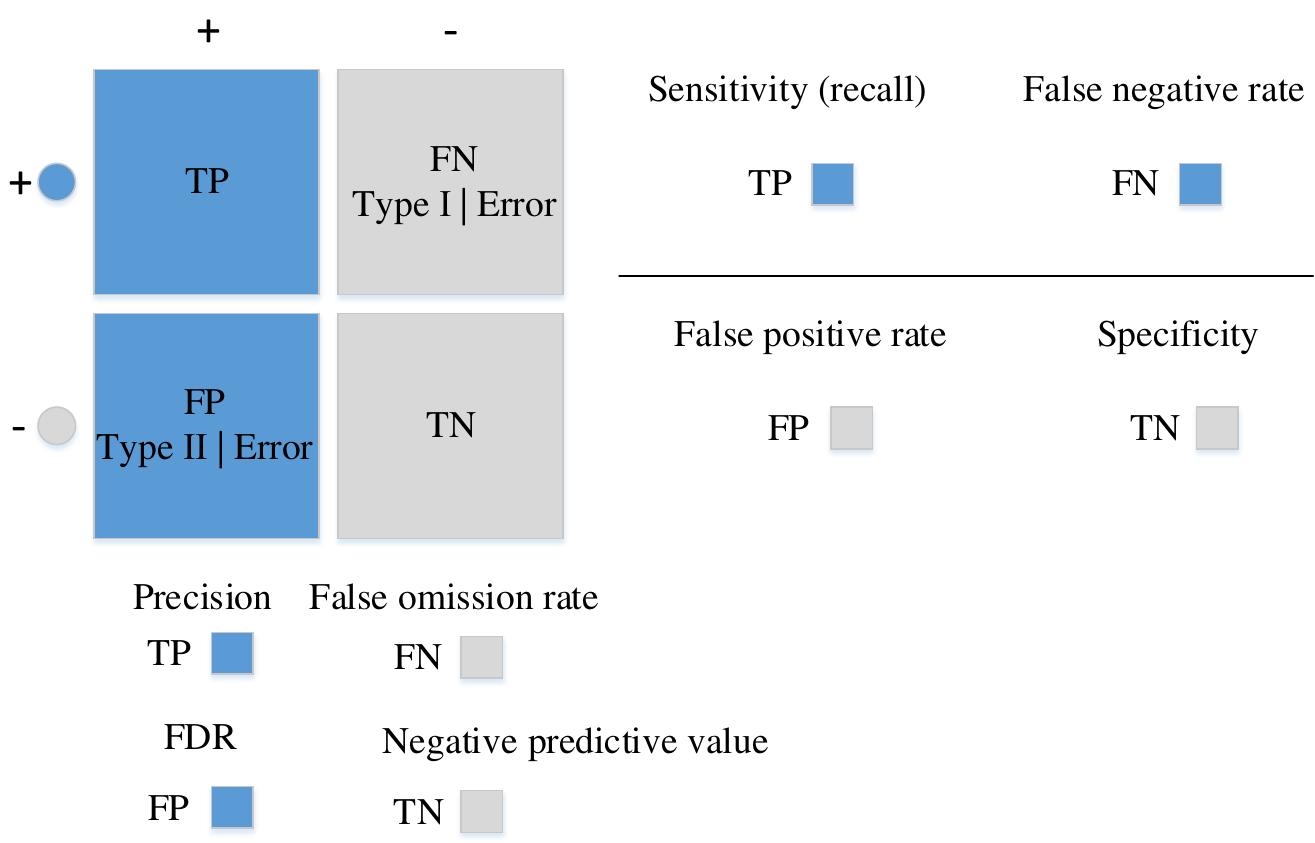}
\caption{Confusion matrix.}  \label{fig:F1}
\end{figure}

In the research community, having shared and comparable performance measures to evaluate algorithms is preferable. However, in reality such measures may only exist for a handful of methods. The major problem when evaluating text classification methods is the absence of standard data collection protocols. Even if a common collection method existed ({e.g.,} Reuters news corpus), simply choosing different training and test sets can introduce inconsistencies in model performance~\cite{yang1999evaluation}. Another challenge with respect to method evaluation is being able to compare different performance measures used in separate experiments. Performance measures generally evaluate specific aspects of classification task performance, and thus do not always present identical information. In this section, we discuss evaluation metrics and performance measures and highlight ways in which the performance of classifiers can be compared. Since the underlying mechanics of different evaluation metrics may vary, understanding what exactly each of these metrics represents and what kind of information they are trying to convey is crucial for comparability. Some examples of these metrics include recall, precision, accuracy, F-measure, micro-average, and macro average. These metrics are based on a~``confusion matrix''~(shown in Figure~\ref{fig:F1}) that comprises true positives~(TP), false positives~(FP), false negatives~(FN), and true negatives~(TN)~\cite{lever2016points}. The significance of these four elements may vary based on the classification application. The fraction of correct predictions over all predictions is called accuracy (Equation~\eqref{eq:acc}). The fraction of known positives that are correctly predicted is called sensitivity i.e., true positive rate or recall (Equation~\eqref{eq:recall}). The ratio of correctly predicted negatives is called specificity (Equation~\eqref{eq:spec}). The proportion of correctly predicted positives to all positives is called precision,~i.e., positive predictive value (Equation~\eqref{eq:pres}).
\begin{align}
accuracy&=\frac{(TP+TN)}{(TP+FP+FN+TN)}\label{eq:acc}\\
sensitivity&=\frac{TP}{(TP+FN)}\label{eq:recall}\\
specificity&=\frac{TN}{(TN+FP)}\label{eq:spec}\\
precision &= \frac{\sum_{l=1}^LTP_l}{\sum_{l=1}^LTP_l+FP_l}\label{eq:pres}\\
recall&= \frac{\sum_{l=1}^LTP_l}{\sum_{l=1}^LTP_l+FN_l}\\
F1-Score &=  \frac{\sum_{l=1}^L2TP_l}{\sum_{l=1}^L2TP_l+FP_l+FN_l}
\end{align}

\subsection{Macro-Averaging and Micro-Averaging}
A single aggregate measure is required when several two-class classifiers are being used to process a collection. Macro-averaging gives a simple average over classes while micro-averaging combines per-document decisions across classes and then outputs an effective measure on the pooled contingency table~\cite{manning2008matrix}. Macro-averaged results can be computed as follows:
\begin{equation}
B_{macro} = \frac{1}{q}\sum_{\lambda=1}^q B(TP_{\lambda}+FP_{\lambda}+TN_{\lambda}+FN_{\lambda})
\end{equation}
where $B$ is a binary evaluation measure calculated based on true positives~(TP), false positives~(FP), false negatives~(FN), and true negatives~(TN), and $L=\{\lambda_j : j=1...q\}$ is the set of all labels.

Micro-averaged results~\cite{sebastiani2002machine,tsoumakas2009mining} can be computed as follows:
\begin{equation}
\begin{split}
B_{macro} = B\bigg(&\sum_{\lambda=1}^q TP_{\lambda}, \sum_{\lambda=1}^q FP_{\lambda},\sum_{\lambda=1}^q TN_{\lambda}, \sum_{\lambda=1}^q FN_{\lambda} \bigg)
\end{split}
\end{equation}

Micro-average score assigns equal weights to every document as a consequence, and it is considered to be a per-document average. On the other hand, macro-average score assigns equal weights to each category without accounting for frequency and therefore, it is a per-category average.

\subsection{F$_{\beta}$ Score}
$F_{\beta}$ is one of the most popular aggregated evaluation metrics for classifier evaluation~\cite{lever2016points}. The~parameter $\beta$ is used to balance recall and precision and is defined as follows:
\begin{equation}{\label{eq:fbeta}}
F_{\beta} = \frac{(1+\beta^2)(precision \times recall)}{\beta^2 \times precision+recall}
\end{equation}

For commonly used $\beta=1$~i.e., $F_1$, recall and precision are given equal weights and Equation~\eqref{eq:fbeta} can be simplified to:
\begin{equation}{\label{eq:f1}}
F_1=\frac{2TP}{2TP+FP+FN}
\end{equation}

Since $F_\beta$ is based on recall and precision, it does not represent the confusion matrix fully.

\subsection{Matthews Correlation Coefficient (MCC)}
The Matthews correlation coefficient~(MCC)~\cite{matthews1975comparison} captures all the data in a confusion matrix and measures the quality of binary classification methods. MCC can be used for problems with uneven class sizes and is still considered a balanced measure. MCC ranges from $-1$ to $0$ (i.e., the classification is always wrong and always true, respectively). MCC can be calculated as follows:
\begin{equation}{\label{eq:mcc}}
MCC=\frac{TP \times TN - FP \times FN}{\sqrt{\splitdfrac{(TP+FP)\times(TP+FN)\times}{(TN+FP)\times(TN+FN)}}}
\end{equation}

While comparing two classifiers, one may have a higher score using MCC and the other one has a higher score using $F_1$ and as a result one specific metric cannot captures all the strengths and weaknesses of a classifier~\cite{lever2016points}.

\subsection{Receiver Operating Characteristics (ROC)}
Receiver operating characteristics~(ROC)~\cite{yonelinas2007receiver} curves are valuable graphical tools for evaluating classifiers. However, class imbalances (i.e., differences in prior class probabilities~\cite{japkowicz2002class}) can cause ROC curves to poorly represent the classifier performance. ROC curve plots true positive rate~(TPR) and false positive rate~(FPR):
\begin{equation}
TPR = \frac{TP}{TP + FN}
\end{equation}
\begin{equation}
FPR = \frac{FP}{FP + TN}
\end{equation}

\subsection{Area Under ROC Curve (AUC)}
The area under ROC curve~(AUC)~\cite{hanley1982meaning,pencina2008evaluating} measures the entire area underneath the ROC curve. AUC leverages helpful properties such as increased sensitivity in the analysis of variance~(ANOVA) tests, independence from decision threshold, invariance to a priori class probabilities, and indication of how well negative and positive classes are in regarding the decision index~\cite{bradley1997use}.\\

For binary classification tasks, AUC can be formulated as:
\begin{equation}
\begin{split}
AUC &= \int_{-\infty}^{\infty} TPR(T) FPR'(T) dT\\
&= \int_{-\infty}^{\infty}\int_{-\infty}^{\infty} I(T'>T)f_1(T')f_0(T)dT dT'\\
&= P(X_1>X_0)
\end{split}
\end{equation}

For multi-class AUC, an average AUC can be defined~\cite{hand2001simple} as follows:
\begin{equation}
AUC = \frac{2}{|C|(|C|-1)} \sum_{i=1}^{|C|} AUC_i
\end{equation}
where \emph{C} is the number of the classes.

Yang~\cite{yang1999evaluation} evaluated statistical approaches for text categorization and reported the following important factors that should be considered when comparing classifier algorithms:

\begin{itemize}
\item Comparative evaluation across methods and experiments which gives insight about factors underlying performance variations and will lead to better evaluation methodology in the future;
\item Impact of collection variability such as including unlabeled documents in training or test set and treat them as negative instances can be a serious problem;
\item Category ranking evaluation and binary classification evaluation show the usefulness of classifier in interactive applications and emphasize their use in a batch mode respectively. Having both types of performance measurements to rank classifiers is helpful in detecting the effects of thresholding strategies;
\item Evaluation of the scalability of classifiers in large category spaces is a rarely investigated area.
\end{itemize}

\section{Discussion}\label{sec:Discussion}
In this article, we aimed to present a brief overview of text classification techniques, alongside a discussion of corresponding pre-processing steps and evaluation methods. In this section, we~compare and contrast each of these techniques and algorithms. Moreover, we discuss the limitations of existing classification techniques and evaluation methods. The main challenge to choosing an efficient classification system is understanding similarity and differences of available techniques in different pipeline steps.

\subsection{Text and Document Feature Extraction}
We outlined the following two main feature extraction approaches:~Weighted words (bag-of-words) and word embedding. Word embedding techniques learn from sequences of words by taking into consideration their occurrence and co-occurrence information. Also, these methods are unsupervised models for generating word vectors. In contrast, weighted words features are based on counting words in documents and can be used as a simple scoring scheme of word representation. Each technique presents unique limitations.

Weighted words computes document similarity directly from the word-count space which increases the computational time for large vocabularies~\cite{wu2008interpreting}. While counts of unique words provide independent evidence of similarity, they do not account for semantic similarities between words~({e.g.,}~\qu{Hello} and \qu{Hi}). Word embedding methods address this issue but are limited by the necessitation of a huge corpus of text data sets for
training~\cite{rezaeinia2017improving}. As a result, scientists prefer to use pre-trained word embedding vectors~\cite{rezaeinia2017improving}. However, this approach cannot work for words missing from these text data corpora.

For example, in some short message service~(SMS) data sets, people use words with multiple meaning such as slang or abbreviation which do not have semantic similarities.
Furthermore, abbreviations are not included in the pre-trained word embedding vectors. To solve these problems, many researchers work on text cleaning as we discussed in Section~\ref{sec:Feature_extraction}. The word embedding techniques such as GloVe, FastText, and Word2Vec were trained based on the word and nearest neighbor of that word, and this contains a very critical limitation~(the meaning of a word could be different in two different sentences). To solve this problem, scientist have come up with the novel methods called contextualized word representations, which train based on the context of the word in a document.

As shown in Table~\ref{tb:_Feature_extraction_comparision}, we compare and evaluate each technique including weighted words, TF-IDF, Word2Vec, Glove, FastText, and contextualized word representations.

\begin{table}[H]
\centering
\caption{Feature extraction comparison.}
\label{tb:_Feature_extraction_comparision}
\begin{tabular}{>{\centering\arraybackslash}m{2cm} >{\raggedright\let\arraybackslash}m{5.5cm}
>{\raggedright\arraybackslash}m{6.5cm}
}

\toprule
\multicolumn{1}{c}{\bf Model}         & \multicolumn{1}{c}{\bf Advantages}
&  \multicolumn{1}{c}{\bf Limitation}                                                                \\  \hline

Weighted Words  & \begin{itemize} [leftmargin=*,rightmargin=1pt,parsep=5pt,topsep=2pt,partopsep=2pt]
\item Easy to compute
\item Easy to compute the similarity between 2 documents using it
\item Basic metric to extract the most descriptive terms in a document
\item Works with unknown word ({e.g.,}~New words in languages)\vspace{-2mm}
\end{itemize} &
\begin{itemize} [leftmargin=*,rightmargin=1pt,parsep=5pt,topsep=2pt,partopsep=2pt]
\item It does not capture position in text (syntactic)
\item It does not capture meaning in text (semantics)
\item Common words effect on the results ({e.g.,}~\qu{am}, \qu{is}, etc.)
\end{itemize} \\  \hline
TF-IDF  &  \begin{itemize} [leftmargin=*,rightmargin=1pt,parsep=5pt,topsep=2pt,partopsep=2pt] \item Easy to compute \item Easy to compute the similarity between 2 documents using it \item Basic metric to extract the most descriptive terms in a document \item Common words do not effect the results due to IDF ({e.g.,} \qu{am}, \qu{is}, etc.)\vspace{-2mm}   \end{itemize} & \begin{itemize} [leftmargin=*,rightmargin=1pt,parsep=5pt,topsep=2pt,partopsep=2pt]   \item It does not capture position in text (syntactic) \item It does not capture meaning in text (semantics) \end{itemize}  \\   \hline
Word2Vec  & \begin{itemize} [leftmargin=*,rightmargin=1pt,parsep=5pt,topsep=2pt,partopsep=2pt]   \item It captures position of the words in the text (syntactic) \item It captures meaning in the words (semantics)  \end{itemize} & \begin{itemize} [leftmargin=*,rightmargin=1pt,parsep=5pt,topsep=2pt,partopsep=2pt]   \item It cannot capture the meaning of the word from the text (fails to capture polysemy) \item It cannot capture  out-of-vocabulary words from corpus\vspace{-2mm}   \end{itemize}
\\
\hline

GloVe (Pre-Trained)  & \begin{itemize} [leftmargin=*,rightmargin=1pt,parsep=5pt,topsep=2pt,partopsep=2pt]   \item It captures position of the words in the text (syntactic) \item It captures meaning in the words (semantics) \item Trained on huge corpus \end{itemize} & \begin{itemize} [leftmargin=*,rightmargin=1pt,parsep=5pt,topsep=2pt,partopsep=2pt] \item It cannot capture the meaning of the word from the text (fails to capture polysemy) \item Memory consumption for storage  \item It cannot capture  out-of-vocabulary words from corpus\vspace{-2mm}  \end{itemize}  \\

\bottomrule
\end{tabular}
\end{table}

\begin{table}[H]\ContinuedFloat
\centering
\caption{\textit{Cont}.}
\label{tb:_Feature_extraction_comparision}
\begin{tabular}{>{\centering\arraybackslash}m{2cm} >{\raggedright\let\arraybackslash}m{5.5cm}
>{\raggedright\arraybackslash}m{6.5cm}
}

\toprule
\multicolumn{1}{c}{\bf Model}         & \multicolumn{1}{c}{\bf Advantages}
&  \multicolumn{1}{c}{\bf Limitation}                                                                \\  \hline

GloVe (Trained)  & \begin{itemize} [leftmargin=*,rightmargin=1pt,parsep=5pt,topsep=2pt,partopsep=2pt]  \item It is very straightforward, e.g., to~enforce the word vectors to capture sub-linear relationships in the vector space (performs better than Word2vec) \item Lower weight for highly frequent word pairs such as stop words like \qu{am}, \qu{is}, etc. Will not dominate training progress\vspace{-2mm}  \end{itemize} & \begin{itemize} [leftmargin=*,rightmargin=1pt,parsep=5pt,topsep=2pt,partopsep=2pt] \item Memory consumption for storage  \item Needs huge corpus to learn \item It cannot capture  out-of-vocabulary words from corpus \item It cannot capture the meaning of the word from the text (fails to capture polysemy) \end{itemize} \\   \hline
FastText   & \begin{itemize} [leftmargin=*,rightmargin=1pt,parsep=5pt,topsep=2pt,partopsep=2pt] \item Works for rare words (rare in their character n-grams which are still shared with other words \item Solves out of vocabulary words with n-gram in character level \end{itemize} & \begin{itemize} [leftmargin=*,rightmargin=1pt,parsep=5pt,topsep=2pt,partopsep=2pt] \item It cannot capture the meaning of the word from the text (fails to capture polysemy) \item Memory consumption for storage \item Computationally is more expensive in comparing with GloVe and Word2Vec\vspace{-2mm}  \end{itemize} \\  \hline
Contextualized Word Representations  &  \begin{itemize} [leftmargin=*,rightmargin=1pt,parsep=5pt,topsep=2pt,partopsep=2pt] \item It captures the meaning of the word from the text (incorporates context, handling polysemy)   \end{itemize} & \begin{itemize} [leftmargin=*,rightmargin=1pt,parsep=5pt,topsep=2pt,partopsep=2pt] \item Memory consumption for storage \item  Improves performance notably on downstream tasks. Computationally is more expensive in comparison to others \item Needs another word embedding for all LSTM and feed forward layers \item It cannot capture  out-of-vocabulary words from corpus \item Works only sentence and document level (it cannot work for individual word level) \vspace{-5.5mm}  \end{itemize}  \\

\bottomrule
\end{tabular}
\end{table}

\subsection{Dimensionality Reduction}
In Section~\ref{sec:Dimensionality}, we outlined many  dimensionality reduction techniques. In this section, we discuss how the efficacy of this step with respect to a text classification system's computational time and robustness. Dimensionality reduction is mostly used for improving computational time and reducing memory complexity.

PCA attempts to find orthogonal projections of the data set which contains the highest variance possible in order to extract linear correlations between variables of the data set. The main limitation of PCA is the computational complexity of this technique for dimensionality reduction~\cite{sharma2007fast}. To~solve this problem, scientists introduced the random projection technique~(Section~\ref{sec:Dimensionality}).

LDA is a supervised technique for dimension reduction that can improve the predictive performance of the extracted features.~However, LDA requires researchers to manually input the number of components, requires labeled data, and produces features that are not easily interpretable~\cite{putthividhya2011bootstrapped}.

Random projection is much faster computationally than PCA. However, this method does not perform well for small data sets~\cite{banerjee2011utility}.

Autoencoders requires more data to train than other DR methods, and thus cannot be used as a general-purpose dimensionality reduction algorithm without sufficient data.

T-SNE is mostly used for data visualization in text and document data sets.

\subsection{Existing Classification Techniques}
In this section, we discuss the limitations and advantages of existing text and document classification algorithms. Then we compare state-of-the-art techniques in two tables.

\subsubsection{Limitations and Advantages}

As shown in Tables~\ref{tb:model_comparison} and~\ref{tb:Model_Comparison_2}, the Rocchio algorithm is limited by the fact that the user can only retrieve a few relevant documents using this model~\cite{selvi2017text}.~Furthermore, the algorithms' results illustrate several limitations in text classification, which could be addressed by taking semantics into consideration~\cite{albitar2012towards}. Boosting and bagging methods also have many limitations and disadvantages, such as  the computational complexity and loss of interpretability~\cite{geurts2000some}. LR works well for predicting categorical outcomes. However, this prediction requires that each data point be independent~\cite{huang2015unconstrained} which is attempting to predict outcomes based on a set of independent variables~\cite{guerin2016using}. Na\"ive Bayes algorithm also has several limitations. NBC makes a strong assumption about the shape of the data distribution~\cite{soheily2017intrusion,wang2012nonparametric}. NBC is also limited by data scarcity for which any possible value in feature space, a likelihood value must be estimated by a frequentist~\cite{ranjan2017document}. KNN is a classification method that is easy to implement and adapts to any kind of feature space. This model also naturally handles multi-class cases~\cite{sahgal2015road,patel2014ant}. However, KNN is limited by data storage constraint for large search problems to find the nearest neighbors.
Additionally, the performance of KNN is dependent on finding a meaningful distance function, thus making this technique a very data-dependent algorithm~\cite{sahgal2014object,sanjay4comparing}.~SVM has been one of the most efficient machine learning algorithms since its introduction in the 1990s~\cite{karamizadeh2014advantage}. However, they are limited by the lack of transparency in results caused by a high number of dimensions. Due to this, it cannot show the company score as a parametric function based on financial ratios nor any other functional form~\cite{karamizadeh2014advantage}. A further limitation is a variable financial ratios rate~\cite{guo2014soft}. The decision tree is a very fast algorithm for both learning and prediction, but it is also extremely sensitive to small perturbations in the data~\cite{giovanelli2017towards}, and can be easily overfit~\cite{quinlan1987simplifying}. These effects can be negated by validation methods and pruning, but this is a grey area~\cite{giovanelli2017towards}. This model also has problems with out-of-sample prediction~\cite{jasimdata}. Random forests~(i.e., ensembles of decision trees) are very fast to train in comparison to other techniques, but quite slow to create predictions once trained~\cite{Himani2018}. Thus, in order to  achieve a faster structure, the number of trees in forest must be reduced, as more trees in forest increases time complexity in the prediction step. With regards to CRF, the~most evident disadvantage of CRF is the high computational complexity of the training step~\cite{sutton2006introduction}, and this algorithm does not perform with unknown words~(i.e., with words that were not present in training data sample)~\cite{tseng2005conditional}. Deep learning~(DL) is one of the most powerful techniques in artificial intelligence~(AI), and many researchers and scientists focus on deep learning architectures to improve the robustness and computational power of this tool.
However, deep learning architectures also have some disadvantages and limitations when applied to classification tasks. One of the main problems of this model is that DL does not facilitate comprehensive theoretical understanding of learning~\cite{shwartz2017opening}. A~well-known disadvantage of DL methods is their \qu{black box} nature~\cite{gray1996alternatives,shrikumar2017learning}. That is, the method by which DL methods come up with the convolved output is not readily understandable. Another limitation of DL is that it usually requires much more data than traditional machine learning algorithms, which means that this technique cannot be applied to classification tasks over small data sets~\cite{anthes2013deep,lampinen2017one}. Additionally, the massive amount of data needed for DL classification algorithms further exacerbates the computational complexity during the training step~\cite{severyn2015learning}.

\begin{table}[H]
\centering
\caption{Text classification comparison~(Rocchio algorithm, boosting, bagging, logistic regression, Na\"ive Bayes classifier, k-nearest Neighbor, and Support Vector Machine).}
\label{tb:model_comparison}
\begin{tabular}{>{\centering\arraybackslash}m{2.5cm} >{\raggedright\let\arraybackslash}m{5.5cm}
>{\raggedright\arraybackslash}m{5.5cm}
}

\toprule
\multicolumn{1}{c}{\bf Model}         & \multicolumn{1}{c}{\bf Advantages}
&  \multicolumn{1}{c}{\bf Pitfall}                                                                \\  \hline
Rocchio Algorithm  & \begin{itemize} [leftmargin=*,rightmargin=1pt,parsep=5pt,topsep=2pt,partopsep=2pt] \item Easy to implement \item Computationally is very cheap \item Relevance feedback mechanism (benefits to ranking documents as not relevant) \end{itemize} & \begin{itemize} [leftmargin=*,rightmargin=1pt,parsep=5pt,topsep=2pt,partopsep=2pt] \item The user can only retrieve a few relevant
documents  \item Rocchio often misclassifies the type for multimodal class \item This techniques is not very robust \item Linear combination in this algorithm is not good for multi-class data sets\vspace{-2mm} \end{itemize}\\  \hline

Boosting and Bagging  & \begin{itemize} [leftmargin=*,rightmargin=1pt,parsep=5pt,topsep=2pt,partopsep=2pt] \item Improves the stability and accuracy (takes  advantage of ensemble learning where in multiple weak learner outperform a single strong learner)  \item Reducing variance which helps to avoid overfitting problems\vspace{-2mm}\end{itemize} & \begin{itemize} [leftmargin=*,rightmargin=1pt,parsep=5pt,topsep=2pt,partopsep=2pt] \item Computational complexity \item Loss of interpretability (if number of model is high, understanding the model is very difficult) \item Requires careful tuning of different hyper-parameters \end{itemize}\\  \hline

Logistic Regression  & \begin{itemize} [leftmargin=*,rightmargin=1pt,parsep=5pt,topsep=2pt,partopsep=2pt] \item Easy to implement \item Does not require too many computational resources \item It does not require input features to be scaled (pre-processing) \item It does not require any tuning  \end{itemize} & \begin{itemize} [leftmargin=*,rightmargin=1pt,parsep=5pt,topsep=2pt,partopsep=2pt] \item It cannot solve non-linear problems \item  Prediction requires that each data point be independent \item Attempting to predict outcomes based on a set of independent variables\vspace{-2mm} \end{itemize}\\  \hline

Na\"ive Bayes Classifier  & \begin{itemize} [leftmargin=*,rightmargin=1pt,parsep=5pt,topsep=2pt,partopsep=2pt] \item It works very well with text data \item Easy to implement \item Fast in comparison to other algorithms  \end{itemize} & \begin{itemize} [leftmargin=*,rightmargin=1pt,parsep=5pt,topsep=2pt,partopsep=2pt] \item A strong assumption about the shape of the data distribution \item Limited by data scarcity for which any possible
value in feature space, a likelihood value must be estimated by a frequentist\vspace{-2mm} \end{itemize}\\ \hline

K-Nearest Neighbor  & \begin{itemize} [leftmargin=*,rightmargin=1pt,parsep=5pt,topsep=2pt,partopsep=2pt] \item Effective for text data sets \item Non-parametric \item More local characteristics of text or document are considered \item Naturally handles multi-class data sets \end{itemize} & \begin{itemize} [leftmargin=*,rightmargin=1pt,parsep=5pt,topsep=2pt,partopsep=2pt] \item Computational of this model is very expensive \item Difficult to find optimal value of k  \item Constraint for large search problems to find nearest neighbors \item Finding a meaningful distance function is difficult for text data sets\vspace{-2mm} \end{itemize}\\ \hline

Support Vector Machine (SVM)  & \begin{itemize} [leftmargin=*,rightmargin=1pt,parsep=5pt,topsep=2pt,partopsep=2pt] \item SVM can model non-linear decision boundaries \item Performs similarly to logistic regression when linear separation \item Robust against overfitting problems~(especially for text data set due to high-dimensional space) \end{itemize} & \begin{itemize} [leftmargin=*,rightmargin=1pt,parsep=5pt,topsep=2pt,partopsep=2pt] \item Lack of transparency in results caused by a high number of dimensions (especially for text data). \item Choosing an efficient kernel function is difficult (susceptible to overfitting/training issues depending on kernel) \item Memory complexity\vspace{-2mm} \end{itemize}\\ \bottomrule

\end{tabular}
\end{table}
\unskip

\begin{table}[H]
\centering
\caption{Text classification comparison~(decision tree, conditional random field(CRF), random forest, and deep learning).}
\label{tb:Model_Comparison_2}
\begin{tabular}{>{\centering\arraybackslash}m{2.5cm} >{\raggedright\let\arraybackslash}m{6cm}
>{\raggedright\arraybackslash}m{5.5cm}
}

\toprule
\multicolumn{1}{c}{\bf Model}         & \multicolumn{1}{c}{\bf Advantages}
&  \multicolumn{1}{c}{\bf Pitfall} \\  \hline

Decision Tree  & \begin{itemize} [leftmargin=*,rightmargin=1pt,parsep=5pt,topsep=2pt,partopsep=2pt] \item Can easily handle qualitative (categorical) features \item Works well with decision boundaries parellel to the feature axis \item  Decision tree is a very fast algorithm for both learning and prediction  \end{itemize} & \begin{itemize} [leftmargin=*,rightmargin=1pt,parsep=5pt,topsep=2pt,partopsep=2pt] \item Issues with diagonal decision boundaries \item Can be easily overfit \item Extremely sensitive to small perturbations in the data  \item Problems with out-of-sample prediction\vspace{-2mm}\end{itemize}\\ \hline

Conditional Random Field (CRF)
& \begin{itemize} [leftmargin=*,rightmargin=1pt,parsep=5pt,topsep=2pt,partopsep=2pt] \item Its feature design is flexible \item Since CRF computes the conditional probability of global optimal output nodes, it overcomes the drawbacks of label bias \item Combining the advantages of classification and graphical modeling which combine the ability to compactly model multivariate data\vspace{-2mm}  \end{itemize} & \begin{itemize} [leftmargin=*,rightmargin=1pt,parsep=5pt,topsep=2pt,partopsep=2pt] \item High computational complexity of the training step \item This algorithm does not perform with unknown words \item Problem about online learning (It~makes it very difficult to re-train the model when newer data becomes available) \end{itemize}\\  \hline

Random Forest  & \begin{itemize} [leftmargin=*,rightmargin=1pt,parsep=5pt,topsep=2pt,partopsep=2pt] \item Ensembles of decision trees are very fast to train in comparison to other techniques \item Reduced variance (relative to regular trees) \item Does not require preparation and pre-processing of the input data \end{itemize} & \begin{itemize} [leftmargin=*,rightmargin=1pt,parsep=5pt,topsep=2pt,partopsep=2pt] \item Quite slow to create predictions once trained \item More trees in forest increases time complexity in the prediction step \item Not as easy to visually interpret \item Overfitting can easily occur \item Need to choose the number of trees at forest\vspace{-2mm}\end{itemize}\\  \hline

Deep Learning  & \begin{itemize} [leftmargin=*,rightmargin=1pt,parsep=5pt,topsep=2pt,partopsep=2pt] \item Flexible with features design (reduces the need for feature engineering, one of the most time-consuming parts of the machine learning practice) \item Architecture that can be adapted to new problems \item  Can deal with complex input-output mappings \item Can easily handle online learning (It~makes it very easy to re-train the model when newer data becomes available) \item Parallel processing capability~(It can perform more than one job at the same time)\vspace{-2mm} \end{itemize} & \begin{itemize} [leftmargin=*,rightmargin=1pt,parsep=5pt,topsep=2pt,partopsep=2pt] \item Requires a large amount of data (if~you only have small sample text data, deep learning is unlikely to outperform other approaches. \item Is extremely computationally expensive to train. \item Model interpretability is the most important problem of deep learning~(deep learning most of the time is a black-box) 
\item Finding an efficient architecture and structure is still the main challenge of this technique \end{itemize}\\ \bottomrule
\end{tabular}
\end{table}

\subsubsection{State-of-the-Art Techniques' Comparison}

Regarding Tables~\ref{tb:comparing_1} and~\ref{tb:comparing_2}, text classification techniques are compared with the criteria: Architecture, author(s), model, novelty, feature extraction, details, corpus, validation measure, and limitation of each technique. Each text classification technique~(system) contains a model which is the classifier algorithm, and also needs a feature extraction technique which means converting texts or documents data set into numerical data~(as discussed in Section~\ref{sec:Feature_extraction}). Another important part in our comparison is the validation measure which is used to evaluate the system.

\newpage
\paperwidth=\pdfpageheight
\paperheight=\pdfpagewidth
\pdfpageheight=\paperheight
\pdfpagewidth=\paperwidth
\newgeometry{layoutwidth=297mm,layoutheight=210 mm, left=2.7cm,right=2.7cm,top=1.8cm,bottom=1.5cm, includehead,includefoot}
\fancyheadoffset[LO,RE]{0cm}
\fancyheadoffset[RO,LE]{0cm}

\begin{table}[H]
\centering
\caption{Comparison of text classification techniques.}\label{tb:comparing_1}
\scalebox{.93}[0.93]{\begin{tabular}{>{\centering\arraybackslash}m{1.5cm} >{\centering\arraybackslash}m{3.5cm} >{\centering\arraybackslash}m{2cm} >{\raggedright\arraybackslash}m{3.5cm} >{\centering\arraybackslash}m{1.5cm} >{\raggedright\arraybackslash}m{3.5cm} >{\centering\arraybackslash}m{2cm}
>{\centering\arraybackslash}m{1.5cm}
>{\raggedright\arraybackslash}m{3cm}}
\toprule
{\bf Model}     & {\bf Author(s)}    &    {\bf Architecture}             & \multicolumn{1}{c}{\bf Novelty}                            & {\bf Feature Extraction} & \multicolumn{1}{c}{\bf Details}            & {\bf Corpus} & {\bf Validation Measure} & \multicolumn{1}{c}{\bf Limitation}                        \\ \midrule

Rocchio Algorithm              &  B.J. Sowmya et al.~\cite{sowmya2016large}              & Hierarchical Rocchio                                & Classificationon hierarchical data                                                                                                                       & TF-IDF                  & Use CUDA on GPU to compute and compare the distances.                                                                                             & Wikipedia                                        & F1-Macro              & Works only on hierarchical data sets and retrieves a few relevant documents                \\ \midrule
Boosting                       & S. Bloehdorn et al.~\cite{bloehdorn2004boosting}           &                                                     & AdaBoost for with semantic features                                                                                                                      & BOW                    & Ensemble learning algorithm                                                                                                                       & Reuters-21578                                    & F1-Macro and F1-Micro & Computational complexity and loss of interpretability                                    \\ \midrule
Logistic Regression            & A. Genkin et al.~\cite{genkin2007large}       & Bayesian Logistic Regression                        & Logistic regression analysis of high-dimensional data                                                                                                    & TF-IDF                  & It is based on Gaussian Priors and Ridge Logistic Regression                                                                                      & RCV1-v2                                          & F1-Macro              & Prediction outcomes is based on a set of independent variables                           \\ \midrule
Na\"ive Bayes                    & Kim, S.B et al.~\cite{kim2006some}               & Weight Enhancing Method                             & Multivariate poisson model for text Classification                                                                                                       & Weights words          & Per-document term frequency normalization to estimate the Poisson parameter                                                                       & Reuters-21578                                    & F1-Macro              & This method makes a strong assumption about the shape of the data distribution           \\ \midrule
SVM and KNN                    & K. Chen et al.~\cite{chen2016turning}                 & Inverse Gravity Moment                              & Introduced TFIGM (term frequency \& inverse gravity moment)                                                                                              & TF-IDF and TFIGM        & Incorporates a statistical model to precisely measure the class distinguishing power of a term                                                    & 20 Newsgroups and Reuters-21578                  & F1-Macro              & Fails to capture polysemy and also still semantic and sentatics is not solved            \\ \midrule
Support Vector Machines        & H. Lodhi et al.~\cite{lodhi2002text}                & String Subsequence Kernel                           & Use of a special kernel                                                                                                                                  & Similarity using TF-IDF & The kernel is an inner product in the feature space generated by all subsequences of length k                                                     & Reuters-21578                                    & F1-Macro              & The lack of transparency in the results                                                        \\ \midrule
Conditional Random Field (CRF) & T. Chen et al.~\cite{chen2017improving}                & BiLSTM-CRF                                          & Apply a neural network based sequence model to classify opinionated sentences into three types according to the number of targets appearing in a sentence & Word embedding         & Improve sentence-level sentiment analysis via sentence type classification                                                                        & Customer reviews                                 & Accuracy              & High computational complexity and this algorithm does not perform with unseen words      \\ \bottomrule

\end{tabular}}
\end{table}
\unskip

\begin{table}[H]
\centering
\caption{Comparison of the text classification techniques~(continue).}\label{tb:comparing_2}
\scalebox{.95}[0.95]{\begin{tabular}{>{\centering\arraybackslash}m{1.5cm} >{\centering\arraybackslash}m{3cm} >{\centering\arraybackslash}m{2cm} >{\raggedright\arraybackslash}m{3cm} >{\centering\arraybackslash}m{1.5cm} >{\raggedright\arraybackslash}m{3.5cm} >{\centering\arraybackslash}m{2cm}
>{\centering\arraybackslash}m{1.5cm}
>{\raggedright\arraybackslash}m{3cm}}
\toprule
{\bf Model}     & {\bf Author(s)}    &  {\bf Architecture}               & \multicolumn{1}{c}{\bf Novelty}                            & {\bf Feature Extraction} & \multicolumn{1}{c}{\bf Details}            & {\bf Corpus} & {\bf Validation Measure} & \multicolumn{1}{c}{\bf Limitation }                        \\ \midrule
Deep Learning                  & Z. Yang et al.~\cite{yang2016hierarchical}                 & Hierarchical Attention Networks                     & It has a hierarchical structure                                                                                                                          & Word embedding         & Two levels of attention mechanisms applied at the word and sentence-level                                                                         & Yelp, IMDB review, and Amazon review              & Accuracy              & Works only for document-level                                                            \\ \midrule
Deep Learning                  & J. Chen et al.~\cite{chen2018verbal} & Deep Neural Networks                                & Convolutional neural networks (CNN) using 2-dimensional TF-IDF features                                                                                  & 2D TF-IDF               & A new solution to the verbal aggression detection task                                                                                           & Twitter comments                                 & F1-Macro and F1-Micro & Data dependent for designed a model architecture                                         \\ \midrule
Deep Learning                  & M. Jiang et al.~\cite{jiang2018text} & Deep Belief Network                                 & Hybrid text classification model based on deep belief network and softmax regression.                                                                    & DBN                    & DBN completes the feature learning to solve the high dimension and sparse matrix problem and softmax regression is employed to classify the texts & Reuters-21578 and 20-Newsgroup                   & Error-rate            & Computationally is expensive and model interpretability is still a problem of this model \\ \midrule
Deep Learning                  & X. Zhang et al.~\cite{zhang2015character}               & CNN                                                 & Character-level convolutional networks (ConvNets) for text classification                                                                                & Encoded Characters     & Character-level ConvNet contains 6 convolutional layers and 3 fully-connected layers                                                              & Yelp, Amazon review and Yahoo! Answers data set   & Relative errors       & This model is only designed to discover position-invariant features of their inputs      \\ \midrule
Deep Learning                  & K. Kowsari~\cite{Kowsari2018RMDL}                    & Ensemble deep learning algorithm (CNN, DNN and RNN) & Solves the problem of finding the best deep learning structure and architecture                                                                          & TF-IDF and GloVe        & Random Multimodel Deep Learning (RDML)                                                                                                            & IMDB review, Reuters-21578, 20NewsGroup, and WOS & Accuracy              & Computationally is expensive                                                             \\ \midrule
Deep Learning                  & K. Kowsari~\cite{kowsari2017HDLTex}                     & Hierarchical structure                              & Employs stacks of deep learning architectures to provide specialized understanding at each level of the document hierarchy                               & TF-IDF and GloVe        & Hierarchical Deep Learning for Text Classification (HDLTex)                                                                                       & Web of science data set                           & Accuracy              & Works only for hierarchical data sets                                                     \\ \bottomrule

\end{tabular}}
\end{table}

\newpage
\restoregeometry
\paperwidth=\pdfpageheight
\paperheight=\pdfpagewidth
\pdfpageheight=\paperheight
\pdfpagewidth=\paperwidth
\headwidth=\textwidth

\subsection{Evaluation}
The experimental evaluation of text classifiers measures effectiveness (i.e., capacity to make the right classification or categorization decision). Precision and recall are widely used to measure the effectiveness of text classifiers. Accuracy and error~($\frac{FP+FN}{TP+TN+FP+FN}=1-\text{accuracy}$), on the other hand, are not widely used for text classification applications because they are insensitive to variations in the number of correct decisions due to the large value of the denominator~($TP+TN$)~\cite{yang1999evaluation}. The pitfalls of each of the above-mentioned metrics are listed in Table~\ref{table:pitfall}.

\begin{table}[H]
\centering
\caption{Metrics pitfalls.}
\label{table:pitfall}
\begin{tabular}{>{\centering\arraybackslash}m{3cm} >{\raggedright\arraybackslash}m{10cm}}
\toprule
& \multicolumn{1}{c}{\bf Limitation}                                                                                                \\ \midrule
Accuracy    & Gives us no information on False Negative (FN) and False Positive (FP)                                                                                        \\ \midrule
Sensitivity & Does not evaluate True Negative (TN) and FP and any classifier that predicts data points as positives considered to have high sensitivity   \\ \midrule
Specificity & Similar to sensitivity and does not account for FN and TP                                                                   \\ \midrule
Precision   & Does not evaluate TN and FN and considered to be very conservative and goes for a case which is most certain to be positive \\ \bottomrule
\end{tabular}
\end{table}

\section{Text Classification Usage}\label{sec:Application}
In the earliest history of ML and AI, text classification techniques have mostly been used for information retrieval systems. However, as technological advances have emerged over time, text classification and document categorization have been globally used in many domains such as medicine, social sciences, healthcare, psychology, law, engineering, etc. In this section, we highlight several domains which make use of text classification techniques.

\subsection{Text Classification Applications}\unskip

\subsubsection{Information Retrieval}

Information retrieval is finding documents of an unstructured data that meet an information need from within large collections of documents~\cite{schutze2008introduction}. With the rapid growth of online information, particularly in text format, text classification has become a  significant technique for managing this type of data~\cite{hoogeveen2018web}. Some of the important methods used in this area are Na\"ive Bayes, SVM, decision tree, J48, KNN, and IBK~\cite{dwivedi2016automatic}. One of the most challenging applications for document and text data set processing is applying document categorization methods for information retrieval~\cite{jones1971automatic,croft2010search}.

\subsubsection{Information Filtering}
Information filtering refers to the selection of relevant information or rejection of irrelevant information from a stream of incoming data. Information filtering systems are typically used to measure and forecast users' long-term interests~\cite{o1997information}. Probabilistic models, such as the Bayesian inference network, are commonly used in information filtering systems. Bayesian inference networks employ recursive inference to propagate values through the inference network and return documents with the highest ranking~\cite{croft2010search}. Buckley, C.~\cite{buckley1985implementation} used vector
space model with iterative refinement for filtering task.

\subsubsection{Sentiment Analysis}
Sentiment analysis is a computational approach toward identifying opinion, sentiment, and subjectivity in text~\cite{pang2008opinion}. Sentiment classification methods classify a document associated with an opinion to be positive or negative. The assumption is that document~$d$ is expressing an opinion on a single entity $e$ and opinions are formed via a single opinion holder $h$~\cite{liu2012survey}. Naive Bayesian classification and SVM are some of the most popular supervised learning methods that have been used for sentiment classification~\cite{pang2002thumbs}. Features such as terms and their respective frequency, part of speech, opinion words and phrases, negations, and syntactic dependency have been used in sentiment classification techniques.

\subsubsection{Recommender Systems}
Content-based recommender systems suggest items to users based on the description of an item and a profile of the user's interests~\cite{aggarwal2016content}.

A user's profile can be learned from user feedback~(history of the search queries or self reports) on items as well as self-explained features~(filter or conditions on the queries) in one's profile.
In this way, input to such recommender systems can be semi-structured such that some attributes are extracted from free-text field while others are directly specified~\cite{pazzani2007content}. Many different types of text classification methods, such as decision trees, nearest neighbor methods, Rocchio's algorithm, linear classifiers, probabilistic methods, and Naive Bayes, have been used to model user's preference.

\subsubsection{Knowledge Management}
Textual databases are significant sources of information and knowledge. A large percentage of corporate information~(nearly~$80\%$) exists in textual data formats~(unstructured). In knowledge distillation, patterns, or knowledge are inferred from immediate forms that can be semi-structured ({e.g.,}~conceptual graph representation) or structured/relational (e.g., data representation).~A given intermediate form can be document-based such that each entity represents an object or concept of interest in a particular domain. Document categorization is one of the most common methods for mining document-based intermediate forms~\cite{sumathy2013text}. In other work, text classification has been used to find the relationship between railroad accidents' causes and their correspondent descriptions in reports~\cite{heidarysafa2018analysis}.

\subsubsection{Document Summarization}
Text classification used for document summarizing in which the summary of a document may employ words or phrases which do not appear in the original document~\cite{mani1999advances}.  Multi-document summarization also is necessitated due to the rapid increase in online information~\cite{cao2017improving}. Thus, many researchers focus on this task using text classification to extract important features out of a document.

\subsection{Text Classification Support}

\subsubsection{Health}
Most textual information in the medical domain is presented in an unstructured or narrative form with ambiguous terms and typographical errors. Such information needs to be available instantly throughout the patient-physicians encounters in different stages of diagnosis and treatment~\cite{lauria2011combining}. Medical coding, which consists of assigning medical diagnoses to specific class values obtained from a large set of categories, is an area of healthcare applications where text classification techniques can be highly valuable. In other research, {J. Zhang et al.} introduced Patient2Vec to learn an interpretable deep representation of longitudinal electronic health record~(EHR) data which is personalized for each patient~\cite{zhang2018patient2vec}. Patient2Vec is a novel technique of text data set feature embedding that can learn a personalized interpretable deep representation of EHR data based on recurrent neural networks and the attention mechanism. Text classification has also been applied in the development of Medical Subject Headings~(MeSH) and Gene Ontology~(GO)~\cite{trieschnigg2009mesh}.

\subsubsection{Social Sciences}
Text classification and document categorization has increasingly been applied to understanding human behavior in past decades~\cite{nobles2018identification,ofoghi2017textual}. Recent data-driven efforts in human behavior research have focused on mining language contained in informal notes and text data sets, including short message service~(SMS), clinical notes, social media, etc.~\cite{nobles2018identification}. These studies have mostly focused on using approaches based on frequencies of word occurrence (i.e., how often a word appears in a document) or features based on linguistic inquiry word count~(LIWC)~\cite{Pennebaker2015}, a well-validated lexicon of categories of words with psychological relevance \cite{Paul2017}.

\subsubsection{Business and Marketing}
Profitable companies and organizations are progressively using social media for marketing purposes~\cite{yu2011classifying}. Opening mining from social media such as Facebook, Twitter, and so on is main target of companies to rapidly increase their profits~\cite{kang2018opinion}. Text and documents classification is a powerful tool for companies to find their customers more easily.

\subsubsection{Law}
Huge volumes of legal text information and documents have been generated by government institutions. Retrieving this information and automatically classifying it can not only help lawyers but also their clients~\cite{turtle1995text}.
In the United States, the law is derived from five sources: Constitutional law, statutory law, treaties, administrative regulations, and the common law~\cite{bergman2016represent}. Many new legal documents are created each year. Categorization of these documents is the main challenge for the lawyer community.

\section{Conclusions}
The classification task is one of the most indispensable problems in machine learning. As~text and document data sets proliferate, the development and documentation of supervised machine learning algorithms becomes an imperative issue, especially for text classification. Having a better document categorization system for this information requires discerning these algorithms. However, the existing text classification algorithms work more efficiently if we have a better understanding of feature extraction methods and how to evaluate them correctly. Currently, text classification algorithms can be chiefly classified in the following manner: (\RNum{1}) Feature extraction methods, such as Term Frequency-Inverse document frequency~(TF-IDF), term frequency~(TF), word-embedding ({e.g.,}~Word2Vec,  contextualized word representations, Global Vectors for Word Representation~(GloVe), and FastText), are widely used in both academic and commercial applications. In this paper, we had addressed these techniques. However, text and document cleaning could help the accuracy and robustness of an application. We described the basic methods of text pre-processing step.~(\RNum{2}) Dimensionality reduction methods, such as principal component analysis~(PCA), linear discriminant analysis~(LDA), non-negative matrix factorization~(NMF), random projection, Autoencoder, and t-distributed Stochastic Neighbor Embedding~(t-SNE), can be useful in reducing the time complexity and memory complexity of existing text classification algorithms. In a separate section, the most common methods of dimensionality reduction were presented. (\RNum{3}) Existing classification algorithms, such as the Rocchio algorithm, bagging and boosting, logistic regression~(LR), Na\"ive Bayes Classifier~(NBC), k-nearest Neighbor~(KNN), Support Vector Machine~(SVM), decision tree classifier~(DTC), random forest, conditional random field~(CRF), and deep learning, are the primary focus of this paper. (\RNum{4}) Evaluation methods, such as accuracy, $F_{\beta}$, Matthew correlation coefficient~(MCC), receiver  operating  characteristics~(ROC), and area  under curve~(AUC), are explained. With these metrics, the text classifcation algorithm can be evaluated. (\RNum{5}) Critical limitations of each component of the text classification pipeline~(i.e., feature extraction, dimensionality reduction, existing classification algorithms, and evaluation) were addressed in order to each technique. And finally we compare the most common text classification algorithm in this section. (\RNum{5}) Finally, the usage of text classification as an application and/or support other majors such as lay, medicine, etc. are covered in a separate section.

In this survey, Recent techniques and trending of text classification algorithm have discussed.

\vspace{6pt}
\authorcontributions{K.K., M.H, and K.J.M. worked on the idea and designed the platform, and also they worked on GitHub sample code for all of these models.  M.H. and S.M. organized and proofread the paper. This work is under the supervision of L.B. and D.B.}

\funding{This work was supported by The United States Army Research Laboratory under Grant W911NF-17-2-0110.}

\acknowledgments{The authors would like to thank Matthew S. Gerber for his feedback and comments.}

\conflictsofinterest{The authors declare no conflict of interest. The funding sponsors had no role in the design
of the study; in the collection, analyses or interpretation of data; in the writing of the manuscript; nor in the
decision to publish the results.}

\reftitle{References}

\end{document}